\documentclass[10pt,a4paper]{article}

\usepackage[utf8]{inputenc} % allow utf-8 input
\usepackage[T1]{fontenc}    % use 8-bit T1 fonts
\usepackage{amsmath}
\usepackage{amsfonts}
\usepackage{mathtools}
\usepackage{amssymb}
\usepackage{graphicx}
\usepackage{bm}
\usepackage{color}
\usepackage{float}
\usepackage{authblk}
\usepackage{subcaption}
\usepackage{caption}
\usepackage{hyperref}       % hyperlinks
\usepackage{url}            % simple URL typesetting
\usepackage{booktabs}       % professional-quality tables
\usepackage{nicefrac}       % compact symbols for 1/2, etc.
\usepackage{microtype}      % microtypography
\usepackage{algorithm,algorithmic}    %for writing the algorithm

\usepackage{cancel}

\usepackage[toc,page]{appendix}

\usepackage[authoryear, round]{natbib}
\title{A Fully Natural Gradient Scheme for Improving Inference of the Heterogeneous Multi-Output Gaussian Process Model}
\author{Juan-Jos\'e Giraldo and Mauricio A. \'Alvarez}
\affil{{\small Department of Computer Science, University of Sheffield}}
\date{}

\begin{document}
\maketitle

\begin{abstract}
	A recent novel extension of multi-output Gaussian processes handles
	heterogeneous outputs assuming that each output has its own
	likelihood function. It uses a vector-valued Gaussian process prior
	to jointly model all likelihoods' parameters as latent functions
	drawn from a Gaussian process with a linear model of
	coregionalisation covariance. By means of an inducing points
	framework, the model is able to obtain tractable variational bounds
	amenable to stochastic variational inference. Nonetheless, the
	strong conditioning between the variational parameters and the
	hyper-parameters burdens the adaptive gradient optimisation methods
	used in the original approach. To overcome this issue we borrow
	ideas from variational optimisation introducing an exploratory
	distribution over the hyper-parameters, allowing inference together
	with the posterior's variational parameters through a fully natural
	gradient optimisation scheme. Furthermore, in this work we introduce
	an extension of the heterogeneous multi-output model, where its
	latent functions are drawn from convolution processes. We show that
	our optimisation scheme can achieve better local optima solutions
	with higher test performance rates than adaptive gradient methods,
	this for both the linear model of coregionalisation and the
	convolution processes model. We also show how to make the
	convolutional model scalable by means of stochastic
	variational inference and how to optimise it through a fully
	natural gradient scheme. We compare the performance of the
	different methods over toy and real databases.
\end{abstract}

\section{Introduction}
A Multi-Output Gaussian Processes (MOGP) model
generalises the Gaussian Process (GP) model by exploiting correlations
not only in the input space, but also in the output space
\citep{Alvarez2012}. Major research about MOGP models has focused on
finding proper definitions of a cross-covariance function between the
multiple outputs \citep{Journel1978,Higdon2002}. Nevertheless few works
have been concerned about targeting the issue that those outputs not
necessarily follow the same statistical data type. To address that
regard, a recent approach known as the Heterogeneous Multi-Output
Gaussian Process (HetMOGP) model extents the MOGP application
\citep{Alvarez2011} to any arbitrary combination of $D$ likelihood
distributions over the output observations \citep{pablo2018}. The
HetMOGP jointly models all likelihoods' parameters as latent functions
drawn from a Gaussian process with a linear model of coregionalisation
(LMC) covariance. It can be seen as a generalisation of a Chained GP
\citep{saul2016} for multiple correlated output functions of an
heterogeneous nature. The HetMOGP's scalability bases on the
schemes of variational inducing variables for single-output GPs \citep{Hensman2013}. This scheme relies on the
idea of augmenting the GP prior probability space, through the
inclusion of a so-called set of inducing points that change the full
GP covariance by a low-rank approximation
\citep{Quinonero-Candela2005,Edward2006}. Such inducing points help
reducing significantly the MOGP's computational costs from
$\mathcal{O}(D^3N^3)$ to $\mathcal{O}(DNM^2)$ and storage from
$\mathcal{O}(D^2N^2)$ to $\mathcal{O}(DNM)$, where $N$, $D$ and
$M \ll N$ represent the number of data observations, outputs and
inducing points, respectively \citep{Rasmussen2006,Alvarez2011}.

The adequate performance of a variational GP model depends on a proper
optimisation process able to find rich local optima solutions for
maximising a bound to the marginal likelihood. Variational GP
models generally suffer from strong conditioning between the
variational posterior distribution, the multiple hyper-parameters of
the GP prior and the inducing points \citep{Mark2018}. In particular,
the HetMOGP model is built upon a linear combinations of $Q$ latent
functions, where each latent function demands a treatment based on the
inducing variables framework. On this model then, such strong
conditionings are enhanced even more due to the dependence of inducing
points per underlying latent function, and the presence of additional
linear combination coefficients. Since the model is extremely
sensitive to any small change on any of those variables, stochastic
gradient updates in combination with adaptive gradient methods (AGMs,
e.g. Adam) tend to drive the optimisation to poor local minima.

With the purpose to overcome the optimisation problems present in
variational GP models, there has recently been a growing interest in
alternative optimisation schemes that adopt the natural gradient (NG)
direction \citep{Amari1998}. For instance, in \citep{Hensman2013} the
authors derived a mathematical analysis that suggested we can make
better progress when optimising a variational GP along the NG direction, but without providing any experimental results of
its performance. The authors in \citep{Khan2015} propose to linearise
the non-conjugate terms of the model for admitting closed-form updates
which are equivalent to optimising in the natural gradient
direction. The work in \citep{Khan17} shows how to convert inference in
non-conjugate models as it is done in the conjugate ones, by way of
expressing the posterior distribution in the mean-parameter
space. Furthermore, it shows that by means of exploiting the mirror
descent algorithm (MDA) one can arrive to NG updates for
tuning the variational posterior distribution. Those works coincide in
improvements of training and testing performance, and also fast
convergence rates. Nonetheless, they only show results in a full GP
model where the kernel hyper-parameters are fixed using a grid
search. On the other hand, the work in \citep{Hugh2018} does show a
broad experimental analysis of the NG method for sparse
GPs. The authors conclude that the NG is not prone to
suffer from ill-conditioning issues in comparison to the AGMs. Also
the NG has been used to ease optimisation of the
variational posterior over the latent functions of a deep GP model
\citep{salimbeni2019}. However, in those two latter cases the NG method only applies for the latent functions' posterior
parameters, while an Adam method performs a cooperative optimisation
for dealing with the hyper-parameters and inducing points. The authors
in \citep{Hugh2018} call this strategy a hybrid between NG and Adam, and termed it NG+Adam. 

The main contributions of this paper include the following:
\begin{itemize}
	\item We propose a fully natural gradient (FNG) scheme for jointly
	tuning the hyper-parameters, inducing points and variational
	posterior parameters of the HetMOGP model. To this end, we borrow ideas from different
	variational optimisation (VO) strategies like
	\citep{Staines2013,Khan2017VAN} and \citep{Khan2018}, by introducing an
	exploratory distribution over the hyper-parameters and inducing
	points. Such VO strategies have shown to be successful
	exploratory-learning tools able to avoid poor local optima solutions;
	they have been broadly studied in the context of reinforcement and
	Bayesian deep learning, but not much in the context of
	GPs.
	\item We provide an extension of the HetMOGP based on a Convolution Processes (CPM) model, rather than an LMC approach
	as in the original model. This is a novel contribution since there are no former MOGP models with convolution processes that involve stochastic variational inference (SVI),
	nor a model of heterogeneous outputs that relies on convolution
	processes. 
	\item We provide a FNG scheme for optimising the new model extension, the HetMOGP with CPM.
	\item To the best
	of our knowledge the NG method has not been performed over any MOGP
	model before. Hence, in this work we also contribute to show how a NG method used in a full scheme over the MOGP's
	parameters and kernel hyper-parameters alleviates the strong
	conditioning problems. This, by achieving better local optima
	solutions with higher test performance rates than Adam and stochastic
	gradient descent (SGD).
	\item We explore for the first time in a MOGP model
	the behaviour of the hybrid strategy NG+Adam, and provide comparative
	results to our proposed scheme.
\end{itemize}

%\color{red}\hfill mds
%\hfill August 26, 2015
\section{Variational Optimisation: an Exploratory Mechanism for Optimisation}
This section introduces the variational optimisation method as an exploratory mechanism for minimising an objective function \citep{Staines2013}. It also shows how Variational Inference (VI) can be seen as a particular case of variational optimisation.
\subsection{Variational Optimisation}
\begin{figure}
	\centering
	\includegraphics[width=0.5\textwidth]{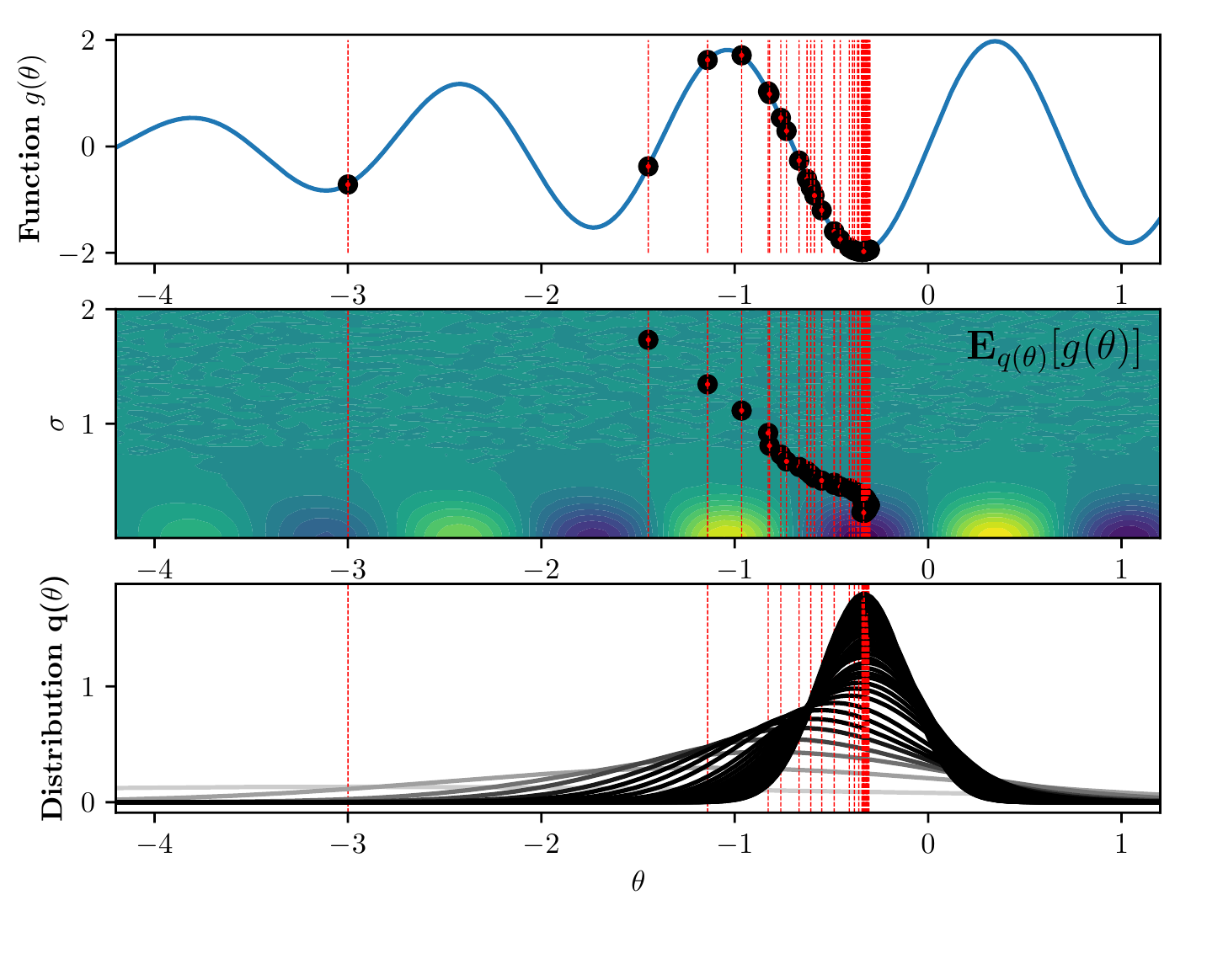}
	\caption{First row shows what happens from the perspective of the original function $g(\theta)=2\exp(-0.09\theta^2)\sin(4.5\theta)$, the black dots represent the position of $\theta=\mu$ at each iteration. Second row shows a contour graph of the space of solutions w.r.t $\sigma$ and $\mu$, here the black dots refer to the position of $\sigma$ and $\mu$ at each iteration, and the low and high colour intensities relate to low and high values of $\mathbb{E}_{q(\theta)}[g(\theta)]$, notice that here we do not include the KL term information for easing the visualisation of the multiple local minima. Third row shows $q(\theta)$'s behaviour, for each Gaussian bell we use a colour code from light-gray to black for representing initial to final stages of the inference. All sub-graphs present vertical lines for aligning iterations, i.e., from left to right the lines represent the occurrence of an iteration. To avoid excessive overlapping, the third row only shows $q(\theta)$ every two iterations.}
	\label{fig:variational_opt} 
\end{figure}

The goal in optimisation is to find a proper set of parameters that minimise a possibly non-convex function $g(\boldsymbol{\theta})$ by solving, $\boldsymbol{\theta}^*=\arg \min_{\boldsymbol{\theta}}g(\boldsymbol{\theta})$, where $\boldsymbol{\theta}^*$ represents the set of parameters that minimise the function. The classical way to deal with the above optimisation problem involves deriving w.r.t $\boldsymbol{\theta}$ and solving in a closed-form, or through a gradient descent method. Usually, gradient methods tend to converge to the closest local minima from the starting point without exploring much the space of solutions \citep{Chong2013} (see appendix A for a comparison between VO and Newton's method). Alternatively the variational optimisation method proposes to solve the same problem \citep{Staines2013}, but introducing exploration in the parameter space of a variational (or exploratory) distribution $q(\boldsymbol{\theta}|\boldsymbol{\psi})$ by bounding the function $g(\boldsymbol{\theta})$ as follows:
\begin{align}
\tilde{\mathcal{L}}(\boldsymbol{\psi})=\mathbb{E}_{q(\boldsymbol{\theta}|\boldsymbol{\psi})}[g(\boldsymbol{\theta})]+\mathbb{D}_{KL}\big(q(\boldsymbol{\theta}|\boldsymbol{\psi})||p(\boldsymbol{\theta})\big),
\label{eq:VO}
\end{align}
where $\mathbb{D}_{KL}(\cdot||\cdot)$ is a Kullback-Leibler (KL) divergence and $p(\boldsymbol{\theta})$ is a penalization distribution. The work of VO in \citep{Staines2013} does not introduce the KL term in the equation above, i.e. $\tilde{\mathcal{L}}(\boldsymbol{\psi})=\mathbb{E}_{q(\boldsymbol{\theta}|\boldsymbol{\psi})}[g(\boldsymbol{\theta})]$, this implies that during an inference process, the exploratory distribution is free to collapse to zero becoming a Dirac's delta $q(\boldsymbol{\theta})=\delta(\boldsymbol{\theta}-\boldsymbol{\mu})$, where $\boldsymbol{\mu}=\boldsymbol{\theta}^*$  and $\boldsymbol{\mu}$ represents the $q(\boldsymbol{\theta})$'s mean \citep{Wierstra2014,Hensman2015b}. This collapsing effect limits the exploration of $\boldsymbol{\theta}$'s space (see appendix A for a graphical example). In contrast, by using the KL term, we can force the exploratory distribution $q(\boldsymbol{\theta}|\boldsymbol{\psi})$ to trade-off between minimising the expectation $\mathbb{E}_{q(\boldsymbol{\theta}|\boldsymbol{\psi})}[g(\boldsymbol{\theta})]$ and not going far away from the imposed $p(\boldsymbol{\theta})$ penalization \citep{Khan2017Vprop}. Indeed, the KL term in Eq. \eqref{eq:VO} reduces the collapsing effect of $q(\boldsymbol{\theta})$ and helps to gain additional exploration when an inference process is carried out. With the aim to better understand such behaviour, let us define an example inspired by the one in \citep{Khan2017VAN}; we define $g(\theta)=2\exp(-0.09\theta^2)\sin(4.5\theta)$, a
function with multiple local minima, $q(\theta)=\mathcal{N}(\theta|\mu,\sigma^2)$ represents a variational distribution over $\theta$, with parameters mean $\mu$ and variance $\sigma^2$, and $p(\theta)=\mathcal{N}(\theta|0,\lambda^{-1})$ with $\lambda=1.5$. We built a graphical experiment to show what happens at
each iteration of the optimisation process. Figure
\ref{fig:variational_opt} shows three perspectives of a such
experiment, where we initialise the parameters $\theta=\mu=-3.0$ and
$\sigma=3.0$.
We can notice from Fig. \ref{fig:variational_opt} that the initial
value of $\theta=\mu=-3.0$ is far away from $g(\theta)$'s global
minimum at $\theta\approx -0.346$. When the inference
process starts, the exploratory distribution $q(\theta)$ modifies its
variance and moves its mean towards a better region in the space of $\theta$. From the third row we can also see that
$q(\theta)$ initially behaves as a broad distribution (in light-gray
colour) with a mean located at $\mu=-3.0$, while the iterations
elapse, the distribution $q(\theta)$ modifies its shape in order to
reach a better local minima solution (at $\mu\approx -0.346$). The distribution
presents such behaviour in spite of being closer to other poor local
minima like the ones between the intervals $(-4,-3)$ and
$(-2,-1)$. Additionally, when the mean $\mu$ is close to $\theta\approx -0.346$
(the global minimum), the variance parameter reduces constantly making
the distribution look narrower, which means it is increasing the certainty
of the solution. This behaviour implies that in the long term $q(\theta)$'s mean will be much closer to $\theta^*$. Therefore, a
feasible minima solution for the original objective function
$g(\theta)$ is $\theta=\mathbb{E}_{q(\theta)}[\theta]=\mu$, this can be seen in the first sub-graph
where at each iteration $\theta=\mu$, in fact, at the end $\mu$ is
fairly close to the value $\theta\approx -0.346$.

\subsection{Variational Inference: VO for the Negative Log Likelihood}
A common way to build a probabilistic model for a set of observations
$\mathbf{X}=\{\mathbf{x}_n\}^N_{n=1} \in \mathbb{R}^{N\times P}$ is to
assume that each observation is drawn
independently and identically distributed
(IID) from a probability distribution
$p(\mathbf{X}|\boldsymbol{\theta})$, commonly known as a
likelihood. Fitting the model consists on finding the parameter
$\boldsymbol{\theta}$ that makes the distribution appropriately
explain the data. This inference process is called maximum likelihood
estimation, given that is equivalent to the optimisation problem of
maximising the log likelihood function
$\log p(\mathbf{X}|\boldsymbol{\theta})$, i.e., minimising the
negative log likelihood (NLL) function
$-\log p(\mathbf{X}|\boldsymbol{\theta})$ \citep{Murphy}. From a
Bayesian perspective, we can introduce a prior distribution
$p(\boldsymbol{\theta})$ over the parameter of interest, $p(\boldsymbol{\theta}|\mathbf{X}) \propto p(\mathbf{X}|\boldsymbol{\theta})p(\boldsymbol{\theta})$, which implies that there also exists a posterior distribution
$p(\boldsymbol{\theta}|\mathbf{X})$ over such parameter, useful to
render future predictions of the model. When the likelihood and prior
are conjugate, the posterior distribution can be
computed in closed form, but that is not always the case. Hence, if
the likelihood and prior are non-conjugate, it is
necessary to approximate the posterior \citep{Bishop2006}. Variational
inference is a powerful framework broadly used in machine
learning, that allows to estimate the posterior by minimising the KL
divergence
$\mathbb{D}_{KL}\big(q(\boldsymbol{\theta}|\boldsymbol{\psi})||p(\boldsymbol{\theta}|\mathbf{X})\big)$
between an approximate variational posterior
$q(\boldsymbol{\theta}|\boldsymbol{\psi})$ and the true posterior
$p(\boldsymbol{\theta}|\mathbf{X})$ \citep{Blei2017}. Since we do not
have access to the true posterior, minimising such KL divergence is
equivalent to maximising a lower bound to the marginal likelihood. It
emerges from the equality: $\log \mathbb{E}_{q(\boldsymbol{\theta}|\boldsymbol{\psi})}\Big[\frac{p(\mathbf{X}|\boldsymbol{\theta})p(\boldsymbol{\theta})}{q(\boldsymbol{\theta}|\boldsymbol{\psi})}\Big] = \log p(\mathbf{X})$, in which, after applying the Jensen's inequality we arrive to,
\begin{align}
-\tilde{\mathcal{L}}(\boldsymbol{\psi})=\mathbb{E}_{q(\boldsymbol{\theta}|\boldsymbol{\psi})}\Bigg[\log \frac{p(\mathbf{X}|\boldsymbol{\theta})p(\boldsymbol{\theta})}{q(\boldsymbol{\theta}|\boldsymbol{\psi})}\Bigg]\leq \log p(\mathbf{X}),
\label{eq:negative_ELBO}
\end{align}  
where $\log p(\mathbf{X})$ represents the log marginal likelihood and
$-\tilde{\mathcal{L}}(\boldsymbol{\psi})$ is an evidence lower bound
(ELBO) \citep{Jordan1999}. It is noteworthy that if we replace
$g(\boldsymbol{\theta})=-\log p(\mathbf{X}|\boldsymbol{\theta})$ in
Eq. \eqref{eq:VO}, we end up with exactly the same lower bound of
Eq. \eqref{eq:negative_ELBO}. Therefore, VI can be seen as a
particular case of VO with a KL divergence penalisation, where the
objective $g(\boldsymbol{\theta})$ is nothing but the NLL. We can
distinguish from two perspectives when using VO for maximum
likelihood: for the Bayesian perspective we are not only
interested in a point estimate for the parameter
$\boldsymbol{\theta}$, but in the uncertainty codified in
$q(\boldsymbol{\theta})$'s (co)variance for making future
predictions; and for the non-Bayesian perspective the main goal in maximum likelihood estimation is to optimise the function
$g(\boldsymbol{\theta})=-\log p(\mathbf{X}|\boldsymbol{\theta})$. For
this case, if $q(\boldsymbol{\theta}|\boldsymbol{\mu},\boldsymbol{\Sigma})$ is a Gaussian distribution, we can make use of only the posterior's mean
$\mathbb{E}_{q(\boldsymbol{\theta}|\boldsymbol{\psi})}[\boldsymbol{\theta}]=\boldsymbol{\mu}$ as a feasible solution for $\boldsymbol{\theta}^*$ without taking into
account the uncertainty. This is also known as the maximum a
posteriori (MAP) solution in the context of VI, due
to the fact that
$\boldsymbol{\theta}_{\text{MAP}}=\arg \max_{\boldsymbol{\theta}}
p(\boldsymbol{\theta}|\mathbf{X})\approx q(\boldsymbol{\theta}|\boldsymbol{\mu},\boldsymbol{\Sigma})$,
where the maximum of the distribution
$q(\boldsymbol{\theta}|\boldsymbol{\mu},\boldsymbol{\Sigma})$ is
located at its mean, thereby
$\boldsymbol{\theta}_{\text{MAP}}=\boldsymbol{\mu}$ (see section VII of
SM for details) \citep{Bishop2006}. 
\section{Exploiting The Mirror Descent Algorithm}

Direct update equations for the parameters of a (posterior) distribution using natural gradients involve the inversion of a Fisher information matrix, which in general it is complex to do. The purpose of this section is to show how an alternative formulation of the NG updates can be derived from the MDA. We introduce the Variational Adaptive-Newton (VAN), a method that benefits from a Gaussian posterior distribution to easily express the parameters updates in the NG direction. And we also introduce the concept of \textit{natural-momentum} which takes advantage of the KL divergence for providing an extra memory information to the MDA.
%\label{ch:background}
\subsection{Connection between Natural-Gradient and Mirror Descent}
The NG allows to solve an optimisation problem like the one in Eq. \eqref{eq:VO}, where the goal consists on finding an optimal distribution $q(\boldsymbol{\theta})$ that best minimises the objective bound \citep{Amari1998}. The method takes advantage of the inverse Fisher information matrix, $\mathbf{F}^{-1}$, associated to the random variable $\boldsymbol{\theta}$, by iteratively weighting the following gradient updates, $\boldsymbol{\lambda}_{t+1}=\boldsymbol{\lambda}_{t}-\alpha_t\mathbf{F}_t^{-1}\hat{\nabla}_{\boldsymbol{\lambda}}\tilde{\mathcal{L}}_{t}$, where $\alpha_t$ is a positive step-size parameter and
$\boldsymbol{\lambda}_t$ represents the natural (or canonical)
parameters of the distribution $q(\boldsymbol{\theta})$. Such natural
parameters can be better noticed by expressing the distribution in the
general form of the exponential family,
$q(\boldsymbol{\theta})=h(\boldsymbol{\theta})\exp\big(\langle\boldsymbol{\lambda},\phi(\boldsymbol{\theta})\rangle-A(\boldsymbol{\lambda})\big)$,
where $A(\boldsymbol{\lambda})$ is the log-partition function,
$\phi(\boldsymbol{\theta})$ is a vector of sufficient statistics and
$h(\boldsymbol{\theta})$ is a scaling constant \citep{Murphy}. The updates for
$\boldsymbol{\lambda}_{t+1}$ are expensive due to involving the
computation of the inverse Fisher matrix at each iteration. Since an
exponential-family distribution has an associated set of
mean-parameters
$\boldsymbol{\eta}=\mathbb{E}[\boldsymbol{\phi}(\boldsymbol{\theta})]$,
then an alternative way to induce the NG updates
consists on formulating a MDA in such
mean-parameter space. Hence, the algorithm bases on solving the
following iterative sub-problems:
\begin{align}
\boldsymbol{\eta}_{t+1}=\arg \min_{\boldsymbol{\eta}} \langle\boldsymbol{\eta},\hat{\nabla}_{\boldsymbol{\eta}}\tilde{\mathcal{L}}_{t}\rangle+\frac{1}{\alpha_t}\mathbb{D}_{KL}(q(\boldsymbol{\theta})||q_t(\boldsymbol{\theta})), \label{eq:mirror}
\end{align}
where $\boldsymbol{\eta}$ is the set of $q(\boldsymbol{\theta})$'s mean-parameters, $\tilde{\mathcal{L}}$ is a VO bound of a function $g(\boldsymbol{\theta})$, $\hat{\nabla}_{\boldsymbol{\eta}}\tilde{\mathcal{L}}_{t}:=\hat{\nabla}_{\boldsymbol{\eta}}\tilde{\mathcal{L}}(\boldsymbol{\eta}_{t})$ denotes a stochastic gradient, $q_t(\boldsymbol{\theta}):=q(\boldsymbol{\theta}|\boldsymbol{\eta}_{t})$ and $\alpha_t$ is a positive step-size parameter \citep{Khan17}. The intention of the above formulation is to exploit the parametrised distribution's structure by controlling its divergence w.r.t its older state $q_t(\boldsymbol{\theta})$. Replacing the distribution $q(\boldsymbol{\theta})$ in its exponential-form, in the above KL divergence, and setting Eq. \eqref{eq:mirror} to zero, let us express,  
\begin{align}
\langle\boldsymbol{\eta},\hat{\nabla}_{\boldsymbol{\eta}}\tilde{\mathcal{L}}_{t}\rangle+\frac{1}{\alpha_t}\big[\langle\boldsymbol{\lambda},\boldsymbol{\eta}\rangle-A(\boldsymbol{\lambda})-\langle\boldsymbol{\lambda}_t,\boldsymbol{\eta}\rangle+A(\boldsymbol{\lambda}_t)\big]=0,\notag
\end{align}
and by deriving w.r.t $\boldsymbol{\eta}$, we arrive to $\boldsymbol{\lambda}_{t+1}=\boldsymbol{\lambda}_{t}-\alpha_t\hat{\nabla}_{\boldsymbol{\eta}}\tilde{\mathcal{L}}_{t}$, where $\boldsymbol{\lambda}_{t+1}:=\boldsymbol{\lambda}$ and $\hat{\nabla}_{\boldsymbol{\eta}}\tilde{\mathcal{L}}_{t}=\mathbf{F}^{-1}\hat{\nabla}_{\boldsymbol{\lambda}}\tilde{\mathcal{L}}_{t}$ as per the work in \citep{Raskutti2015}, where the authors provide a formal proof of such equivalence. The formulation in Eq. \eqref{eq:mirror} is advantageous since it is easier to compute derivatives w.r.t $\boldsymbol{\eta}$ than computing the inverse Fisher information matrix $\mathbf{F}^{-1}$. Therefore, the MDA for solving iterative sub-problems in the mean-parameter space is equivalent to updating the canonical parameters in the NG direction (see appendix B for more details).  
\subsection{Variational Adaptive-Newton and Natural-Momentum}
The VAN method aims to solve the problem in Eq. \eqref{eq:mirror} using a Gaussian distribution $q(\boldsymbol{\theta}):=q(\boldsymbol{\theta}|\boldsymbol{\mu},\boldsymbol{\Sigma})$ as the exploratory mechanism for optimisation \citep{Khan2017VAN}. This implies that if $\boldsymbol{\mu}$ and $\boldsymbol{\Sigma}$ represent the mean and covariance respectively, then $q(\boldsymbol{\theta})$'s mean-parameters are $\boldsymbol{\eta}=\{\boldsymbol{\mu},\boldsymbol{\Sigma}+\boldsymbol{\mu}\boldsymbol{\mu}^{\top}\}$, and also its analogous natural-parameters are $\boldsymbol{\lambda}=\{\boldsymbol{\Sigma}^{-1}\boldsymbol{\mu},-\frac{1}{2}\boldsymbol{\Sigma}^{-1}\}$. When plugging these parametrisations and solving for the MDA in Eq. \eqref{eq:mirror}, we end up with the following updates: $\boldsymbol{\Sigma}_{t+1}^{-1}=\boldsymbol{\Sigma}_{t}^{-1}+2\alpha_t\hat{\nabla}_{\boldsymbol{\Sigma}}\tilde{\mathcal{L}}_t$ and $\boldsymbol{\mu}_{t+1}=\boldsymbol{\mu}_{t}-\alpha_t\boldsymbol{\Sigma}_{t+1}\hat{\nabla}_{\boldsymbol{\mu}}\tilde{\mathcal{L}}_t$,
where $\boldsymbol{\mu}_t$ and $\boldsymbol{\Sigma}_t$ are the mean and covariance parameters at the instant $t$ respectively; the stochastic gradients are  $\hat{\nabla}_{\boldsymbol{\mu}}\tilde{\mathcal{L}}_{t}:=\hat{\nabla}_{\boldsymbol{\mu}}\tilde{\mathcal{L}}(\boldsymbol{\mu}_t,\boldsymbol{\Sigma}_t)$ and $\hat{\nabla}_{\boldsymbol{\Sigma}}\tilde{\mathcal{L}}_{t}:=\hat{\nabla}_{\boldsymbol{\Sigma}}\tilde{\mathcal{L}}(\boldsymbol{\mu}_t,\boldsymbol{\Sigma}_t)$. These latter updates represent a NG descent algorithm for exploring the space of solutions of the variable $\boldsymbol{\theta}$ through a Gaussian distribution \citep{Khan17}. It is possible to keep exploiting the
structure of the distribution $q(\boldsymbol{\theta})$, this by including
an additional KL divergence term in the MDA of Eq. \eqref{eq:mirror} as follows: $\boldsymbol{\eta}_{t+1}$
\begin{align}
=\arg \min_{\boldsymbol{\eta}} \langle\boldsymbol{\eta},\hat{\nabla}_{\boldsymbol{\eta}}\tilde{\mathcal{L}}_{t}\rangle+\frac{1}{\tilde{\alpha}_t}\text{KL}(\boldsymbol{\theta})_t -\frac{\tilde{\gamma}_t}{\tilde{\alpha}_t}\text{KL}(\boldsymbol{\theta})_{t-1},\label{eq:mirror_mom}
\end{align}
where $q_{t}(\boldsymbol{\theta}):=q(\boldsymbol{\theta}|\boldsymbol{\mu}_{t},\boldsymbol{\Sigma}_{t})$ represents the exploratory distributions $q(\boldsymbol{\theta})$ with the parameters obtained at time $t$, and  $\text{KL}(\cdot)_t:=\mathbb{D}_{KL}(q(\cdot)||q_t(\cdot))$. Such additional KL term, called as a \textit{natural-momentum} in \citep{Khan2018}, provides extra memory information to the MDA for potentially improving its convergence rate. This momentum can be controlled by the relation between the positive step-sizes $\tilde{\alpha}_t$ and $\tilde{\gamma}_t$. When solving for Eq. \eqref{eq:mirror_mom}, we arrive to the following NG update equations:
\begin{align}
\boldsymbol{\Sigma}_{t+1}^{-1}=\boldsymbol{\Sigma}_{t}^{-1}&+2\alpha_t\hat{\nabla}_{\boldsymbol{\Sigma}}\tilde{\mathcal{L}}_t\label{eq:update_nat1}\\
\boldsymbol{\mu}_{t+1}=\boldsymbol{\mu}_{t}-&\alpha_t\boldsymbol{\Sigma}_{t+1}\hat{\nabla}_{\boldsymbol{\mu}}\tilde{\mathcal{L}}_t+\gamma_t\boldsymbol{\Sigma}_{t+1}\boldsymbol{\Sigma}^{-1}_{t}(\boldsymbol{\mu}_t-\boldsymbol{\mu}_{t-1})\label{eq:update_nat2},
\end{align}    
where $\alpha_t=\tilde{\alpha}_t/(1-\tilde{\gamma}_t)$ and $\gamma_t=\tilde{\gamma}_t/(1-\tilde{\gamma}_t)$ are positive step-size parameters \citep{Khan2017Vprop,Khan2018}. 
\section{Heterogeneous Multi-Output Gaussian Process Model}
This section provides a brief summary of the state of the art in multi
output GPs. It later describes the HetMOGP model. Also,
how the inducing points framework allows the model to obtain tractable
variational bounds amenable to SVI.
\subsection{Multi-Output Gaussian Processes Review}
A MOGP generalises the GP model by exploiting correlations not only in
the input space, but also in the output space \citep{Alvarez2012}. Major research about MOGPs has focused on finding
proper definitions of a cross-covariance function between multiple
outputs. Classical approaches that define such cross-covariance
function include the LMC
\citep{Journel1978} or process convolutions
\citep{Higdon2002}. The works in \citep{Alvarez2012,Alvarez2011} provide a review of MOGPs that use either LMC or convolution processes approaches. MOGPs
have been applied in several problems including sensor networks with
missing signals \citep{Michael2008}; motion capture data for completing
a sequence of missing frames \citep{Zhao2016}; and natural language
processing, where annotating linguistic data is often a complex and
time consuming task, and MOGPs can learn from the outputs of multiple
annotators \citep{Cohn2013}. They have been also used in computer
emulation, where the LMC, also termed as a Multiple-Output emulator,
can be used as a substitute of a computationally expensive
deterministic model \citep{Conti2009,Conti2010}. Likewise, MOGPs have been useful for learning the couplings between multiple time series and helping to enhance their forecasting capabilities \citep{Boyle2005}. Recent approaches have focused on building
cross-covariances between outputs in the spectral domain
\citep{Parra2017}. For instance, by constructing a
multi-output Convolution Spectral Mixture kernel which incorporates
time and phase delays in the spectral density \citep{Chen2019}. Other
works have concentrated in tackling the issues regarding inference
scalability and computation efficiency \citep{Alvarez2011}, for example
in the context of large datasets using collaborative MOGPs
\citep{Nguyen2014}; introducing an scalable inference procedure with a
mixture of Gaussians as a posterior approximation \citep{Dezfouli2015}. Other works have
investigated alternative paradigms to MOGPs. For instance, the work in
\citep{Gordon2012} has explored combinations of GPs with Bayesian
neural networks (BNN) so as to take advantage from the GPs’
non-parametric flexibility and the BNN’s structural properties for
modelling multiple-outputs. Another recent work relies on a
product rule to decompose the joint distribution of the outputs given
the inputs into conditional distributions, i.e. decoupling the model
into single-output regression tasks
\citep{Requeima2019}. Most work on MOGPs including \citep{Conti2010,Chen2019,Requeima2019} has focused on Gaussian multivariate regression. As
we have mentioned before, in this paper we focus on the HetMOGP that concerns about outputs with different statistical data types, and extends the MOGPs'
application to heterogeneous outputs \citep{pablo2018}.
\subsection{The Likelihood Function for the HetMOGP}
The HetMOGP model is an extension of the Multi-Output GP that allows different kinds of likelihoods as per the statistical data type each output demands \citep{pablo2018}. For instance, if we have two outputs problem, where one output is binary $y_1\in \{0,1\}$ while the other is a real value $y_2\in \mathbb{R}$, we can assume our likelihood as the product of a Bernoulli and Gaussian distribution for each output respectively. In general the HetMOGP likelihood for $D$ outputs can be written as:
\begin{align}
p(\mathbf{y}|\mathbf{f})=\prod_{n=1}^{N}\prod_{d=1}^{D}p(y_{d,n}|\psi_{d,1}(\mathbf{x}_n),...,\psi_{d,J_d}(\mathbf{x}_n)),\label{eq:likelihood}
\end{align}
where the vector $\mathbf{y}=[\mathbf{y}^{\top}_1,...,\mathbf{y}^{\top}_D]^{\top}$ groups all the output observations and each $\psi_{d,j}(\mathbf{x}_n)$ represents the $j$-th parameter that belongs to the $d$-th likelihood. It is worth noticing that each output vector $\mathbf{y}_d$ can associate a particular set of input observations $\mathbf{X}_d$. Though, in order to ease the explanation of the model and to be consistent with the equation above, we have assumed that all outputs $\mathbf{y}_d=[y_{d,1},...,y_{d,N}]^\top$ associate the same input observations $\mathbf{X}=[\mathbf{x}_{1},...,\mathbf{x}_{N}]^\top \in \mathbb{R}^{N\times P}$. Each likelihoods' parameter $\psi_{d,j}(\mathbf{x}_n)$ is plugged to a latent function $f_{d,j}(\cdot)$ that follows a GP prior, through a link function $\phi(\cdot)$, i.e., $\psi_{d,j}(\mathbf{x}_n)=\phi(f_{d,j}(\mathbf{x}_n))$. For instance, if we have two outputs where the first likelihood is a Heteroscedastic Gaussian, then its parameters mean and variance are respectively chained as $\psi_{1,1}(\mathbf{x}_n)=f_{1,1}(\mathbf{x}_n)$ and $\psi_{1,2}(\mathbf{x}_n)=\exp(f_{1,2}(\mathbf{x}_n))$; if the second likelihood is a Gamma, its parameters are linked as $\psi_{2,1}(\mathbf{x}_n)=\exp(f_{2,1}(\mathbf{x}_n))$ and $\psi_{2,2}(\mathbf{x}_n)=\exp(f_{2,2}(\mathbf{x}_n))$ \citep{saul2016}. Notice that $J_d$ accounts for the number of latent functions necessary to parametrise the $d$-th likelihood, thus the total number of functions $f_{d,j}(\cdot)$ associated to the model becomes $J=\sum_{d=1}^{D}J_d$. Each $f_{d,j}(\cdot)$ is considered a latent
parameter function (LPF) that comes from a LMC as follows:   
\begin{align}
f_{d,j}(\mathbf{x})=\sum_{q=1}^{Q}\sum_{i=1}^{R_q}a^i_{d,j,q}u^i_q(\mathbf{x}), 
\label{eq:linear_comb2}
\end{align}
where $u^i_q(\mathbf{x})$ are IID samples from GPs $u_q(\cdot)\sim\mathcal{GP}(0,k_q(\cdot,\cdot))$ and $a^i_{d,j,q}\in \mathbb{R}$ is a linear combination coefficient (LCC). In Section \ref{sec:CPM}, we introduce a different way to model
$f_{d,j}(\mathbf{x})$ based on convolution processes. For the sake of future explanations let us assume that $R_{q}=1$. In this way the number of LCCs per latent function $u_q(\cdot)$ becomes $J$. Such number of coefficients per function $u_q(\cdot)$ can be grouped in a vector $\mathbf{W}_q = [a_{1, 1, q},...,a_{1, J_1, q},...,a_{D, J_D, q}]^\top \in \mathbb{R}^{J\times 1}$; and we can cluster all vectors $\mathbf{W}_q$ in a specific vector of LCCs $\mathbf{w} = [\text{vec}(\mathbf{W}_1)^\top,...,\text{vec}(\mathbf{W}_Q)^\top]^\top \in \mathbb{R}^{QJ\times 1}$.
\subsection{The Inducing Points Method}
A common approach for reducing computational complexity in GP models
is to augment the GP prior with a set of \textit{inducing
	variables}. For the specific case of the HetMOGP model with LMC prior, the vector of
\textit{inducing variables}
$\mathbf{u}=[\mathbf{u}_1^\top,...,\mathbf{u}_Q^\top]^\top \in
\mathbb{R}^{QM\times 1}$ is built from $\mathbf{u}_{q}=[u_{q}(\mathbf{z}^{(1)}_{q}),...,u_{q}(\mathbf{z}^{(M)}_{q})]^\top \in \mathbb{R}^{M\times 1}$. Notice that the vector $\mathbf{u}_{q}$ is constructed by additional evaluations of the functions $u_q(\cdot)$ at some unknown inducing points $\mathbf{Z}_q=[\mathbf{z}^{(1)}_{q},...,\mathbf{z}^{(M)}_{q}]^\top \in \mathbb{R}^{M\times P}$. The vector of all inducing variables can be expressed as $\mathbf{Z}=[\text{vec}(\mathbf{Z}_1)^\top,...,\text{vec}(\mathbf{Z}_Q)^\top]^\top
\in \mathbb{R}^{QMP\times 1}$ \citep{Edward2006,Titsias2009}. We can write the augmented GP prior as follows,
\begin{align}
p(\mathbf{f}|\mathbf{u})p(\mathbf{u})=\prod_{d=1}^{D}\prod_{j=1}^{J_d}p(\mathbf{f}_{d,j}|\mathbf{u})\prod_{q=1}^{Q}p(\mathbf{u}_{q}),
\label{eq:poster_HetMOGP}
\end{align}
where $\mathbf{f}=[\mathbf{f}^{\top}_{1,1},...,\mathbf{f}^{\top}_{1,J_1},...,\mathbf{f}^{\top}_{D,J_D}]^\top$ is a vector-valued function built from $\mathbf{f}_{d,j}=[f_{d,j}(\mathbf{x}_{n}),...,f_{d,j}(\mathbf{x}_N)]^\top \in \mathbb{R}^{N\times 1}$. Following the conditional Gaussian properties we can express, 
\begin{align}
p(\mathbf{f}_{d,j}|\mathbf{u})=\mathcal{N}(\mathbf{f}_{d,j}|\mathbf{A}_{\mathbf{f}_{d,j}\mathbf{u}}\mathbf{u},\widetilde{\mathbf{Q}}_{\mathbf{f}_{d,j}\mathbf{f}_{d,j}}), p(\mathbf{u})=\mathcal{N}(\mathbf{u}|\mathbf{0},\mathbf{K}_\mathbf{uu}),\notag
\end{align}
where the matrix $\mathbf{K}_\mathbf{uu}\in \mathbb{R}^{QM\times QM}$ is a
block-diagonal with blocks $\mathbf{K}_{\mathbf{u}_q\mathbf{u}_q}\in \mathbb{R}^{M\times M}$ built from evaluations of
$\operatorname{cov}\left[u_q(\cdot), u_q(\cdot)\right]=k_{q}(\cdot,
\cdot)$ between all pairs of inducing points $\mathbf{Z}_q$
respectively; and we have introduced the following definitions,
$
\mathbf{A}_{\mathbf{f}_{d,j}\mathbf{u}}=\mathbf{K}_{\mathbf{f}_{d,j}\mathbf{u}}\mathbf{K}^{-1}_\mathbf{uu},\quad\widetilde{\mathbf{Q}}_{\mathbf{f}_{d,j}\mathbf{f}_{d,j}} = \mathbf{K}_{\mathbf{f}_{d,j}\mathbf{f}_{d,j}}-\mathbf{Q}_{\mathbf{f}_{d,j}\mathbf{f}_{d,j}}$, $ \mathbf{Q}_{\mathbf{f}_{d,j}\mathbf{f}_{d,j}}=\mathbf{K}_{\mathbf{f}_{d,j}\mathbf{u}}\mathbf{K}^{-1}_\mathbf{uu}\mathbf{K}_{\mathbf{u}\mathbf{f}_{d,j}}$, $ \mathbf{K}_{\mathbf{f}_{d,j}\mathbf{u}}=\mathbf{K}^{\top}_{\mathbf{u}\mathbf{f}_{d,j}}$. Here the covariance matrix
$\mathbf{K}_{\mathbf{f}_{d,j}\mathbf{f}_{d,j}}\in \mathbb{R}^{N\times
	N}$ is built from the evaluation of all pairs of input data $\mathbf{X}$ in the covariance function
$\operatorname{cov}\left[f_{d,j}(\cdot),
f_{d,j}(\cdot)\right]=\sum_{q=1}^{Q} a_{d, j, q} a_{d, j, q}
k_{q}\left(\cdot, \cdot\right)$; and the cross covariance matrix
$\mathbf{K}_{\mathbf{f}_{d,j}\mathbf{u}}=[\mathbf{K}_{\mathbf{f}_{d,j}\mathbf{u}_1},...,\mathbf{K}_{\mathbf{f}_{d,j}\mathbf{u}_Q}]\in
\mathbb{R}^{N\times QM}$ is constructed with the blocks
$\mathbf{K}_{\mathbf{f}_{d,j}\mathbf{u}_q} \in \mathbb{R}^{N\times M}$, formed by the evaluations of
$\operatorname{cov}\left[f_{d,j}(\cdot), u_q(\cdot)\right]=a_{d, j, q}
k_{q}(\cdot, \cdot)$ between inputs $\mathbf{X}$ and $\mathbf{Z}_q$. Each kernel covariance $k_{q}\left(\cdot, \cdot\right)$ has an Exponentiated Quadratic (EQ) form as follows:
\begin{align}
\mathcal{E}(\boldsymbol{\tau}|\mathbf{0},\mathbf{L})=\frac{\left|\mathbf{L}\right|^{-1 / 2}}{(2 \pi)^{p / 2}} \exp \left[-\frac{1}{2}\boldsymbol{\tau}^{\top} \mathbf{L}^{-1}\boldsymbol{\tau}\right],
\label{eq:EQ_kern}
\end{align}
where $\boldsymbol{\tau}:=\mathbf{x}-\mathbf{x}^{\prime}$ and $\mathbf{L}$ is a diagonal matrix of length-scales. Thus, each $k_{q}\left(\mathbf{x}, \mathbf{x}^\prime\right) = \mathcal{E}(\boldsymbol{\tau}|\mathbf{0},\mathbf{L}_q)$.
\subsection{The Evidence Lower Bound}
We follow a VI derivation similar to the one used
for single output GPs \citep{Hensman2013,saul2016}. This approach allows the use of HetMOGP for large data. The goal is to approximate
the true posterior $p(\mathbf{f},\mathbf{u}|\mathbf{y})$ with a
variational distribution $q(\mathbf{f},\mathbf{u})$ by optimising the
following negative ELBO:
\begin{align}
\mathcal{\tilde{L}}=\sum_{n,d=1}^{N,D}\mathbb{E}_{q(\mathbf{f}_{d,1}) \cdots q(\mathbf{f}_{d,J_d})}\left[g_{d,n}\right]+\sum_{q=1}^{Q} \mathbb{D}_{KL}\left(\mathbf{u}_{q}\right),
\label{eq:HetMOGP}
\end{align}
where $g_{d,n}=-\log p(y_{d,n}|\psi_{d,1}(\mathbf{x}_n),...,\psi_{d,J_d}(\mathbf{x}_n))$ is the NLL function associated to each output, $\mathbb{D}_{KL}\left(\mathbf{u}_{q}\right):= \mathbb{D}_{KL}\left(q(\mathbf{u}_{q}) \| p(\mathbf{u}_{q})\right)$, and we have set a tractable posterior $q(\mathbf{f},\mathbf{u})=p(\mathbf{f}|\mathbf{u})q(\mathbf{u})$, where $p(\mathbf{f}|\mathbf{u})$ is already defined in Eq. \eqref{eq:poster_HetMOGP}, $q(\mathbf{u}|\mathbf{m},\mathbf{V})=\prod_{q=1}^{Q}q(\mathbf{u}_q)$, and each $q(\mathbf{u}_{q})=\mathcal{N}(\mathbf{u}_{q}|\mathbf{m}_q,\mathbf{V}_q)$ is a Gaussian distribution with mean $\mathbf{m}_q$ and covariance $\mathbf{V}_q$ \citep{Hensman2015} (see appendix C for details on the ELBO derivation). The above expectation associated to the NLL is computed using the marginal posteriors, 
\begin{align}
q(\mathbf{f}_{d,j}):=\mathcal{N}(\mathbf{f}_{d,j}|\mathbf{\tilde{m}}_{\mathbf{f}_{d,j}},\mathbf{\tilde{V}}_{\mathbf{f}_{d,j}}),
\label{eq:posterior}
\end{align}
with the following definitions, $\mathbf{\tilde{m}}_{\mathbf{f}_{d,j}}:=\mathbf{A}_{\mathbf{f}_{d,j}\mathbf{u}}\mathbf{m}$, $\mathbf{\tilde{V}}_{\mathbf{f}_{d,j}}:=\mathbf{K}_{\mathbf{f}_{d,j}\mathbf{f}_{d,j}}+\mathbf{A}_{\mathbf{f}_{d,j}\mathbf{u}}(\mathbf{V}-\mathbf{K_{uu}})\mathbf{A}_{\mathbf{f}_{d,j}\mathbf{u}}^{\top}$, where mean $\mathbf{m}=[\mathbf{m}^\top_1,...,\mathbf{m}^\top_Q]^\top \in \mathbb{R}^{QM\times 1}$ and the covariance matrix $\mathbf{V} \in \mathbb{R}^{QM\times QM}$ is a block-diagonal matrix with blocks given by $\mathbf{V}_q \in \mathbb{R}^{M\times M}$.\footnote{Each marginal posterior derives from: $q(\mathbf{f}_{d,j})=\int p(\mathbf{f}_{d,j}|\mathbf{u})q(\mathbf{u})d\mathbf{u}$.} The objective function derived in Eq. \eqref{eq:HetMOGP} for the HetMOGP model with LMC requires fitting the parameters of each posterior $q(\mathbf{u}_q)$, the inducing points $\mathbf{Z}$, the kernel hyper-parameters  $\mathbf{L}_\text{kern}=[\mathbf{L}_1^\top,...,\mathbf{L}_Q^\top]^\top$ and the coefficients $\mathbf{w}$. With the aim to fit said variables in a FNG scheme, later on we will apply the VO perspective on Eq. \eqref{eq:HetMOGP} for inducing randomness and gain exploration over $\mathbf{Z}$, $\mathbf{L}_{\text{kern}}$ and $\mathbf{w}$; and by means of the MDA we will derive the inference updates for all the model's variables.
%\subsection{Predictive Distribution for the HetMOGP with LMC}
%In order to make predictions with the model, it is necessary to compute the following distribution:
%\begin{align}
%p(\mathbf{y}_*|\mathbf{y})\approx \int p(\mathbf{y}_*|\mathbf{f}_*)q(\mathbf{f}_*)d\mathbf{f}_*,
%\label{eq:predict_HetMOGP}
%\end{align}
%where $q(\mathbf{f}_*)=\prod_{d=1}^{D}\prod_{j=1}^{J_d}q(\mathbf{f}_{d,j,*})$, and each $q(\mathbf{f}_{d,j,*})$ can be computed with Eq. \eqref{eq:posterior} by evaluating $\mathbf{K}_{\mathbf{f}_{d,j,*}\mathbf{u}}$ and $\mathbf{K}_{\mathbf{f}_{d,j,*}\mathbf{f}_{d,j,*}}$ at the new inputs $\mathbf{X}_*$. 

\section{Heterogeneous Multi-Output GPs with Convolution Processes}\label{sec:CPM}
The HetMOGP model with convolution processes follows the same
likelihood defined in Eq. \eqref{eq:likelihood}, though each
$f_{d,j}(\mathbf{x}_n)$ is considered a LPF that comes from a
convolution process as follows:
\begin{align}
f_{d,j}\left(\mathbf{x}\right)=\sum_{q=1}^{Q} \sum_{i=1}^{R_q}\int_{\mathcal{X}} G^{i}_{d,j, q}\left(\mathbf{x}-\mathbf{r}^{\prime}\right)u^{i}_{q}\left(\mathbf{r}^{\prime}\right) \mathrm{d} \mathbf{r}^{\prime},\notag
\end{align}
where $u^i_q(\mathbf{x})$ are IID samples from Gaussian Processes
$u_q(\cdot)\sim\mathcal{GP}(0,k_q(\cdot,\cdot))$ and each $G_{d,j, q}(\cdot)$
represents a smoothing kernel. We will also use $R_q = 1$ as in the LMC for
simplicity in the following derivations \citep{Boyle2005}. 
\subsection{The Inducing Points Method}\label{sec:inducing_CPM}
With the purpose to reduce the
computational complexities involved in GPs we follow
the inducing variables framework by augmenting the
probability space as,
\begin{align}
p(\mathbf{f}|\mathbf{\check{u}})p(\mathbf{\check{u}})=\prod_{d=1}^{D}\prod_{j=1}^{J_d}p(\mathbf{f}_{d,j}|\mathbf{\check{u}}_{d,j})p(\mathbf{\check{u}}_{d,j}),
\label{eq:inducing_v2}
\end{align}
with $p(\mathbf{\check{u}})=\prod_{d=1}^{D}\prod_{j=1}^{J_d}p(\mathbf{\check{u}}_{d,j})$, and $p(\mathbf{f}|\mathbf{\check{u}})=\prod_{d=1}^{D}\prod_{j=1}^{J_d}p(\mathbf{f}_{d,j}|\mathbf{\check{u}}_{d,j})$, where the vector	$\mathbf{\check{u}}=[\mathbf{\check{u}}^\top_{1,1},...,\mathbf{\check{u}}^\top_{1,J_1},...,\mathbf{\check{u}}^\top_{D,J_D}]^\top \in \mathbb{R}^{JM\times 1}$ is built from the inducing variables $\mathbf{\check{u}}_{d,j}=[f_{d,j}(\mathbf{z}^{(1)}_{d,j}),...,f_{d,j}(\mathbf{z}^{(M)}_{d,j})]^\top \in \mathbb{R}^{M\times 1}$. As it can be seen, these inducing variables are additional evaluations of the functions $f_{d,j}(\cdot)$ at each set of inducing points $\mathbf{Z}_{d,j}=[\mathbf{z}^{(1)}_{d,j},...,\mathbf{z}^{(M)}_{d,j}]^\top \in \mathbb{R}^{M\times P}$, thus the set of all inducing variables is $\mathbf{Z}=[\text{vec}(\mathbf{Z}_{1,1})^\top,...,\text{vec}(\mathbf{Z}_{1,J_1})^\top,...,\text{vec}(\mathbf{Z}_{D,J_D})^\top]^\top \in \mathbb{R}^{JMP\times 1}$. Using the properties of Gaussian distributions we can express, $p(\mathbf{f}_{d,j}|\mathbf{\check{u}}_{d,j})=\mathcal{N}(\mathbf{f}_{d,j}|\mathbf{A}_{\mathbf{f}_{d,j}\mathbf{\check{u}}_{d,j}}\mathbf{\check{u}}_{d,j},\bar{\mathbf{Q}}_{\mathbf{f}_{d,j}})$, $p(\mathbf{\check{u}}_{d,j})=\mathcal{N}(\mathbf{\check{u}}_{d,j}|\mathbf{0},\mathbf{K}_{\mathbf{\check{u}}_{d,j}})$, with the following definitions:
$
\mathbf{A}_{\mathbf{f}_{d,j}\mathbf{\check{u}}_{d,j}}=\mathbf{K}_{\mathbf{f}_{d,j}\mathbf{\check{u}}_{d,j}}\mathbf{K}^{-1}_{\mathbf{\check{u}}_{d,j}}$, $\bar{\mathbf{Q}}_{\mathbf{f}_{d,j}} =\mathbf{K}_{\mathbf{f}_{d,j}\mathbf{f}_{d,j}}-\mathbf{\check{Q}}_{\mathbf{f}_{d,j}}$, $\mathbf{\check{Q}}_{\mathbf{f}_{d,j}}=\mathbf{K}_{\mathbf{f}_{d,j}\mathbf{\check{u}}_{d,j}}\mathbf{K}^{-1}_{\mathbf{\check{u}}_{d,j}}\mathbf{K}_{\mathbf{\check{u}}_{d,j}\mathbf{f}_{d,j}}$, $\mathbf{K}_{\mathbf{f}_{d,j}\mathbf{\check{u}}_{d,j}}=\mathbf{K}^{\top}_{\mathbf{\check{u}}_{d,j}\mathbf{f}_{d,j}}
$. Here the covariance matrix $\mathbf{K}_{\mathbf{f}_{d,j}\mathbf{f}_{d,j}}\in \mathbb{R}^{N\times N}$ is built from the evaluation of all pairs of input data $\mathbf{X}\in \mathbb{R}^{N\times P}$ in the covariance function $$\text{cov}\left[f_{d,j}\left(\mathbf{x}\right) f_{d^{\prime},j^{\prime}}\left(\mathbf{x}^{\prime}\right)\right]
=\sum_{q=1}^{Q} \int_{\mathcal{X}} G_{d,j, q}\left(\mathbf{x}-\mathbf{r}\right) \int_{\mathcal{X}} G_{d^{\prime},j^{\prime}, q}\left(\mathbf{x}^{\prime}-\mathbf{r}^{\prime}\right) k_q(\mathbf{r},\mathbf{r}^{\prime}) \mathrm{d} \mathbf{r} \mathrm{d} \mathbf{r}^{\prime},$$ the cross covariance matrix $\mathbf{K}_{\mathbf{f}_{d,j}\mathbf{\check{u}}_{d,j}}\in \mathbb{R}^{N\times M}$ is formed by evaluations of the equation above between inputs $\mathbf{X}$ and $\mathbf{Z}_{d,j}$, and the matrix $\mathbf{K}_{\mathbf{\check{u}}_{d,j}}\in \mathbb{R}^{M\times M}$ is also built from evaluations of the equation above between all pairs of inducing points $\mathbf{Z}_{d,j}$ respectively. We can compute the above
covariance function analytically for certain forms of $G_{d,j, q}\left(\cdot\right)$ and $ k_q(\mathbf{r},\mathbf{r}^{\prime})$. In this paper, we
follow the work in \citep{Alvarez2011} by defining the kernels in the EQ form of Eq. \eqref{eq:EQ_kern}: $k_{q}\left(\mathbf{x},\mathbf{x}^{\prime}\right)= \mathcal{E}(\boldsymbol{\tau}|\mathbf{0},\mathbf{L}_q)$ and $G_{d,j,q}(\boldsymbol{\tau})=S_{d,j,q}\mathcal{E}(\boldsymbol{\tau}|\mathbf{0},\boldsymbol{\kappa}_{d,j})$, where $S_{d,j,q}$ is a weight associated to the LPF indexed by $f_{d,j}(\cdot)$ and to the latent function $u_q(\cdot)$, and $\boldsymbol{\kappa}_{d,j}$ is a diagonal covariance matrix particularly associated to each $f_{d,j}(\cdot)$, $\boldsymbol{\kappa}_{d,j}$ can be seen as a matrix of length-scales in its diagonal. Therefore, when solving for the $\text{cov}\left[f_{d,j}\left(\mathbf{x}\right) f_{d^{\prime},j^{\prime}}\left(\mathbf{x}^{\prime}\right)\right]$ above we end up with the closed-form,
\begin{align}
k_{f_{d,j}, f_{d^{\prime},j^{\prime}}}\left(\boldsymbol{\tau}\right)=\sum_{q=1}^{Q}S_{d,j, q} S_{d^{\prime},j^{\prime}, q} \mathcal{E}(\boldsymbol{\tau}|\mathbf{0},\mathbf{P}_{d,j,d^{\prime},j^{\prime},q})\label{eq:smooth_kern},
\end{align}
where $\mathbf{P}_{d,j,d^{\prime},j^{\prime},q}$ represents a diagonal matrix of length-scales, $\mathbf{P}_{d,j,d^{\prime},j^{\prime},q}=\boldsymbol{\kappa}_{d,j}+\boldsymbol{\kappa}_{d^{\prime},j^{\prime}}+\mathbf{L}_{q}.$
\subsection{The Evidence Lower Bound}
We now introduce the negative ELBO for the HetMOGP that uses convolution processes. It follows as
\begin{align}
\mathcal{\tilde{L}}=\sum_{n,d=1}^{N,D}&\mathbb{E}_{q(\mathbf{f}_{d,1}) \cdots q(\mathbf{f}_{d,J_d})}\left[g_{d,n}\right]+\sum_{d,j=1}^{D,J_d} \mathbb{D}_{KL}\left(\mathbf{\check{u}}_{d,j}\right),\label{eq:convHetMOGP}
\end{align}
where $g_{d,n}=-\log p(y_{d,n}|\psi_{d,1}(\mathbf{x}_n),...,\psi_{d,J_d}(\mathbf{x}_n))$ is the NLL function associated to each output, $\mathbb{D}_{KL}\left(\mathbf{\check{u}}_{d,j}\right) :=\mathbb{D}_{KL}\left(q(\mathbf{\check{u}}_{d,j}) \| p(\mathbf{\check{u}}_{d,j})\right)$, and we have set a tractable posterior $q(\mathbf{f},\mathbf{\check{u}})=p(\mathbf{f}|\mathbf{\check{u}})q(\mathbf{\check{u}})$, where $p(\mathbf{f}|\mathbf{\check{u}})$ is already defined in Eq. \eqref{eq:inducing_v2}, $q(\mathbf{\check{u}}|\mathbf{m},\mathbf{V})=\prod_{d=1}^{D}\prod_{j=1}^{J_d}q(\mathbf{\check{u}}_{d,j})$, and each $q(\mathbf{\check{u}}_{d,j})=\mathcal{N}(\mathbf{\check{u}}_{d,j}|\mathbf{m}_{d,j},\mathbf{V}_{d,j})$ is a Gaussian distribution with mean $\mathbf{m}_{d,j}\in \mathbb{R}^{M\times 1}$ and covariance $\mathbf{V}_{d,j}\in \mathbb{R}^{M\times M}$ (see appendix D for details on the ELBO derivation) \citep{Hensman2013,pablo2018}. The above expectation is computed w.r.t the marginals,  
\begin{align}
q(\mathbf{f}_{d,j})=\int p(\mathbf{f}_{d,j}|\mathbf{\check{u}}_{d,j})q(\mathbf{\check{u}}_{d,j})d\mathbf{\check{u}}_{d,j}=\mathcal{N}(\mathbf{f}_{d,j}|\mathbf{\tilde{m}}_{\mathbf{f}_{d,j}},\mathbf{\tilde{V}}_{\mathbf{f}_{d,j}}),
\label{eq:convposterior}
\end{align}
with the following definitions, $\mathbf{\tilde{m}}_{\mathbf{f}_{d,j}}:=\mathbf{A}_{\mathbf{f}_{d,j}\mathbf{\check{u}}_{d,j}}\mathbf{m}_{d,j}$, $\mathbf{\tilde{V}}_{\mathbf{f}_{d,j}}:=\mathbf{K}_{\mathbf{f}_{d,j}\mathbf{f}_{d,j}}+\mathbf{A}_{\mathbf{f}_{d,j}\mathbf{\check{u}}_{d,j}}(\mathbf{V}_{d,j}-\mathbf{K}_{\mathbf{\check{u}}_{d,j}})\mathbf{A}_{\mathbf{f}_{d,j}\mathbf{\check{u}}_{d,j}}^{\top}$. The objective derived in Eq. \eqref{eq:convHetMOGP} for the HetMOGP model with convolution processes requires fitting the parameters of each posterior $q(\mathbf{\check{u}}_{d,j})$, the inducing points $\mathbf{Z}$, the kernel hyper-parameters $\mathbf{L}_\text{kern}$, the smoothing-kernels' length-scales $\boldsymbol{\kappa}_{\text{smooth}}=[\boldsymbol{\kappa}_{1,1}^\top,...,\boldsymbol{\kappa}_{1,J_1}^\top,...,\boldsymbol{\kappa}_{D,J_D}^\top]^\top
\in \mathbb{R}_{+}^{JP\times 1}$ and the weights $\mathbf{S}_q=[S_{1, 1, q},...,S_{1, J_1, q},...,S_{D, J_D, q}]^\top \in \mathbb{R}^{J\times 1}$ associated to each smoothing-kernel. In the interest of fitting those variables in a FNG scheme, in the following section we will explain how to apply the VO perspective over Eq. \eqref{eq:convHetMOGP} so as to introduce stochasticity over $\mathbf{Z}$, $\mathbf{L}_{\text{kern}}$, $\boldsymbol{\kappa}_{\text{smooth}}$ and $\mathbf{S}_q$; and through the MDA we will derive closed-form updates for all parameters of the model.
%\subsection{Predictive Distribution for the HetMOGP with CPM}
%The predictive distribution for the HetMOGP with CPM is estimated using Eq. \eqref{eq:predict_HetMOGP}, though unlike the LMC, for this CPM each $q(\mathbf{f}_{d,j,*})$ has to be computed using Eq. \eqref{eq:convposterior} by evaluating $\mathbf{K}_{\mathbf{f}_{d,j,*}\mathbf{\check{u}}_{d,j}}$ and $\mathbf{K}_{\mathbf{f}_{d,j,*}\mathbf{f}_{d,j,*}}$ at the new inputs $\mathbf{X}_*$. 

\section{Deriving a Fully Natural Gradient Scheme}
This section describes how to derive the FNG updates for optimising both the LMC and CPM schemes of the HetMOGP model. We first detail how to induce an exploratory distribution over the hyper-parameters and inducing points, then we write down the MDA for the model and derive the update equations. Later on, we get into specific details about the algorithm's implementation. 
\subsection{An Exploratory Distribution for HetMOGP with LMC}
In the context of sparse GPs, the kernel hyper-parameters and inducing points of the model have usually been treated as deterministic variables. Here, we use the VO perspective as a mechanism to induce randomness over such variables, this with the aim to gain exploration for finding better solutions during the inference process. To this end we define and connect random real vectors to the variables through a link function $\phi(\cdot)$ as follows: for the inducing points
$\mathbf{Z}=\boldsymbol{\theta}_\mathbf{Z}$, for the
kernel hyper-parameters
$\mathbf{L}_\text{kern}=\exp(\boldsymbol{\theta}_{\mathbf{L}})$ with
$\boldsymbol{\theta}_{\mathbf{L}}=[\boldsymbol{\theta}_{\mathbf{L}_1}^\top,...,\boldsymbol{\theta}_{\mathbf{L}_Q}^\top]^\top \in \mathbb{R}^{QP\times 1¸}$, 
and for the vector of LCC $\mathbf{w}=\boldsymbol{\theta}_{\mathbf{w}}$, that are used to generate
the LPFs in Eq. \eqref{eq:linear_comb2}. We have defined the real random vectors
$\boldsymbol{\theta}_\mathbf{Z}\in \mathbb{R}^{QMP\times 1}$,
$\boldsymbol{\theta}_{\mathbf{L}_q} \in \mathbb{R}^{P\times1}$ and
$\boldsymbol{\theta}_{\mathbf{w}} \in \mathbb{R}^{QJ\times 1}$ to plug
the set of inducing points, the kernel hyper-parameters
per latent function $u_q(\cdot)$, and the vector $\mathbf{w}$ of
LCCs. We cluster the random vectors defining
$\boldsymbol{\theta}=[\boldsymbol{\theta}_\mathbf{Z}^\top,\boldsymbol{\theta}_{\mathbf{L}}^\top,\boldsymbol{\theta}^\top_{\mathbf{w}}]^\top
\in \mathbb{R}^{(QMP+QP+QJ)\times 1}$ to refer to all the parameters
in a single variable. Hence, we can specify an exploratory
distribution
$q(\boldsymbol{\theta}):=\mathcal{N}(\boldsymbol{\theta}|\boldsymbol{\mu},\boldsymbol{\Sigma})$
for applying the VO approach in Eq. \eqref{eq:VO}, though for our case
the objective to bound is $\mathcal{\tilde{L}}$,
already derived in Eq. \eqref{eq:HetMOGP} for the HetMOGP with a
LMC. Therefore our VO bound is defined as follows:
\begin{align}
\tilde{\mathcal{F}}&= \mathbb{E}_{q(\boldsymbol{\theta})}\big[\tilde{\mathcal{L}}\big]+\mathbb{D}_{KL}(q(\boldsymbol{\theta})||p(\boldsymbol{\theta})),\label{eq:new_ELBO}	
\end{align} 
where $p(\boldsymbol{\theta})=\mathcal{N}(\boldsymbol{\theta}|\mathbf{0},\lambda_1^{-1}\mathbf{I})$ is a Gaussian distribution with precision $\lambda_1$ that forces further exploration of $\boldsymbol{\theta}$'s space \citep{Khan2017Vprop}. 

\subsection{An Exploratory Distribution for the HetMOGP with CPM}
The case of the CPM has the same the kernel hyper-parameters
$\mathbf{L}_\text{kern}=\exp(\boldsymbol{\theta}_{\mathbf{L}})$ and inducing points
$\mathbf{Z}=\boldsymbol{\theta}_\mathbf{Z}$, but differs from the LMC one since the smoothing
kernels involve a new set of hyper-parameters, the smoothing-kernels'
length-scales. The way we define and connect the new random real vectors is as follows:
$\boldsymbol{\kappa}_{\text{smooth}}=\exp(\boldsymbol{\theta}_{\boldsymbol{\kappa}})$,
with
$\boldsymbol{\theta}_{\boldsymbol{\kappa}}=[\boldsymbol{\theta}_{\boldsymbol{\kappa}_{1,1}}^\top,...,\boldsymbol{\theta}_{\boldsymbol{\kappa}_{1,J_1}}^\top,...,\boldsymbol{\theta}_{\boldsymbol{\kappa}_{D,J_D}}^\top]^\top
\in \mathbb{R}^{JP\times 1}$, where
$\boldsymbol{\theta}_{\boldsymbol{\kappa}_{d,j}} \in
\mathbb{R}^{P\times1}$ is a real random vector associated to each
smoothing kernel $G_{d,j,q}(\cdot)$ from Eq. \eqref{eq:smooth_kern}. Also,
instead of the combination coefficients $\mathbf{w}$ of the LMC, for
the CPM we have an analogous set of weights from the
smoothing-kernels in Eq. \eqref{eq:smooth_kern},
$\mathbf{S}=\boldsymbol{\theta}_{\mathbf{S}}$, where
$\mathbf{S} = [\mathbf{S}_1^\top,...,\mathbf{S}_Q^\top]^\top \in
\mathbb{R}^{QJ\times 1}$ is a vector that groups all the weights that belong to the smoothing kernels. Thus, the real random vectors for the CPM are:  
$\boldsymbol{\theta}_\mathbf{Z}\in \mathbb{R}^{JMP\times 1}$,
$\boldsymbol{\theta}_{\mathbf{L}} \in \mathbb{R}^{QP\times1}$,
$\boldsymbol{\theta}_{\boldsymbol{\kappa}} \in
\mathbb{R}^{JP\times1}$, and
$\boldsymbol{\theta}_{\mathbf{S}} \in \mathbb{R}^{QJ\times 1}$. We group the random vectors by defining $\boldsymbol{\theta}=[\boldsymbol{\theta}_\mathbf{Z}^\top,\boldsymbol{\theta}_{\mathbf{L}}^\top,\boldsymbol{\theta}_{\boldsymbol{\kappa}}^\top,\boldsymbol{\theta}^\top_{\mathbf{S}}]^\top
\in \mathbb{R}^{(JMP+QP+JP+QJ)\times 1}$. Notice that, for the CPM, the dimensionality of the real random vector $\boldsymbol{\theta}$ differs from the one for LMC, this is due to the way the inducing variables are treated in subsection \ref{sec:inducing_CPM} and the additional set of smoothing-kernel's hyper-parameters. In the same way as defined for the LMC, we specify an
exploratory distribution
$q(\boldsymbol{\theta}):=\mathcal{N}(\boldsymbol{\theta}|\boldsymbol{\mu},\boldsymbol{\Sigma})$
and follow the VO approach in Eq. \eqref{eq:VO}. In this case the
objective to bound is the one derived for the CPM, i.e., the new
bound, $\tilde{\mathcal{F}}$, is exactly the same as
Eq. \eqref{eq:new_ELBO}, but using the corresponding
$\mathcal{\tilde{L}}$ from Eq. \eqref{eq:convHetMOGP}.

\subsection{Mirror Descent Algorithm for the HetMOGP with LMC}
With the purpose of minimising our VO objective in Eq. \eqref{eq:new_ELBO}, we use the MDA in Eq. \eqref{eq:mirror_mom} which additionally exploits the natural-momentum. In the interest of easing the derivation, we use the mean-parameters of distributions $q(\mathbf{u}_q)$ and $q(\boldsymbol{\theta})$ defining $\boldsymbol{\rho}_q=\{\mathbf{m}_q,\mathbf{m}_q\mathbf{m}_q^\top+\mathbf{V}_q\}$ and $\boldsymbol{\eta}=\{\boldsymbol{\mu},\boldsymbol{\mu}\boldsymbol{\mu}^\top+\boldsymbol{\Sigma}\}$. In this way we can write the MDA as: 
\begin{align}
\boldsymbol{\eta}_{t+1},\{\boldsymbol{\rho}_{q,t+1}\}^Q_{q=1}&=\begin{matrix}\arg \min\\{\boldsymbol{\eta},\{\boldsymbol{\rho}_q\}^Q_{q=1}} \end{matrix} \langle\boldsymbol{\eta},\hat{\nabla}_{\boldsymbol{\eta}}\tilde{\mathcal{F}}_t\rangle+ \frac{1}{\tilde{\alpha}_t}\text{KL}(\boldsymbol{\theta})_t-\frac{\tilde{\gamma}_t}{\tilde{\alpha}_t}\text{KL}(\boldsymbol{\theta})_{t-1}\label{eq:mirror_HetMOGP}\\&+\sum_{q=1}^{Q}\Big[\langle\boldsymbol{\rho}_q,\hat{\nabla}_{\boldsymbol{\rho}_q}\tilde{\mathcal{F}}_t\rangle+\frac{1}{\tilde{\beta}_t}\text{KL}(\mathbf{u}_q)_t\notag-\frac{\tilde{\upsilon}_t}{\tilde{\beta}_t}\text{KL}(\mathbf{u}_{q})_{t-1}\Big],\notag
\end{align}
where $\tilde{\mathcal{F}}_t:=\tilde{\mathcal{F}}(\mathbf{m}_{t},\mathbf{V}_{t},\boldsymbol{\mu}_t,\boldsymbol{\Sigma}_{t})$ and $\tilde{\beta}_t$, $\tilde{\alpha}_t$, $\tilde{\upsilon}_t$, and $\tilde{\gamma}_t$ are positive step-size parameters.
\subsection{Mirror Descent Algorithm for the HetMOGP with CPM}
For the HetMOGP with CPM we follow a similar procedure carried out for the LMC. We use the MDA in Eq. \eqref{eq:mirror_mom} and the mean-parameters of distributions $q(\mathbf{\check{u}}_{d,j})$ and $q(\boldsymbol{\theta})$ defining $\boldsymbol{\rho}_{d,j}=\{\mathbf{m}_{d,j},\mathbf{m}_{d,j}\mathbf{m}_{d,j}^\top+\mathbf{V}_{d,j}\}$ and $\boldsymbol{\eta}=\{\boldsymbol{\mu},\boldsymbol{\mu}\boldsymbol{\mu}^\top+\boldsymbol{\Sigma}\}$ for minimising Eq. \eqref{eq:new_ELBO}. Then, our algorithm for the CPM can be written as: 
\begin{align}
\boldsymbol{\eta}_{t+1},\{\boldsymbol{\rho}_{{d,j},t+1}\}^{D,J_d}_{d=1,j=1}&=\begin{matrix}\arg \min\\{\boldsymbol{\eta},\{\boldsymbol{\rho}_{d,j}\}^{D,J_d}_{d=1,j=1}} \end{matrix}  \langle\boldsymbol{\eta},\hat{\nabla}_{\boldsymbol{\eta}}\tilde{\mathcal{F}}_t\rangle+\frac{1}{\tilde{\alpha}_t}\text{KL}(\boldsymbol{\theta})_t-\frac{\tilde{\gamma}_t}{\tilde{\alpha}_t}\text{KL}(\boldsymbol{\theta})_{t-1}\label{eq:mirror_convHetMOGP}\\&+\sum_{d,j=1}^{D,J_d}\Big[\langle\boldsymbol{\rho}_{d,j},\hat{\nabla}_{\boldsymbol{\rho}_{d,j}}\tilde{\mathcal{F}}_t\rangle+\frac{1}{\tilde{\beta}_t}\text{KL}(\mathbf{\check{u}}_{d,j})_t\notag-\frac{\tilde{\upsilon}_t}{\tilde{\beta}_t}\text{KL}(\mathbf{\check{u}}_{d,j})_{t-1}\Big],\notag
\end{align}
where we have used the same variables $\tilde{\beta}_t$, $\tilde{\alpha}_t$, $\tilde{\upsilon}_t$, and $\tilde{\gamma}_t$ for the step-size parameters as in the LMC. This for the sake of a unified derivation of the FNG updates in the next subsection.
\subsection{Fully Natural Gradient Updates}
We can solve for Eq. \eqref{eq:mirror_HetMOGP} and \eqref{eq:mirror_convHetMOGP} by computing derivatives w.r.t $\boldsymbol{\eta}$ and $\boldsymbol{\rho}$, and setting to zero. This way we obtain results similar to Eq. \eqref{eq:update_nat1} and \eqref{eq:update_nat2}, we call them FNG updates:
\begin{align}
\boldsymbol{\Sigma}_{t+1}^{-1}&=\boldsymbol{\Sigma}_{t}^{-1}+2\alpha_t\hat{\nabla}_{\boldsymbol{\Sigma}}\tilde{\mathcal{F}}_t\label{eq:sigma_update}\\
\boldsymbol{\mu}_{t+1}&=\boldsymbol{\mu}_{t}-\alpha_t\boldsymbol{\Sigma}_{t+1}\hat{\nabla}_{\boldsymbol{\mu}}\tilde{\mathcal{F}}_t+\gamma_t\boldsymbol{\Sigma}_{t+1}\boldsymbol{\Sigma}^{-1}_{t}(\boldsymbol{\mu}_t-\boldsymbol{\mu}_{t-1})\label{eq:mu_update}\\
\mathbf{V}_{(\cdot),t+1}^{-1}&=\mathbf{V}_{(\cdot),t}^{-1}+2\beta_t\hat{\nabla}_{\mathbf{V}_{(\cdot)}}\tilde{\mathcal{F}}_t\label{eq:Vdj_update}\\
\mathbf{m}_{(\cdot),t+1}&=\mathbf{m}_{(\cdot),t}-\beta_t\mathbf{V}_{(\cdot),t+1}\hat{\nabla}_{\mathbf{m}_{(\cdot)}}\tilde{\mathcal{F}}_t \label{eq:mdj_update}\\&\quad\quad\quad\quad+\upsilon_t\mathbf{V}_{(\cdot),t+1}\mathbf{V}^{-1}_{(\cdot),t}(\mathbf{m}_{(\cdot),t}-\mathbf{m}_{(\cdot),t-1}),\notag
\end{align}
where we have defined, $\mathbf{m}_{(\cdot),t}$, as a way of referring to either $\mathbf{m}_{q,t}$ or $\mathbf{m}_{{d,j},t}$ depending on the case of LMC or CPM. This also applies for $\mathbf{V}_{(\cdot),t}$ without loss of generality. And $\alpha_t=\tilde{\alpha}_t/(1-\tilde{\gamma}_t)$, $\beta_t=\tilde{\beta}_t/(1-\tilde{\upsilon}_t)$, $\gamma_t=\tilde{\gamma}_t/(1-\tilde{\gamma}_t)$ and $\upsilon_t=\tilde{\upsilon}_t/(1-\tilde{\upsilon}_t)$  are positive step-size  parameters (see appendix E for details on the gradients derivation).
%Notice that if we assume that $p(\boldsymbol{\theta})=q(\boldsymbol{\theta})$, then the KL divergence to the right hand side of \eqref{eq:new_ELBO} vanishes \citep{Hensman2015b,Khan2017VAN}. So, the above gradients end up with only the first expectation term to the right hand side. This means that, when the optimisation is carried out, the covariance $\mathbf{\boldsymbol{\Sigma}}$ collapses rapidly ($\mathbf{\boldsymbol{\Sigma}}\rightarrow \mathbf{0}$) without much exploration around the mean $\boldsymbol{\mu}$.

\subsection{Implementation}

In order to implement the proposed method, we have to take into account that our computational complexity depends on inverting the covariance matrix $\boldsymbol{\Sigma}$ in Eq. \eqref{eq:sigma_update}. Such complexity can be expressed as $\mathcal{O}((QMP+QP+QJ)^3)$ for the LMC, or $\mathcal{O}((JMP+QP+JP+QJ)^3)$ for the CPM, where the terms with the number of inducing points and/or input dimensionality tend to dominate the complexity in both cases. Likewise, the gradient $\hat{\nabla}_{\boldsymbol{\Sigma}}\tilde{\mathcal{F}}$ involves computing the Hessian $\hat{\nabla}^{2}_{\boldsymbol{\theta}\boldsymbol{\theta}}\tilde{\mathcal{L}}$ which can be computationally expensive and prone to suffer from non-positive definiteness. To alleviate those complexity issues we assume $\boldsymbol{\Sigma}=\text{diag}(\boldsymbol{\sigma}^2)$, where $\boldsymbol{\sigma}$ is a vector of standard deviations, and $\text{diag}(\boldsymbol{\sigma}^2)$ represents a matrix with the elements of $\boldsymbol{\sigma}^2$ on its diagonal. Additionally, we estimate the Hessian by means of the Gauss-Newton (GN) approximation $\hat{\nabla}^{2}_{\boldsymbol{\theta}\boldsymbol{\theta}}\tilde{\mathcal{L}}\approx \hat{\nabla}_{\boldsymbol{\theta}}\tilde{\mathcal{L}} \circ \hat{\nabla}_{\boldsymbol{\theta}}\tilde{\mathcal{L}}$ \citep{Bertsekas99,Khan2017Vprop}. The authors in \citep{Khan2018} term this method as the variational RMSprop with momentum. They alternatively express Eq. \eqref{eq:sigma_update} and \eqref{eq:mu_update} as: 
\begin{align} 
\mathbf{p}_{t+1} &=\left(1-\alpha_{t}\right) \mathbf{p}_{t}+\alpha_{t}\mathbb{E}_{q(\boldsymbol{\theta})}\big[\hat{\nabla}_{\boldsymbol{\theta}}\tilde{\mathcal{L}}\circ \hat{\nabla}_{\boldsymbol{\theta}}\tilde{\mathcal{L}}\big]\label{eq:p_vprop}\\
\boldsymbol{\mu}_{t+1} &=\boldsymbol{\mu}_{t}-\alpha_{t}(\mathbf{p}_{t+1}+{\lambda_1\mathbf{1}})^{-1}\circ\hat{\nabla}_{\boldsymbol{\mu}}\tilde{\mathcal{F}}\label{eq:mu_vprop}\\&\quad+\gamma_t(\mathbf{p}_{t}+{\lambda_1\mathbf{1}})\circ (\mathbf{p}_{t+1}+{\lambda_1\mathbf{1}})^{-1}\circ\left(\boldsymbol{\mu}_{t}-\boldsymbol{\mu}_{t-1}\right),\notag 
\end{align}
where $\hat{\nabla}_{\boldsymbol{\mu}}\tilde{\mathcal{F}}=(\mathbb{E}_{q(\boldsymbol{\theta})}\big[\hat{\nabla}_{\boldsymbol{\theta}}\tilde{\mathcal{L}}\big]+{\lambda_1}\boldsymbol{\mu}_{t})$, $\circ$ represents an element-wise product and we have made a
variable change defining a vector $\mathbf{p}_{t}:=\boldsymbol{\sigma}_{t}^{-2}-\lambda_1 \mathbf{1}$,
with $\mathbf{1}$ as a vector of ones. The GN approximation provides
stronger numerical stability by preventing that
$\boldsymbol{\sigma}^2$ becomes negative. Also, using
$\text{diag}(\boldsymbol{\sigma}^2)$ we reduce the computational
complexity from $\mathcal{O}((QMP+QP+QJ)^3)$ to
$\mathcal{O}(QMP+QP+QJ)$ for the LMC, or
$\mathcal{O}((JMP+QP+JP+QJ)^3)$ to $\mathcal{O}(JMP+QP+JP+QJ)$ for the
CPM (see appendix F for a pseudo-code algorithm implementation). 
\subsection{Predictive Distribution}
In order to make predictions with the HetMOGP model, it is necessary to compute the following distribution: $p(\mathbf{y}_*|\mathbf{y})\approx \int p(\mathbf{y}_*|\mathbf{f}_*)q(\mathbf{f}_*)d\mathbf{f}_*,\notag$ 
where $q(\mathbf{f}_*)=\prod_{d=1}^{D}\prod_{j=1}^{J_d}q(\mathbf{f}_{d,j,*})$. Given that we have introduced a variational distribution $q(\boldsymbol{\theta})$ over all hyper-parameters and inducing points of the model, we could apply a fully Bayesian treatment when calculating $q(\mathbf{f}_{d,j,*})$, either for the LMC $q(\mathbf{f}_{d,j,*})=\int p(\mathbf{f}_{d,j,*}|\mathbf{u},\boldsymbol{\theta})q(\mathbf{u})q(\boldsymbol{\theta})d\boldsymbol{\theta}d\mathbf{u},$; or the CPM $q(\mathbf{f}_{d,j,*})=\int p(\mathbf{f}_{d,j,*}|\mathbf{\check{u}},\boldsymbol{\theta})q(\mathbf{\check{u}})q(\boldsymbol{\theta})d\boldsymbol{\theta}d\mathbf{\check{u}}$.
In practice, we found that $q(\boldsymbol{\theta})$'s covariance
converged to very small values, in general
$\text{diag}(\boldsymbol{\sigma}^2)\le 10^{-15}$, and almost all the
uncertainty information was concentrated on $q(\mathbf{u})$'s
covariance for LMC, or $q(\mathbf{\check{u}})$'s covariance for
CPM. Since making predictions with the equations above becomes
computationally expensive and most of the uncertainty is represented by
the distribution $q(\mathbf{u})$ or $q(\mathbf{\check{u}})$, we can
trade-off the computation by using the MAP solution for
$q(\boldsymbol{\theta})$ and completely integrating over the remaining
distribution as follows: for LMC $q(\mathbf{f}_{d,j,*})=\int p(\mathbf{f}_{d,j,*}|\mathbf{u},\boldsymbol{\theta}=\boldsymbol{\mu})q(\mathbf{u})d\mathbf{u}$, and for CPM $q(\mathbf{f}_{d,j,*})=\int p(\mathbf{f}_{d,j,*}|\mathbf{\check{u}},\boldsymbol{\theta}=\boldsymbol{\mu})q(\mathbf{\check{u}})d\mathbf{\check{u}}$. When solving these integrals we arrive to exactly the same solutions in Eq. \eqref{eq:posterior} if we aim to make predictions for the LMC, or Eq. \eqref{eq:convposterior} if the case for CPM, where we simply have to evaluate the matrix covariances $\mathbf{K}_{\mathbf{f}_{d,j,*}\mathbf{u}}$ for LMC or $\mathbf{K}_{\mathbf{f}_{d,j,*}\mathbf{\check{u}}}$ for CPM, and $\mathbf{K}_{\mathbf{f}_{d,j,*}\mathbf{f}_{d,j,*}}$, all at the new inputs $\mathbf{X}_*$.

\section{EXPERIMENTS}
In this section, we explore the performance of the proposed
FNG method for jointly optimising all variational parameters,
hyper-parameters and inducing points. We also test
the hybrid (HYB) method proposed by \citep{Hugh2018}, and compare the
performance against Adam and SGD methods. We run experiments on different toy and real datasets, for
all datasets we use a splitting of $75\%$ and $25\%$ for training
and testing respectively. The experiments consist on evaluating the
method's performance when starting with 20 different initialisations
of $q(\boldsymbol{\theta})$'s parameters to be optimised. We report the
negative evidence lower bound (NELBO) shown in Eq. \eqref{eq:HetMOGP} for LMC and Eq. \eqref{eq:convHetMOGP} for CPM over the training set, and the negative log predictive density (NLPD) error for the test set; this error metric takes into account the
predictions' uncertainty \citep{metricNLPD}.

\begin{figure*}[hbtp]
	\centering
	{\hspace{-0.2cm}\includegraphics[width=0.33\textwidth,height=0.19\textheight]{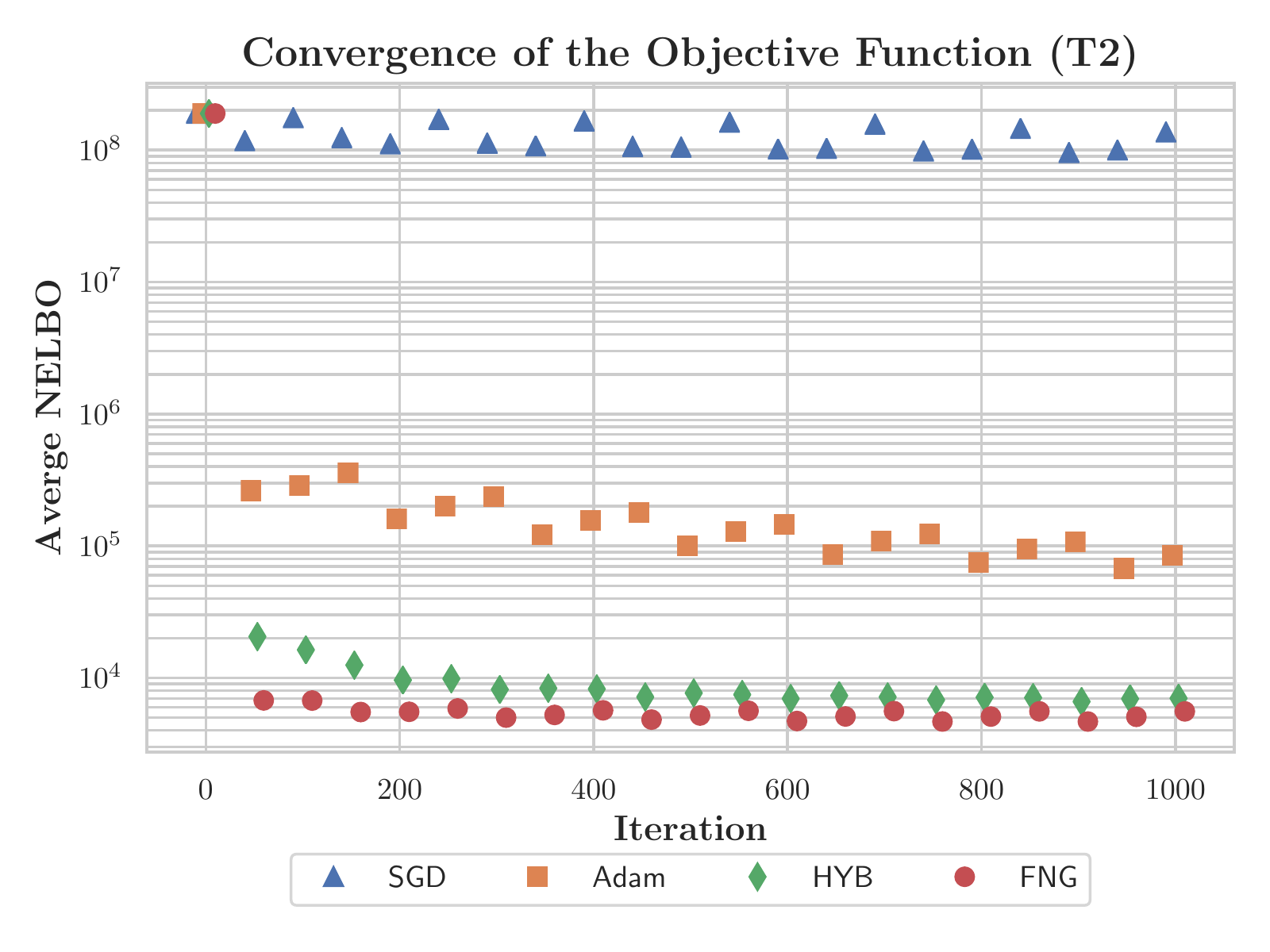}
		\hspace{-0.3cm}\includegraphics[width=0.35\textwidth,height=0.19\textheight]{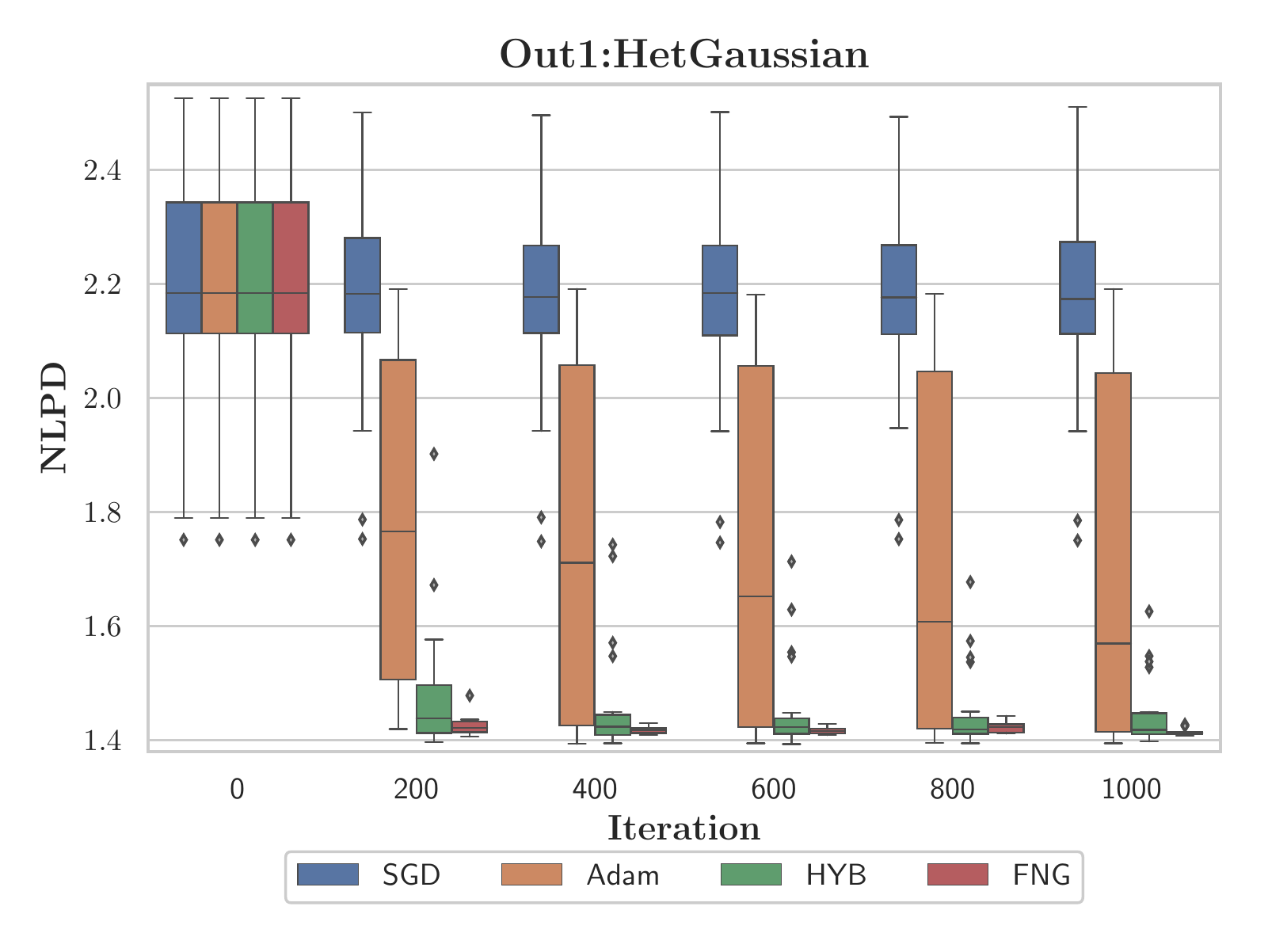}
		\hspace{-0.3cm}\includegraphics[width=0.35\textwidth,height=0.19\textheight]{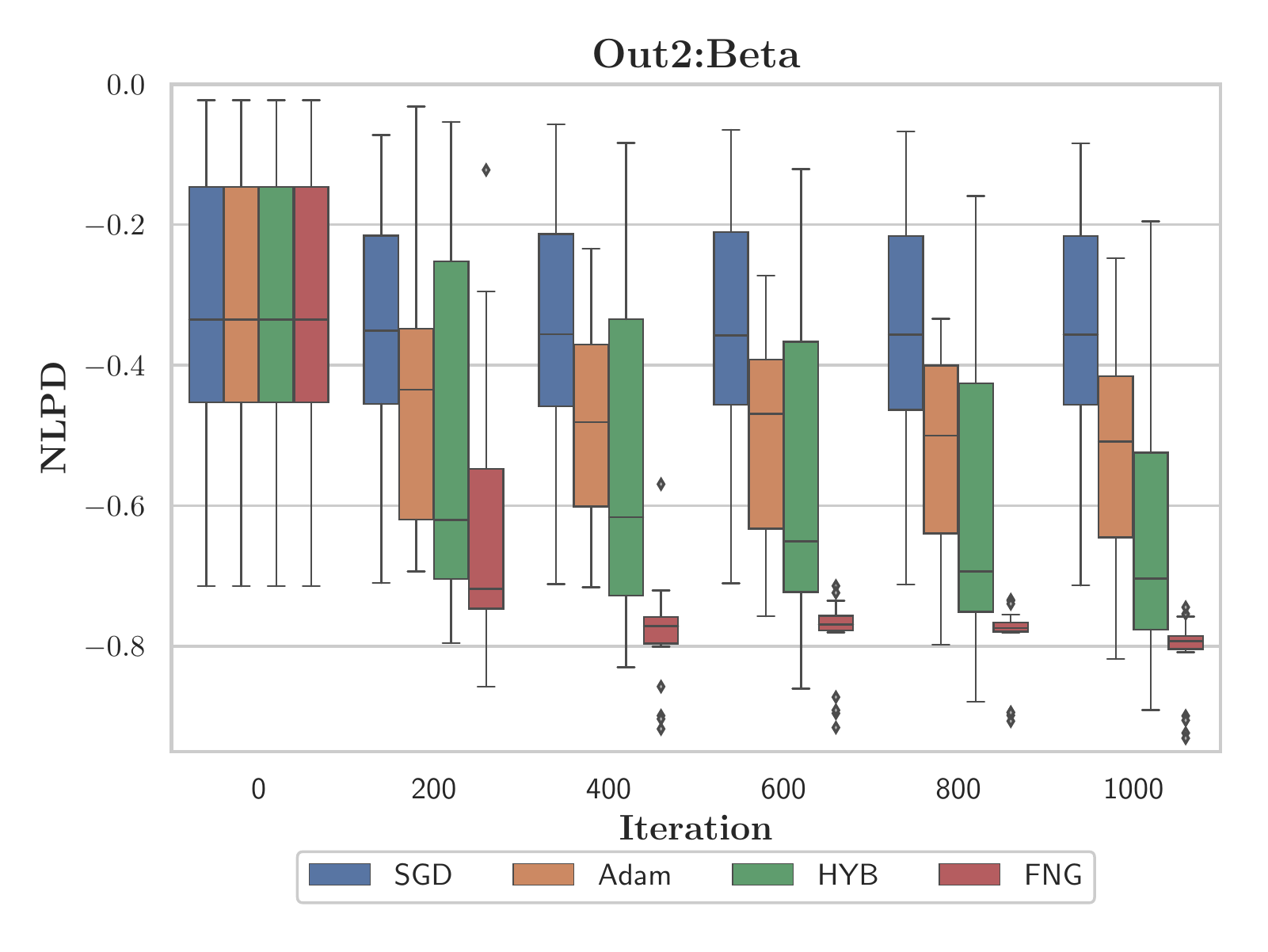}
		\hspace{-0.3cm}\includegraphics[width=0.335\textwidth,height=0.19\textheight]{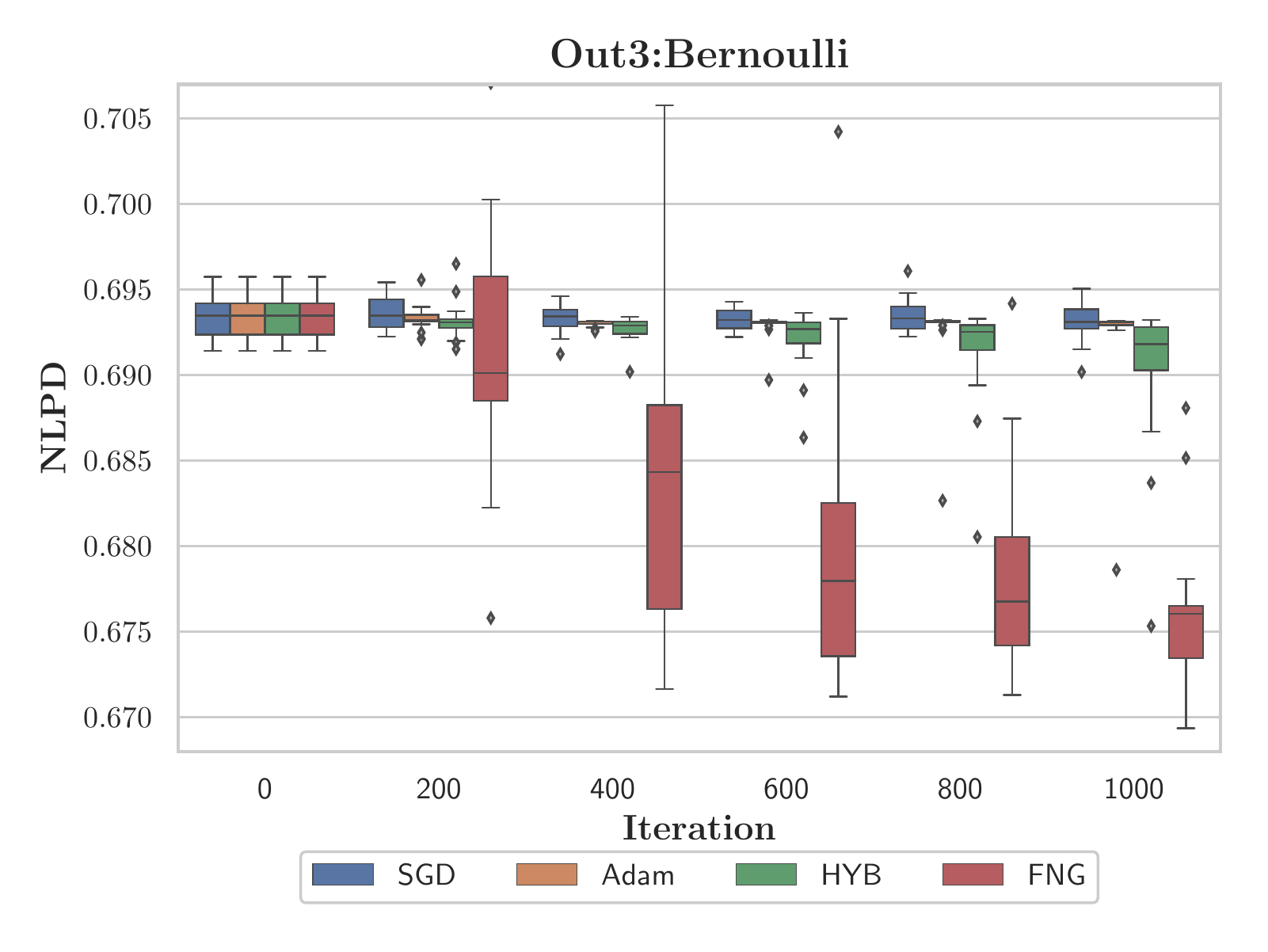}
		\hspace{-0.3cm}\includegraphics[width=0.34\textwidth,height=0.19\textheight]{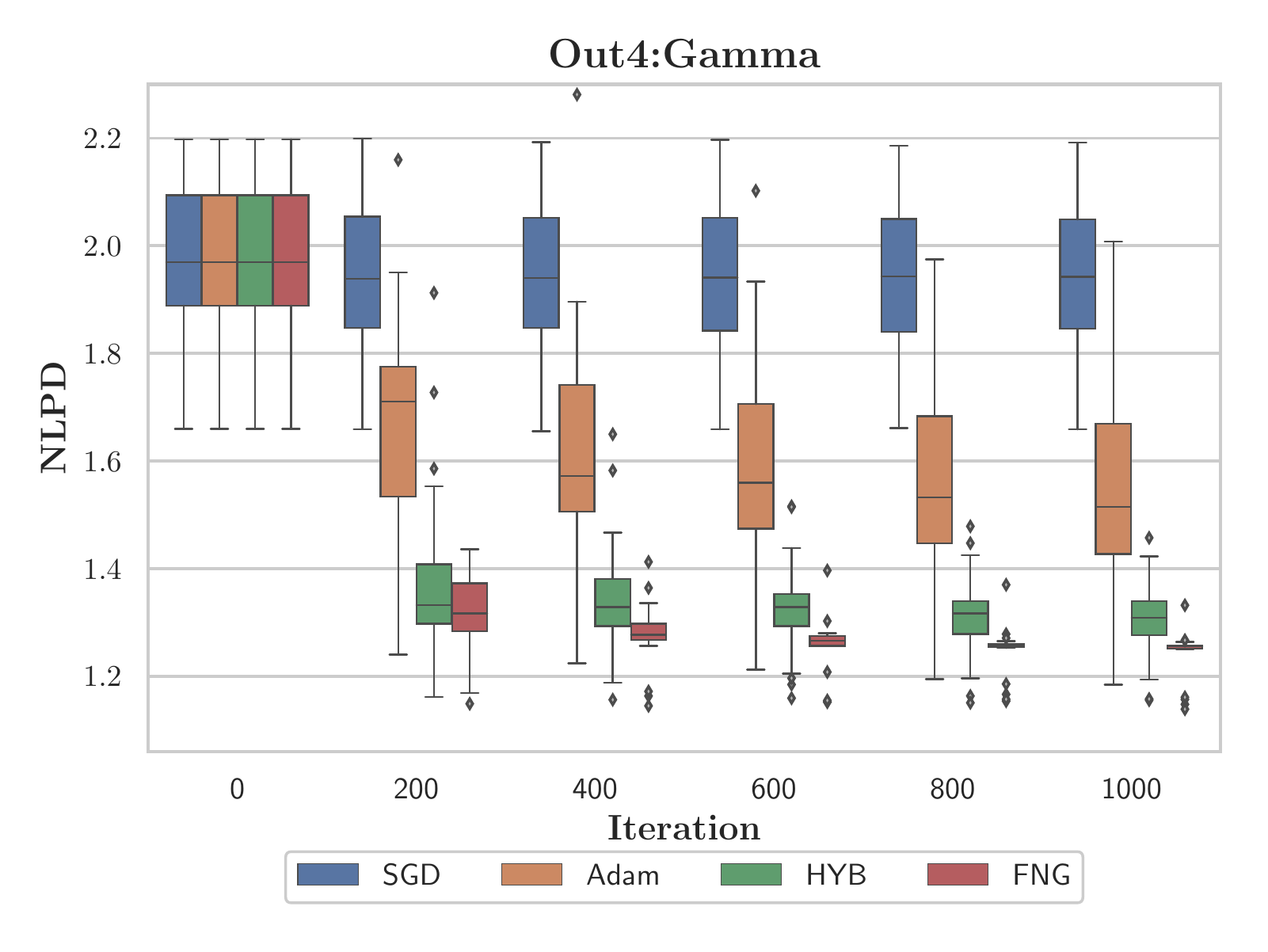}
		\hspace{-0.30cm}\includegraphics[width=0.34\textwidth,height=0.19\textheight]{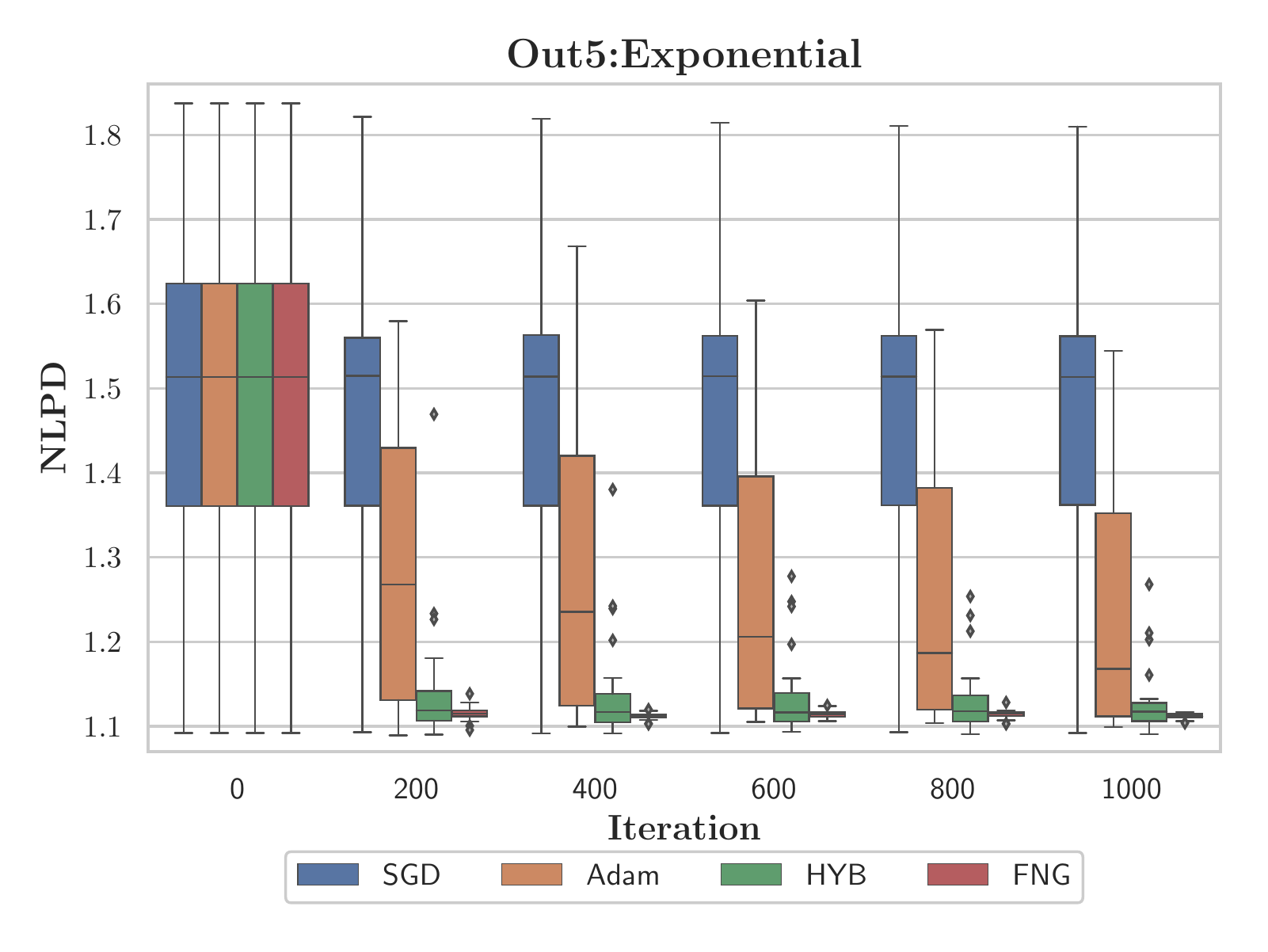}}
	\caption{Performance of the different inference methods on the T2 dataset for $P=10$ using 20 different initialisations. The top left sub-figure shows the average NELBO convergence. The other sub-figures show the box-plot trending of the NLPD over the test set for each output. The box-plots at each iteration follow the legend's order from left to right: SGD, Adam, HYB and FNG. The isolated diamonds that appear in the outputs' graphs represent ``outliers".}
	\label{fig:toy2}
\end{figure*}

\begin{figure*}[hbtp]
	\centering
	{\hspace{-0.15cm}\includegraphics[width=0.33\textwidth,height=0.19\textheight]{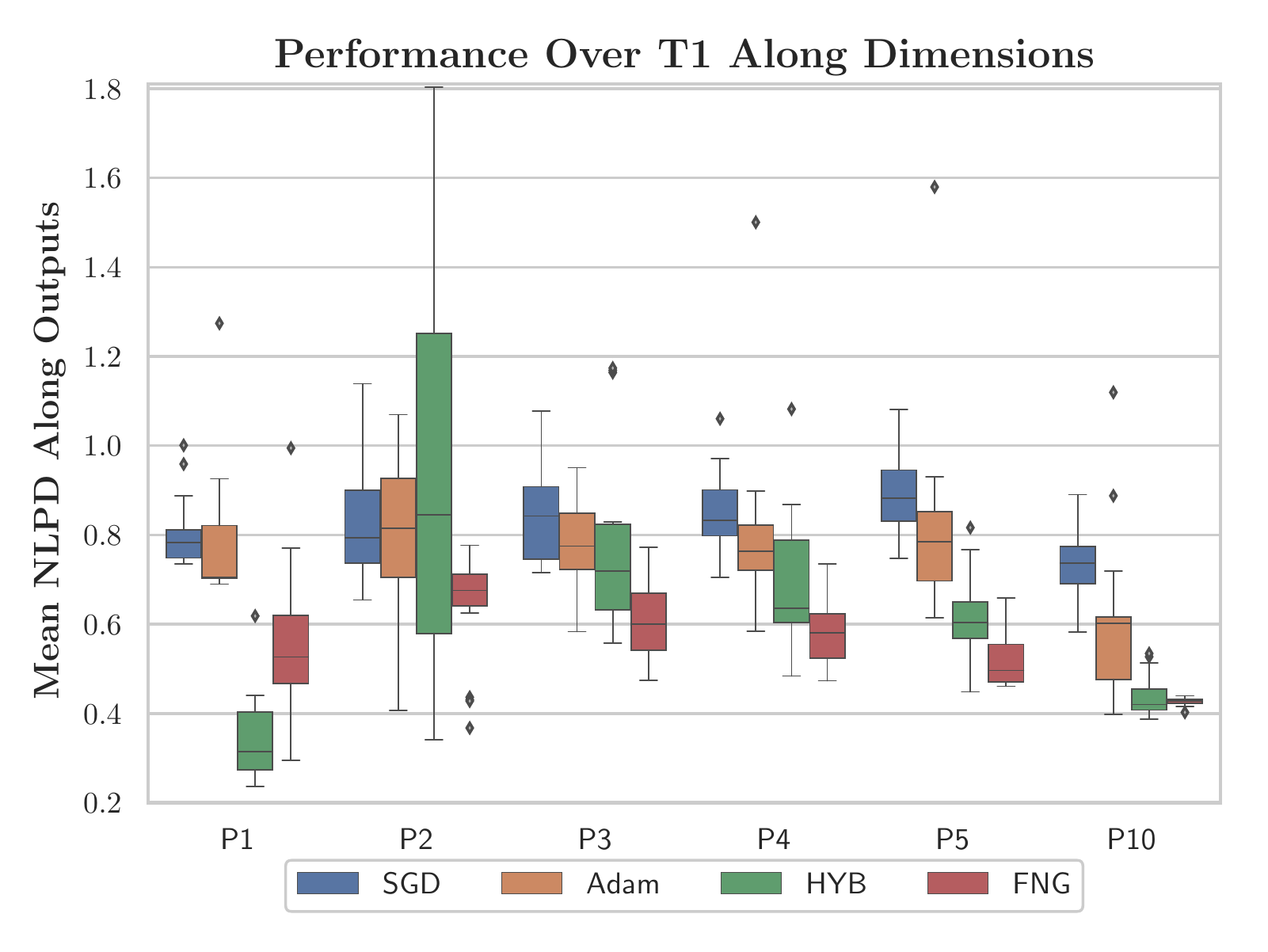}
		\hspace{-0.15cm}\includegraphics[width=0.33\textwidth,height=0.19\textheight]{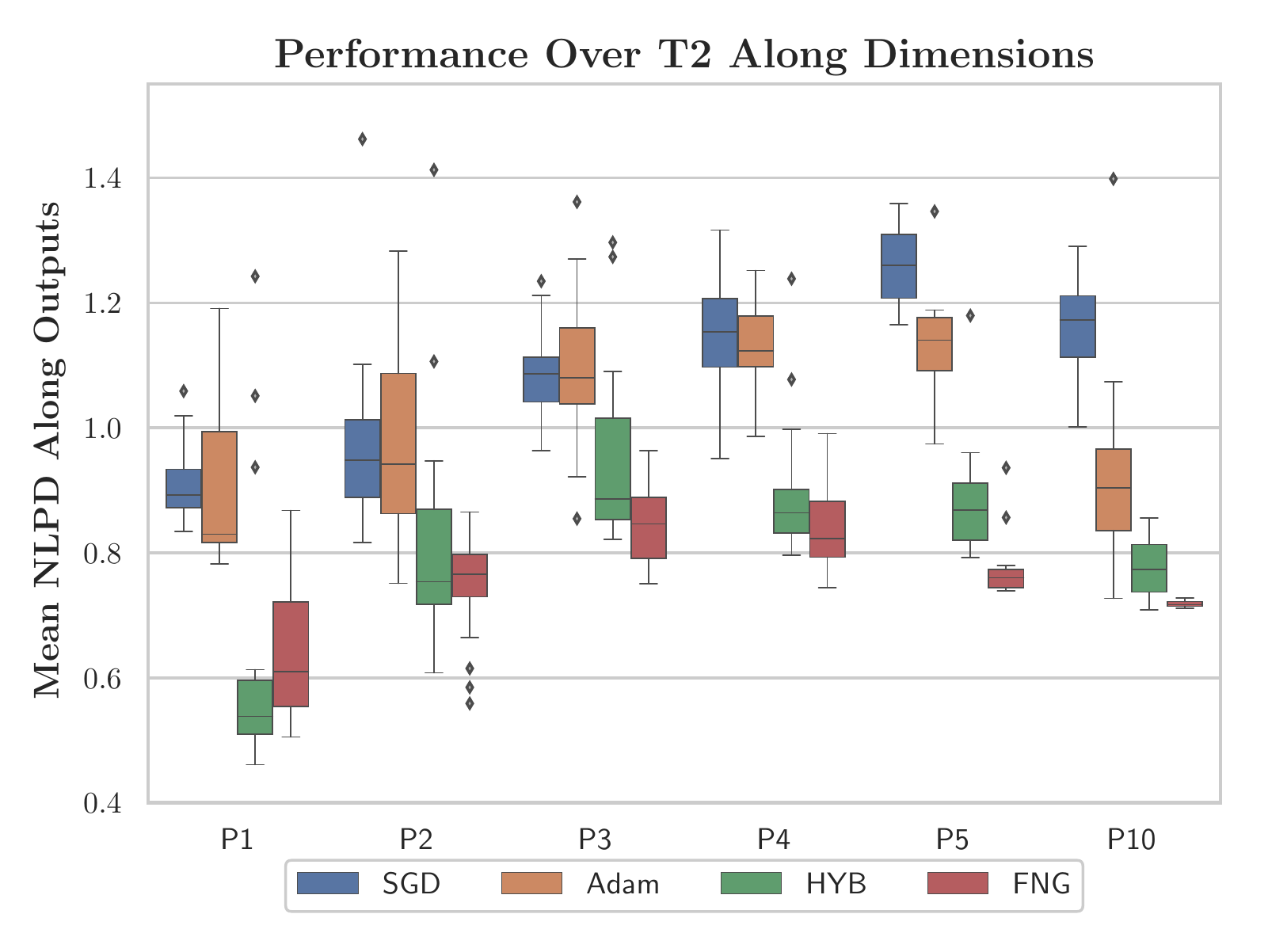}
		\hspace{-0.15cm}\includegraphics[width=0.33\textwidth,height=0.19\textheight]{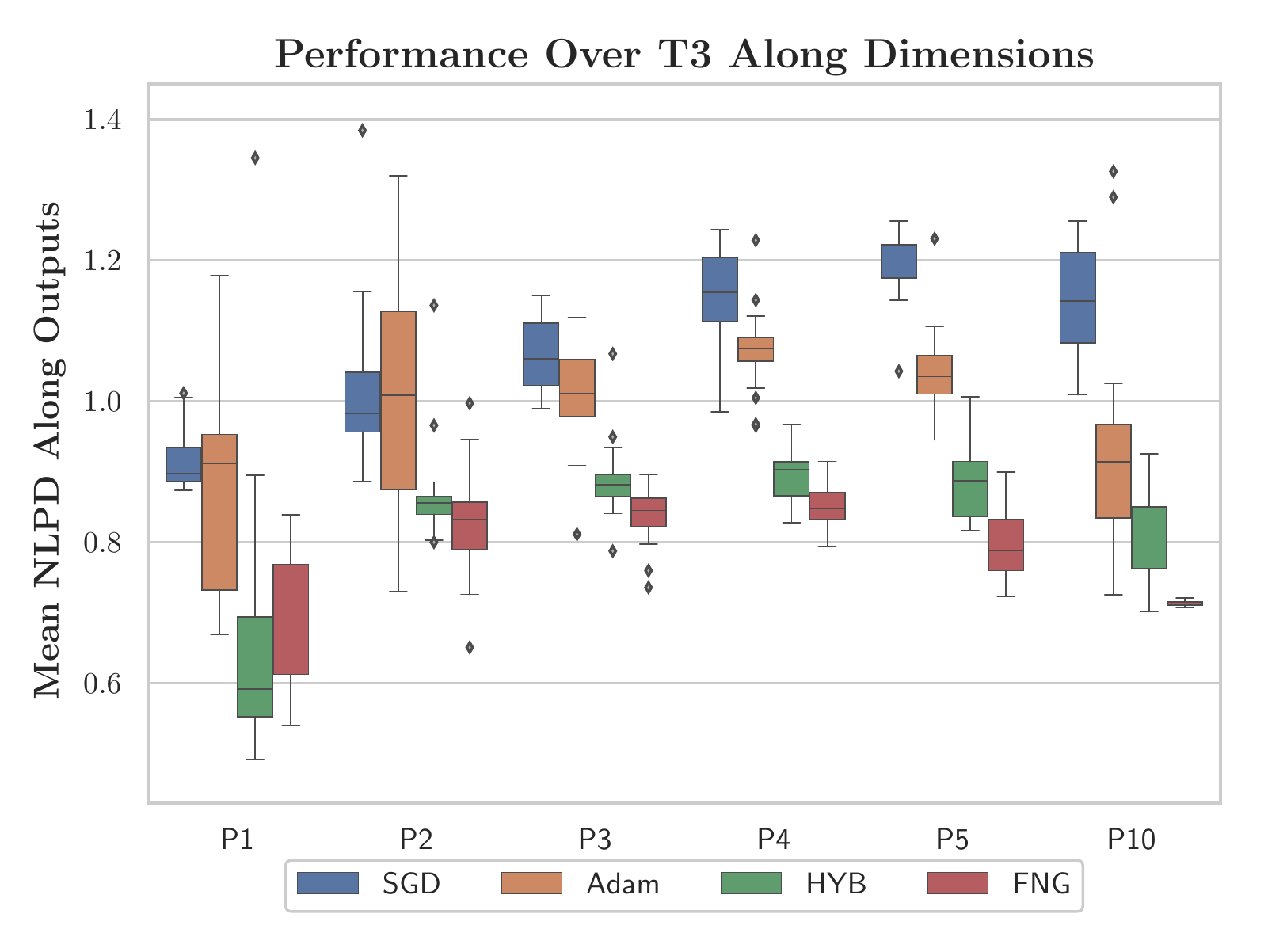}}
	\caption{Trending of the Mean NLPD along outputs for 20 different initialisations. Performance over: T1 (left), T2 (middle) and T3. Each sub-figure summarises the Mean NLPD of SGD, Adam, HYB and FNG methods along dimensions $P=\{1,2,3,4,5,10\}$. The box-plots at each $P$ follow the legend's order.}
	\label{fig:toys}
\end{figure*}

%\begin{figure*}[hbtp]
%	\centering
%	{}
%	\caption{Performance of the different inference methods on the TOY2 dataset for $D=20$. The average training performance NELBO (first column). From second column the average test NLPD for the outputs with Gaussian likelihoods. The average NLPD shows a deviation for 10 different hyper-parameters initialization.}
%	\label{fig:toy6}
%\end{figure*}

\subsection{Optimising the HetMOGP with LMC on Toy Data}

Particularly we are interested in looking at the performance of HetMOGP with LMC when increasing the number of outputs, which implies rising also the heterogeneity of the output data. Given that the inducing points $\mathbf{Z}$ have the same input space dimensionality and strongly affect the performance of sparse MOGPs, we are also interested in assessing the behaviour when increasing the input space dimensionality. For all the toys we define an input space $\mathbf{X}\in [0,1]^{N\times P}$ with $N=2\times 10^3$ observations, we analyse a set of different dimensions $P=\{1,2,3,4,5,10\}$. We assume a number of $Q=3$ with an EQ kernel $k_q(\cdot,\cdot)$, and the inducing points $\mathbf{Z}_q \in \mathbb{R}^{M \times P}$, with $M=80$. We run the experiments using mini-batches of $50$ samples at each iteration, and we use one sample to approximate the expectations w.r.t $q(\boldsymbol{\theta})$ in Eq. \eqref{eq:new_ELBO}. Below we describe the characteristics of each toy dataset. \medskip\\
\textbf{Toy Data 1 (T1):} the first toy example consists of three outputs $D=3$; the first output is $y_1 \in \mathbb{R}$, the second $y_2 \in [0,1]$ and the third $y_3 \in \{0,1\}$. We use a Heteroscedastic-Gaussian (HetGaussian), a Beta and Bernoulli distribution as the likelihoods for each output, respectively.\medskip\\
\textbf{Toy Data 2 (T2):} the second toy example consists of five
outputs $D=5$, where the first three are exactly the same ones as T1
with the same likelihoods and the two additional ones are
$y_4 \in [0,\infty]$, and $y_5 \in [0,\infty]$. We use a
Gamma and an Exponential distribution for those latter outputs, respectively.\medskip\\
\textbf{Toy Data 3 (T3):} the third toy example consists of ten
outputs $D=10$, where the data type of the first five outputs
$\{y_d\}_{d=1}^5$ is exactly the same as T2. Also, the last five
outputs $\{y_d\}_{d=6}^{10}$ share the same data type of the outputs in
T2. We use the following ten likelihoods: HetGaussian, Beta,
Bernoulli, Gamma, Exponential, Gaussian (with
$\sigma_{\text{lik}}=0.1$), Beta, Bernoulli, Gamma and
Exponential. The data of the first five outputs is not the same as the
last ones since the distributions of the generative model depend on
the LCCs $a_{d,j,q}$ that generate the LPFs
in Eq. \eqref{eq:linear_comb2}. \footnote{The code with all toy
	configurations is publicly available in the repository:\\
	https://github.com/juanjogg1987/Fully\_Natural\_Gradient\_HetMOGP}

In order to visualise the convergence performance of the methods, we
show results for T2 which consists of five outputs, where all of them
are used in T3 and three of them in T1. We focus on the example for
which $P=10$ as the dimensionality. Fig.\ref{fig:toy2} shows the
behaviour of the different algorithms over T2, where its top left
sub-figure shows the average convergence of the NELBO after
running 20 different initialisations. The figure shows that our FNG
method tends to find a better local optima solution that minimises the
NELBO followed by the HYB, Adam and SGD. The other sub-figures titled
from Out1 to Out5 show the model's average NLPD achieved by each of
the methods over the test set. From Fig.\ref{fig:toy2} we can notice
that the SGD method does not progress much through the inference
process achieving the poorest performance along the diverse
outputs. The Adam method presents a big variance along the different
outputs, showing its ability to explore feasible solutions,
but arriving at many different poor local minima. Particularly, for
the output 3, a Bernoulli likelihood, the method hardly moves from its
initial NLPD value, showing in the figure a tiny variance without much
improvement. This means the method lacks exploration and rapidly
becomes trapped in a very poor local minima. The HYB method in general
shows smaller error bars than Adam and SGD. Indeed, it reaches low
NLPD results for Gamma, HetGaussian and Exponential likelihoods, with
similar behaviour to our FNG method in the two latter
distributions. Although, it is difficult for HYB to achieve a proper
NLPD performance on the distributions Beta and Bernoulli; though for the Beta distribution presents boxes with big variance meaning that it
arrives to many different solutions, the NLPD's mean shows a
trending to weak solutions. For the Bernoulli is deficient in exploring,
so it also ends up in poor solutions. Our FNG method is consistent
along the diverse outputs, usually tending to richer local minima
solutions than the other methods. For the Beta and Gamma outputs, FNG
makes a confident progress and even shows some ``outliers" below its
boxes which means that our method has the ability to eventually provide better solutions than the other methods. For the Bernoulli
distribution, Fig.\ref{fig:toy2} shows that FNG presents big variance
boxes, but with a tendency to much better solutions than the other
methods. This big variance effect let us confirm that our proposed method actually
takes advantage of the stochastic exploration induced over the model
hyper-parameters for avoiding poor local minima solutions.
\begin{figure*}[hbtp]
	\centering
	{\hspace{-0.25cm}\includegraphics[width=0.255\textwidth,height=0.19\textheight]{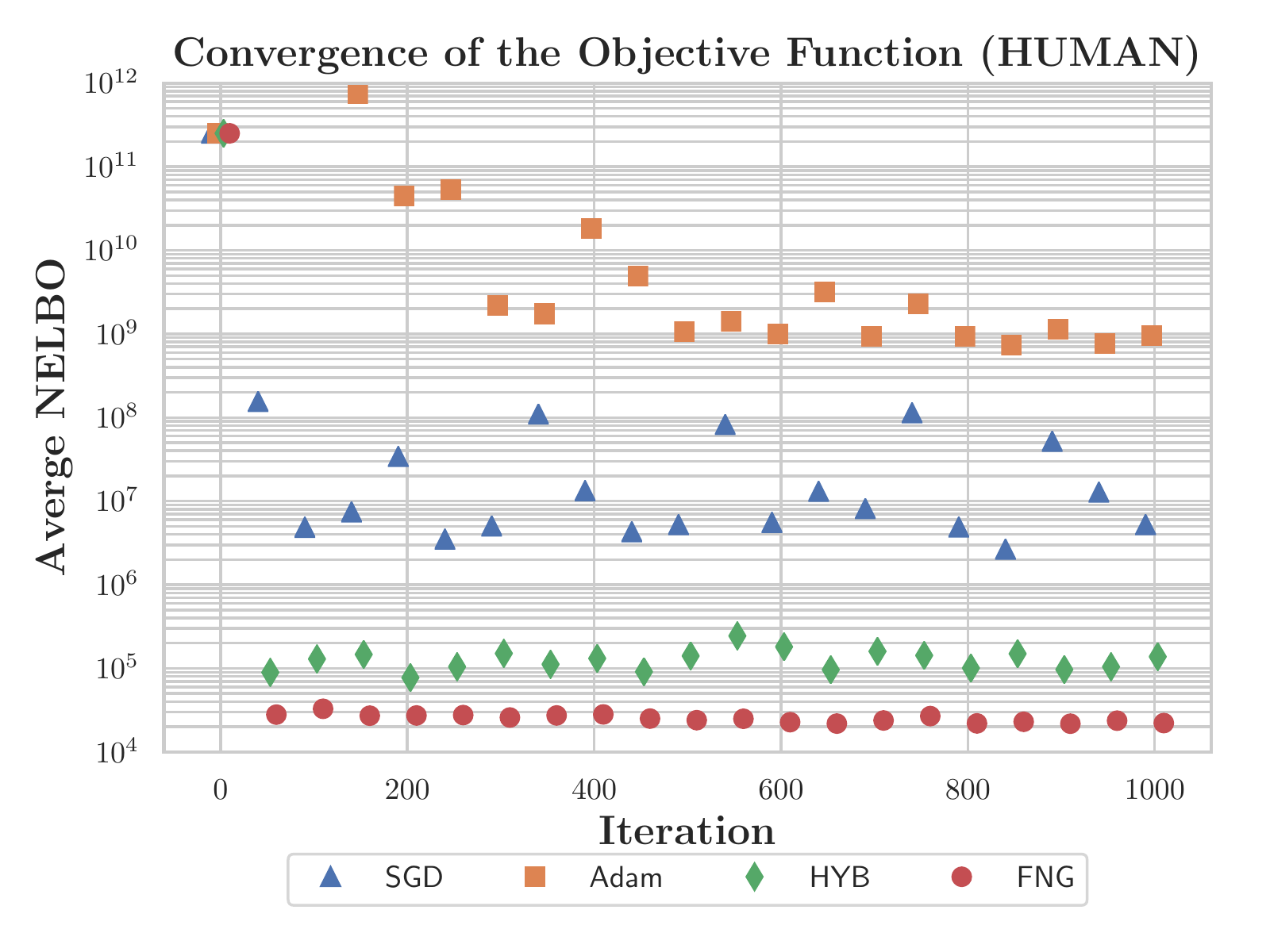}
		\hspace{-0.25cm}\includegraphics[width=0.255\textwidth,height=0.19\textheight]{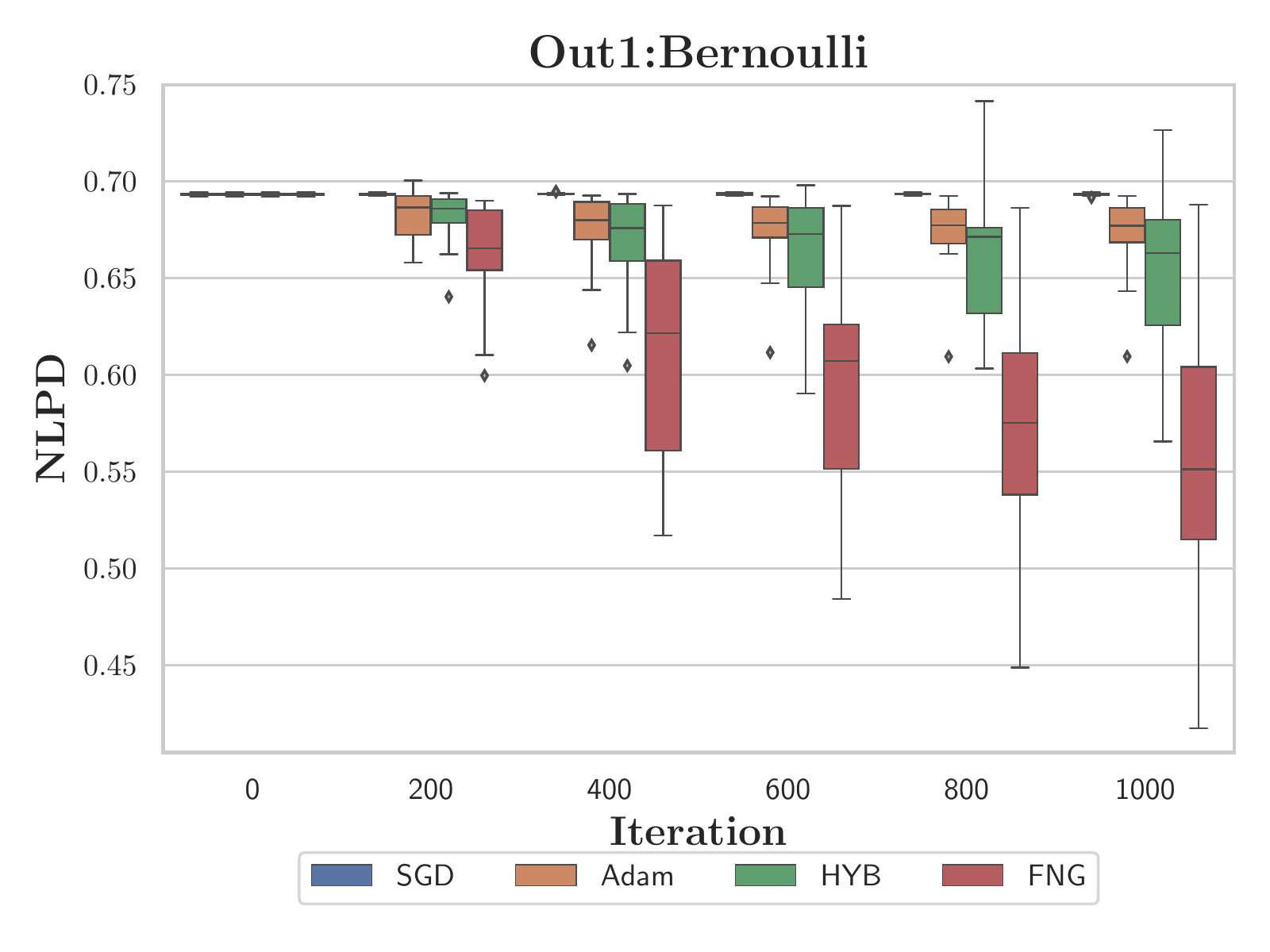}
		\hspace{-0.25cm}\includegraphics[width=0.255\textwidth,height=0.19\textheight]{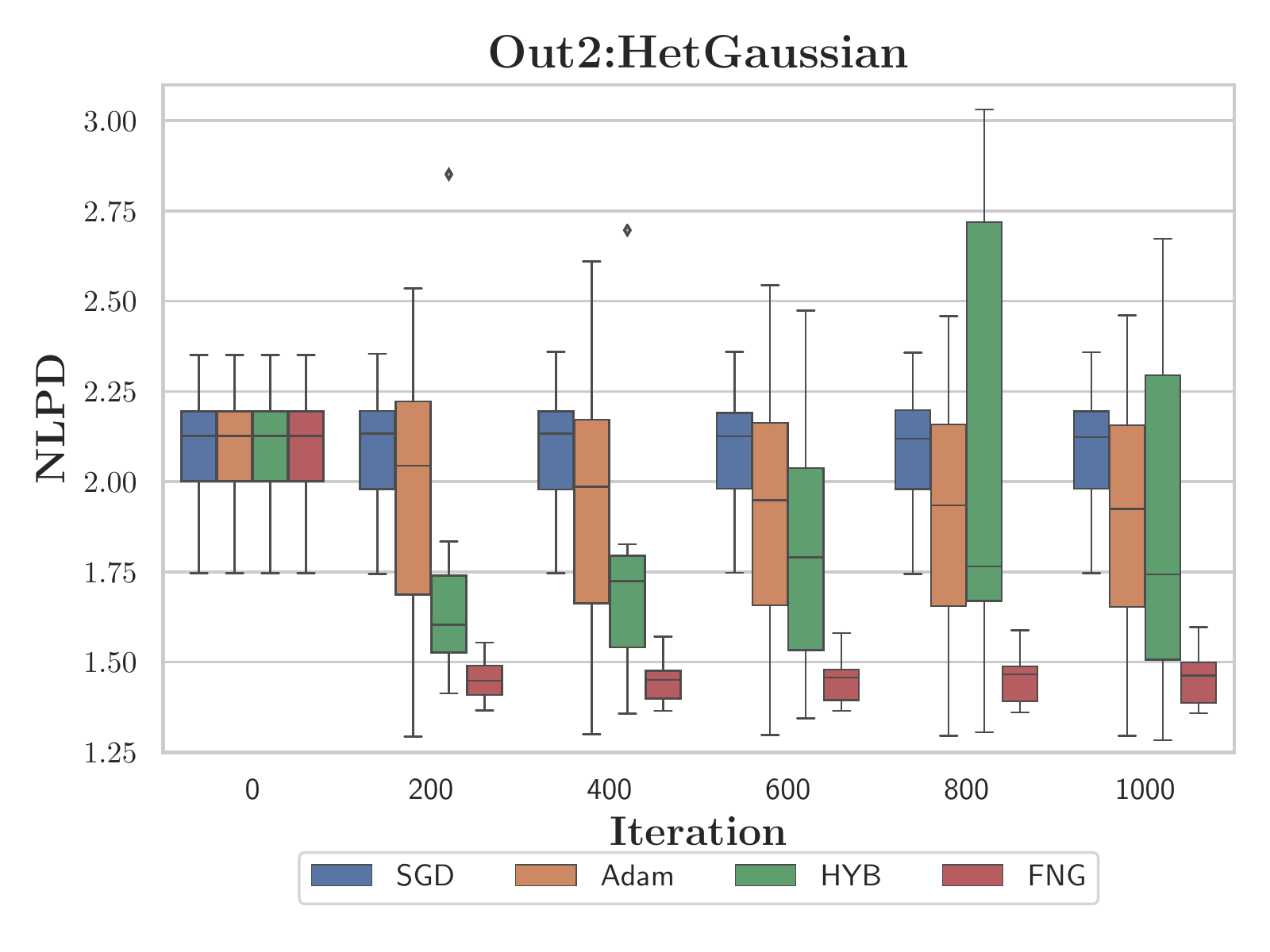}
		\hspace{-0.25cm}\includegraphics[width=0.255\textwidth,height=0.19\textheight]{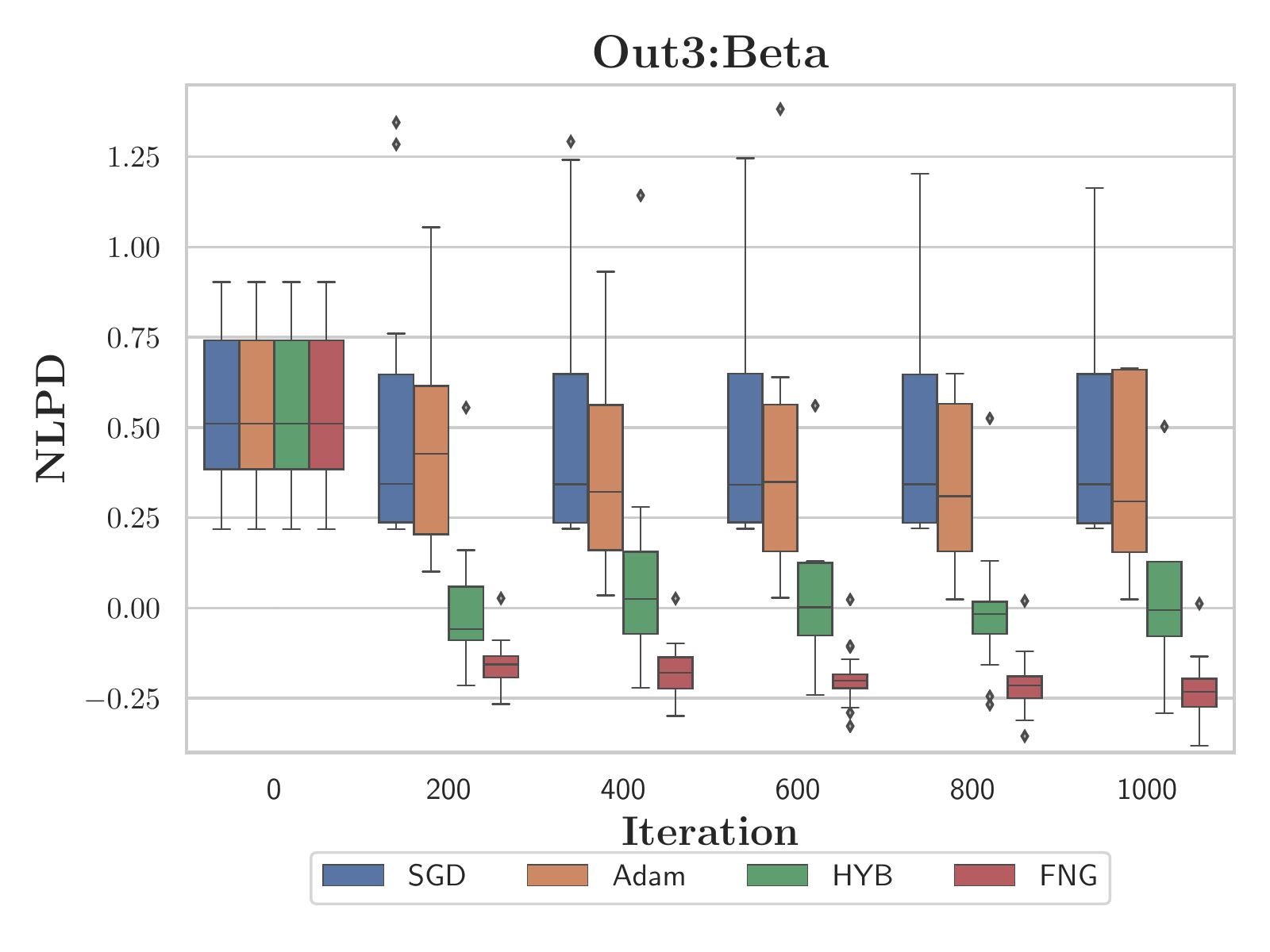}}
	\caption{Performance of the diverse inference methods on the HUMAN dataset using 20 different initialisations. The left sub-figure shows the average NELBO convergence of each method. The other sub-figures show the box-plot trending of the NLPD over the test set for each output. The box-plots at each iteration follow the legend's order from left to right: SGD, Adam, HYB and FNG. The isolated diamonds that appear in the outputs' graphs represent ``outliers".}
	\label{fig:human}
\end{figure*}
\begin{figure*}[hbtp]
	\centering
	{\hspace{-0.25cm}\includegraphics[width=0.255\textwidth,height=0.19\textheight]{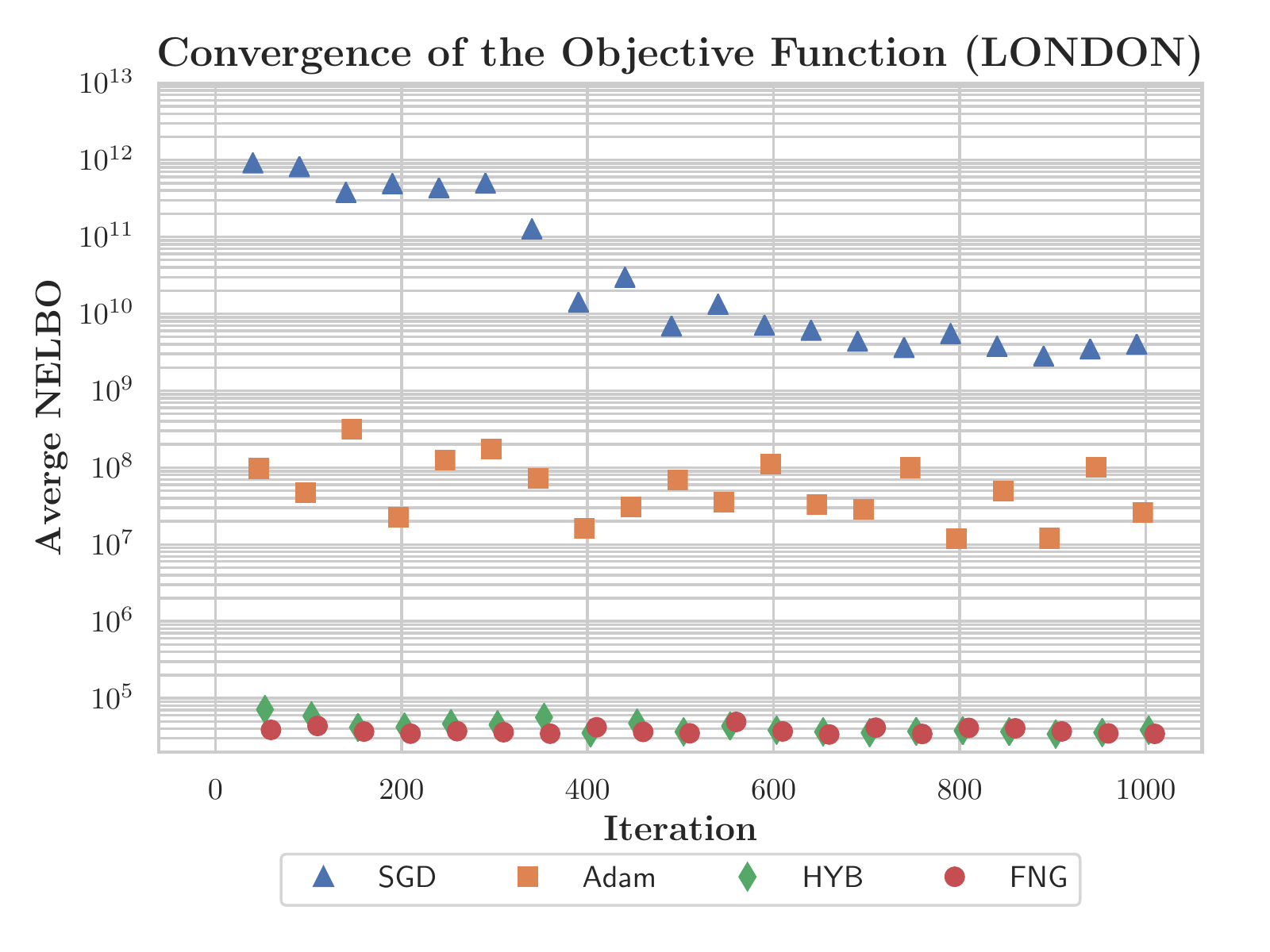}
		\hspace{-0.25cm}\includegraphics[width=0.255\textwidth,height=0.19\textheight]{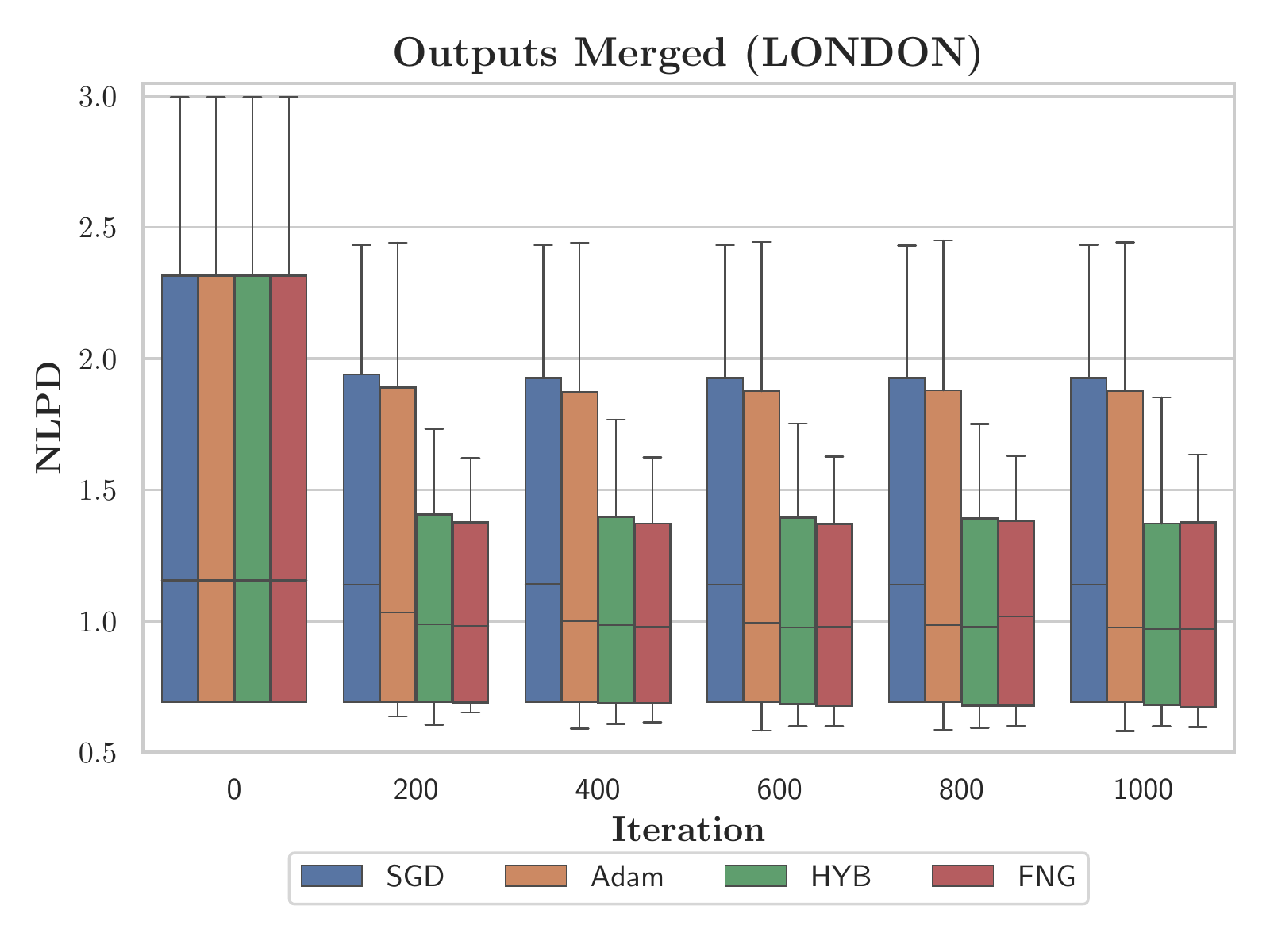}
		\hspace{-0.25cm}\includegraphics[width=0.255\textwidth,height=0.19\textheight]{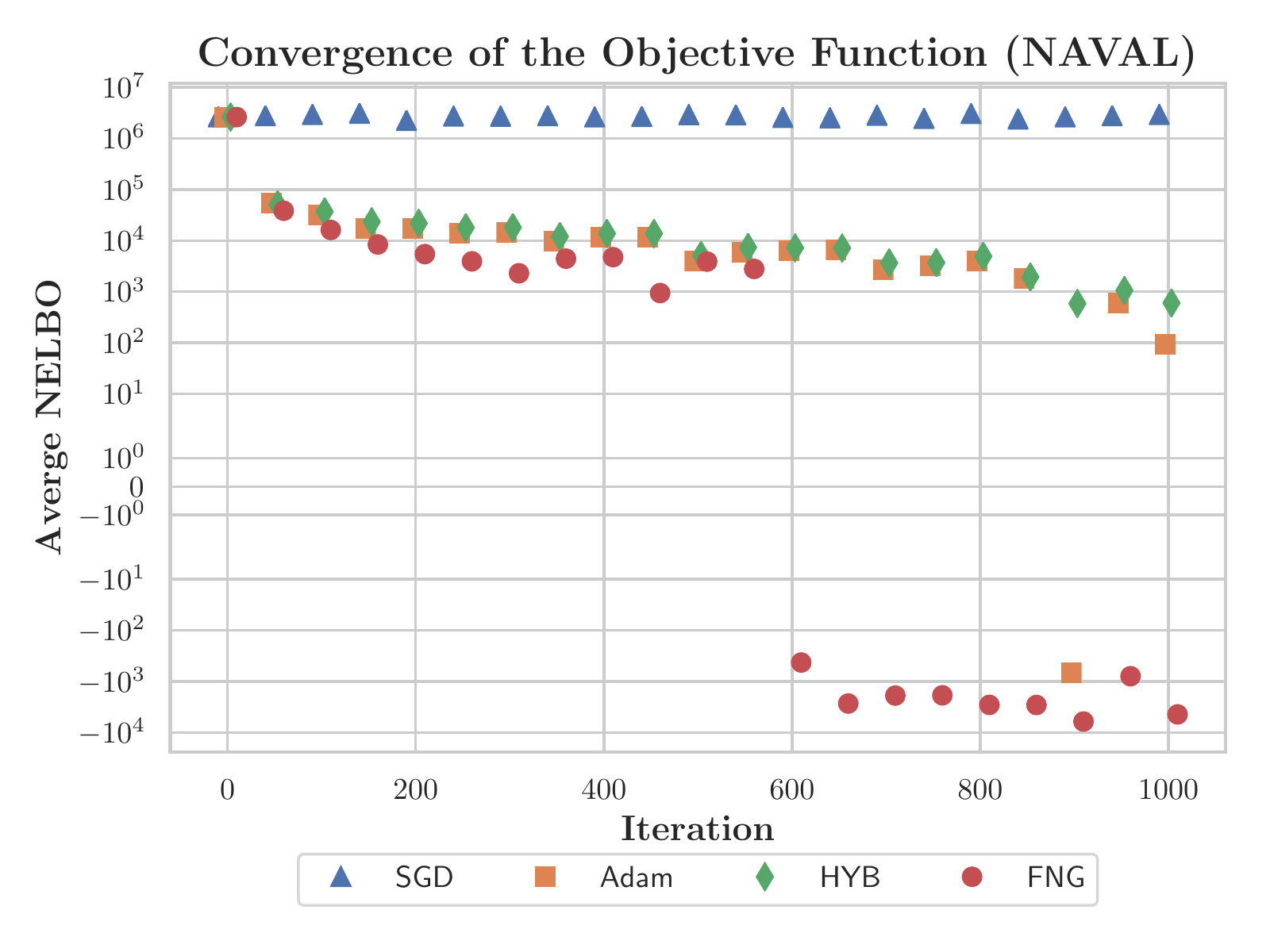}
		\hspace{-0.25cm}\includegraphics[width=0.255\textwidth,height=0.19\textheight]{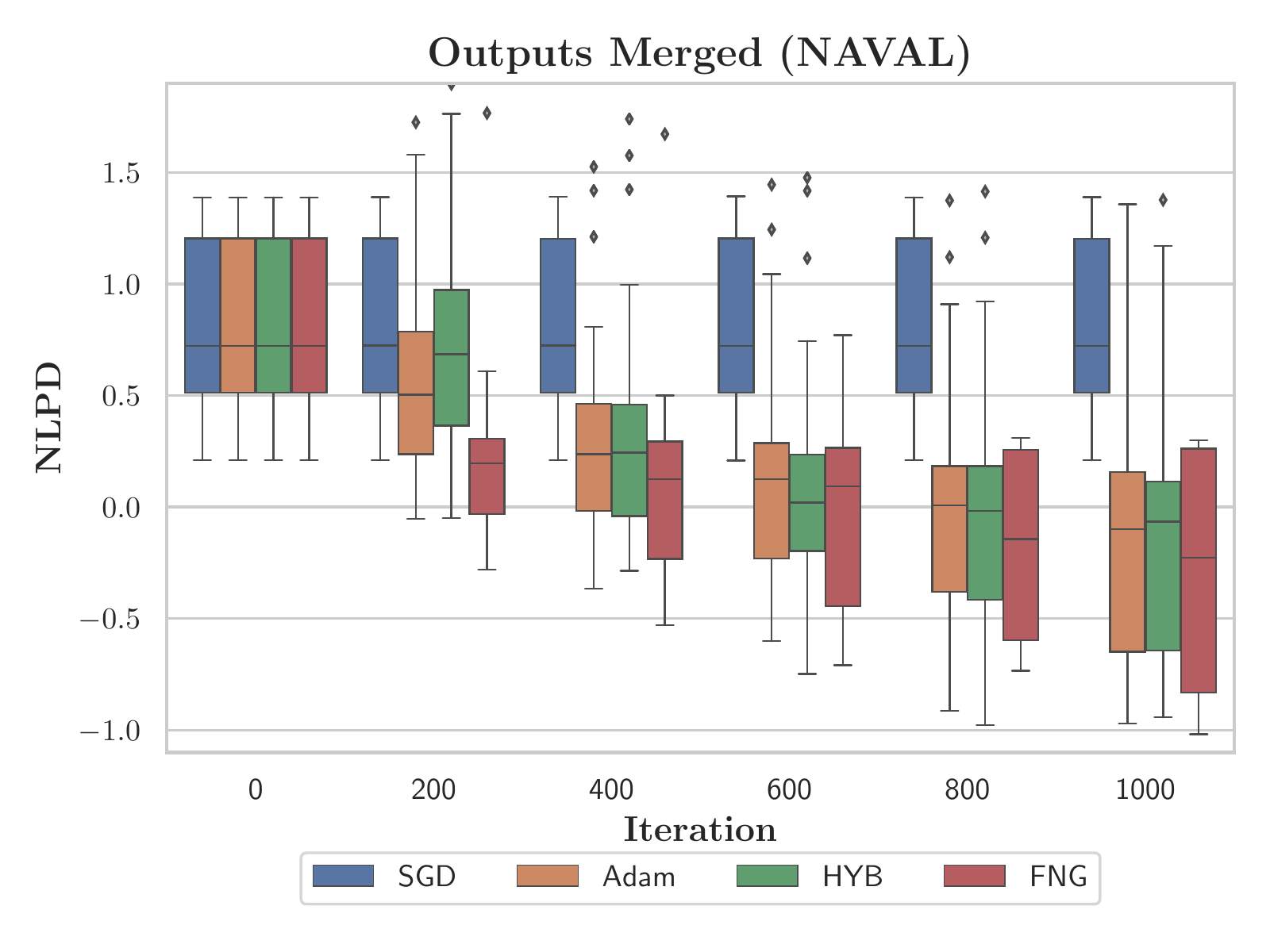}}
	\caption{Performance of the diverse inference methods on the LONDON and NAVAL datasets using 20 different initialisations. Sub-figures left and middle-left correspond to LONDON; middle-right and right refer to NAVAL. For each dataset we show the average NELBO convergence of each method and the box-plot trending of the NLPD over the test set across all output. The box-plots at each iteration follow the legend's order from left to right: SGD, Adam, HYB and FNG. The isolated diamonds that appear in the outputs' graphs represent ``outliers".}
	\label{fig:LMC_london_naval}
\end{figure*}
\\Figure \ref{fig:toys} summarises
the behaviour along the different dimensions $P$ for each toy example. We notice from Fig. \ref{fig:toys} that our FNG method achieves
better test performance along distinct dimensions for all toys,
followed by the HYB, Adam and SGD methods, though HYB presents better results than FNG when $P=1$. All methods in general tend
to present big variances for T1 which consists of three outputs,
although this effect is reduced when the number of outputs is
increased. Our
FNG in general presents the smallest variance showing its
ability to find better local minima even with many
outputs. When increasing the dimensionality, the methods tend to
degrade their performance, but the less sensitive to such behaviour
are the HYB and FNG methods, where the latter, in general achieves the
lowest mean NLPD along outputs for the different toy examples. Apart
from the heterogeneous toys shown in this paper, we also ran
experiments for dimensions higher than $P=10$, although we noticed
that all methods behaved similar except for the SGD which demands a
very small step-size parameter that makes it progress slowly. We
believe that the toy examples become difficult to control in such
dimensions and probably the data observations become broadly
scattered. We also explored experiments increasing the mini-batch size
at each iteration, we noticed the gradient's stochasticity is reduced
helping to increase the convergence rates of all methods, but the ones
using NG perform better. When reducing the mini-batch size, our FNG method
usually performs better than the others probably due to the fact that it 
additionally exploits the probability distribution $q(\boldsymbol{\theta})$, imposed over the hyper-parameters and inducing points.
\subsection{Settings for Real Datasets Experiments}
In this subsection we describe the different real datasets used for our experiments (See appendix G for information about the web-pages where we took the datasets from).\medskip

\noindent\textbf{HUMAN Dataset:} the human behaviour dataset (HUMAN, $N_1,N_2=5\times 10^3,N_3=21\times 10^3, P=1$, $N_d$ associates the number of observations per output) contains information for monitoring psychiatric patients with a smartphone \textit{app}. It consists of three outputs; the first monitors use/non-use of \textit{WhatsApp}, $y_1 \in \{0,1\}$, the second represents distance from the patient’s home location, $y_2 \in \mathbb{R}$, and the third accounts for the number of smartphone active \textit{apps}, we rescale it to $y_3 \in [0,1]$. We use a Bernoulli, HetGaussian and a Beta distribution as the likelihoods for each output, respectively. We assume $Q=5$ latent functions.\medskip\\
\textbf{LONDON Dataset:} the London dataset (LONDON, $N=20\times 10^3 , P=2$) is a register of properties sold in London in 2017, consists of two outputs; the first represents house prices with $y_1 \in \mathbb{R}$ and the second accounts for the type of house, we use two types (flat/non-flat) with $y_2 \in \{0,1\}$, we use a HetGaussian and Bernoulli distribution as the likelihood for each output respectively. We assume $Q=3$ latent functions.\medskip\\
\textbf{NAVAL Dataset:} the naval dataset (NAVAL, $N=11\times 10^3, P=15$) contains information of condition based maintenance of naval propulsion plants, consists of two outputs: plant's compressor decay state coefficient and turbine decay state coefficient. We re-scaled both as $y_1,y_2 \in [0,1]$, and used a Beta and Gamma distribution as the likelihood for each output respectively. We assume $Q=4$ functions.\medskip\\
\textbf{SARCOS Dataset:} a seven degrees-of-freedom SARCOS anthropomorphic robot arm data, where the task is to map from a 21-dimensional input space (7 joint positions, 7 joint velocities, 7 joint accelerations) to the corresponding 7 joint output torques (SARCOS, $N=44.5\times 10^3, P=21, D=7$). We use a HetGaussian distribution as the likelihood for each output and assume $Q=3$ functions.\medskip\\
\textbf{MOCAP Dataset:} a motion capture data for a walking subject (MOCAP7, $N=744, P=1, D=40$). We use a HetGaussian distribution as the likelihood for each output and assume $Q=3$ functions.\medskip\\
For the first three datasets, the
number of inducing points per latent function is $M=80$ and for each
function $u_q(\cdot)$ we define an EQ kernel
like Eq. \eqref{eq:EQ_kern}. We run the experiments using mini-batches
of $50$ samples at each iteration, and we use one sample to
approximate the expectations with regard to $q(\boldsymbol{\theta})$
in Eq. \eqref{eq:new_ELBO}. For SARCOS we use mini-batches of $200$
due to its large number of observations, and given that MOCAP7 is
not a large dataset we use mini-batches of $5$ with $M=20$. The $Q$ assumption made for the datasets above follow that $Q=J$, for providing more flexibility to the LMC and CPM priors. Though, for SARCOS and MOCAP7, which present a high number of outputs, we selected $Q=3$. We did not assume $Q=J$ for these cases since the number $J$ of LPFs would be much bigger than the number of outputs overloading the computational complexity. This model selection problem of $Q$ has been studied in other works like \citep{Guarnizo2015,Tong2019}, we consider such a model selection problem beyond the scope of our work.  
\subsection{Optimising the HetMOGP with LMC on Real Data}
\noindent For this sub-section we explore our method's
behaviour over the HetMOGP with LMC on HUMAN, LONDON, NAVAL, SARCOS and MOCAP7.\medskip

\begin{figure*}[hbtp]
	\centering
	{\hspace{-0.25cm}\includegraphics[width=0.255\textwidth,height=0.19\textheight]{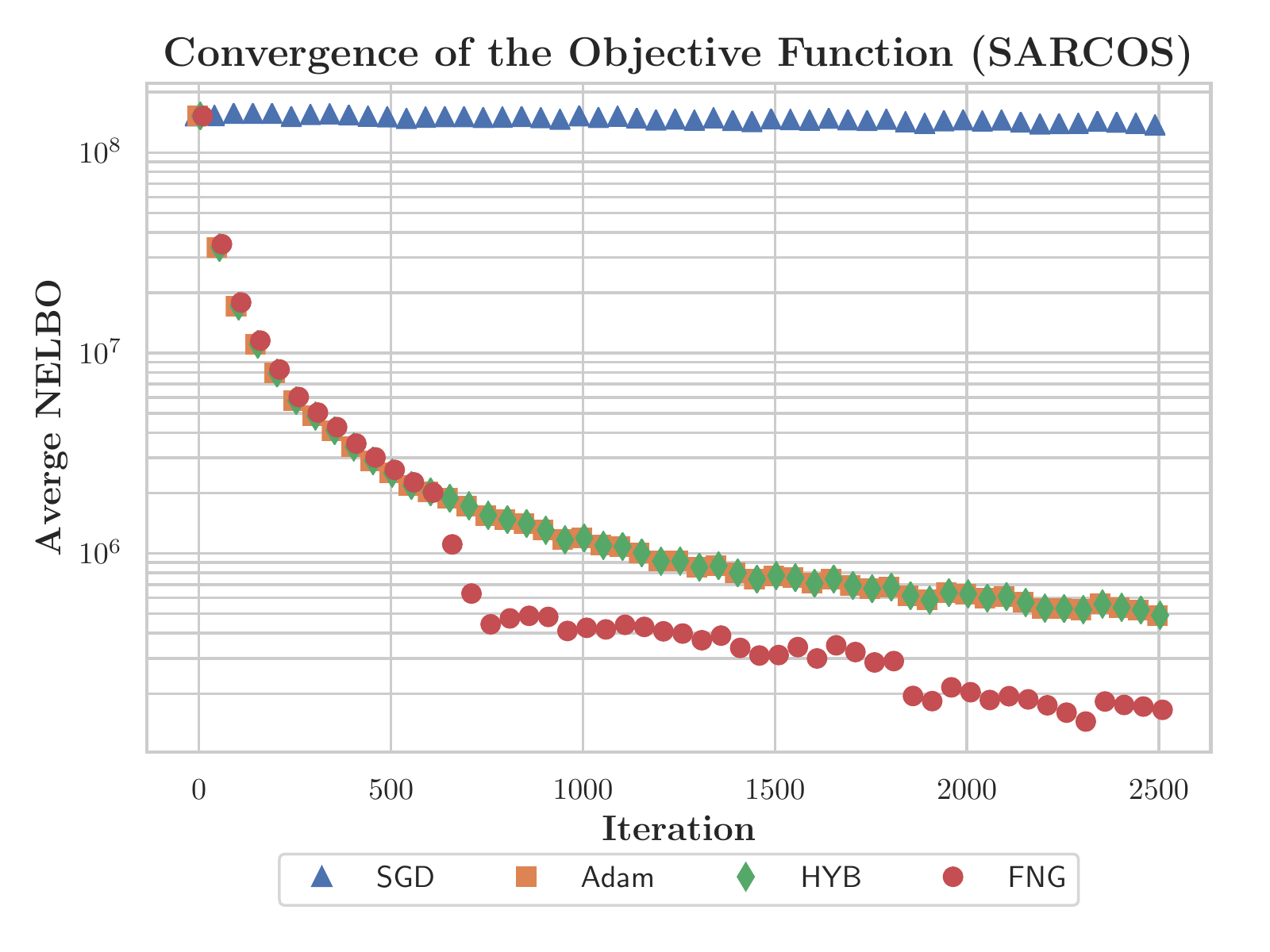}
		\hspace{-0.25cm}\includegraphics[width=0.255\textwidth,height=0.19\textheight]{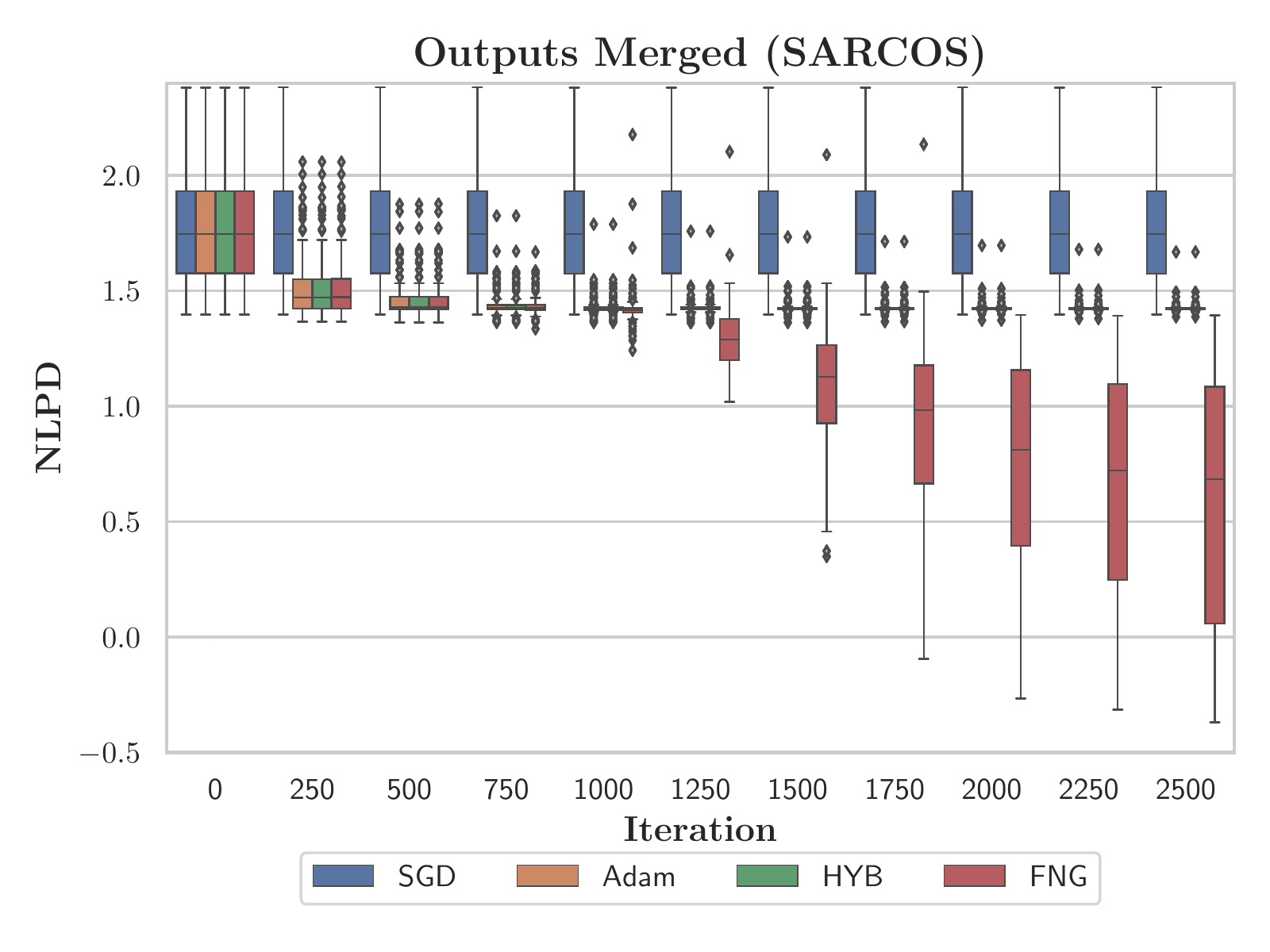}
		\hspace{-0.25cm}\includegraphics[width=0.255\textwidth,height=0.19\textheight]{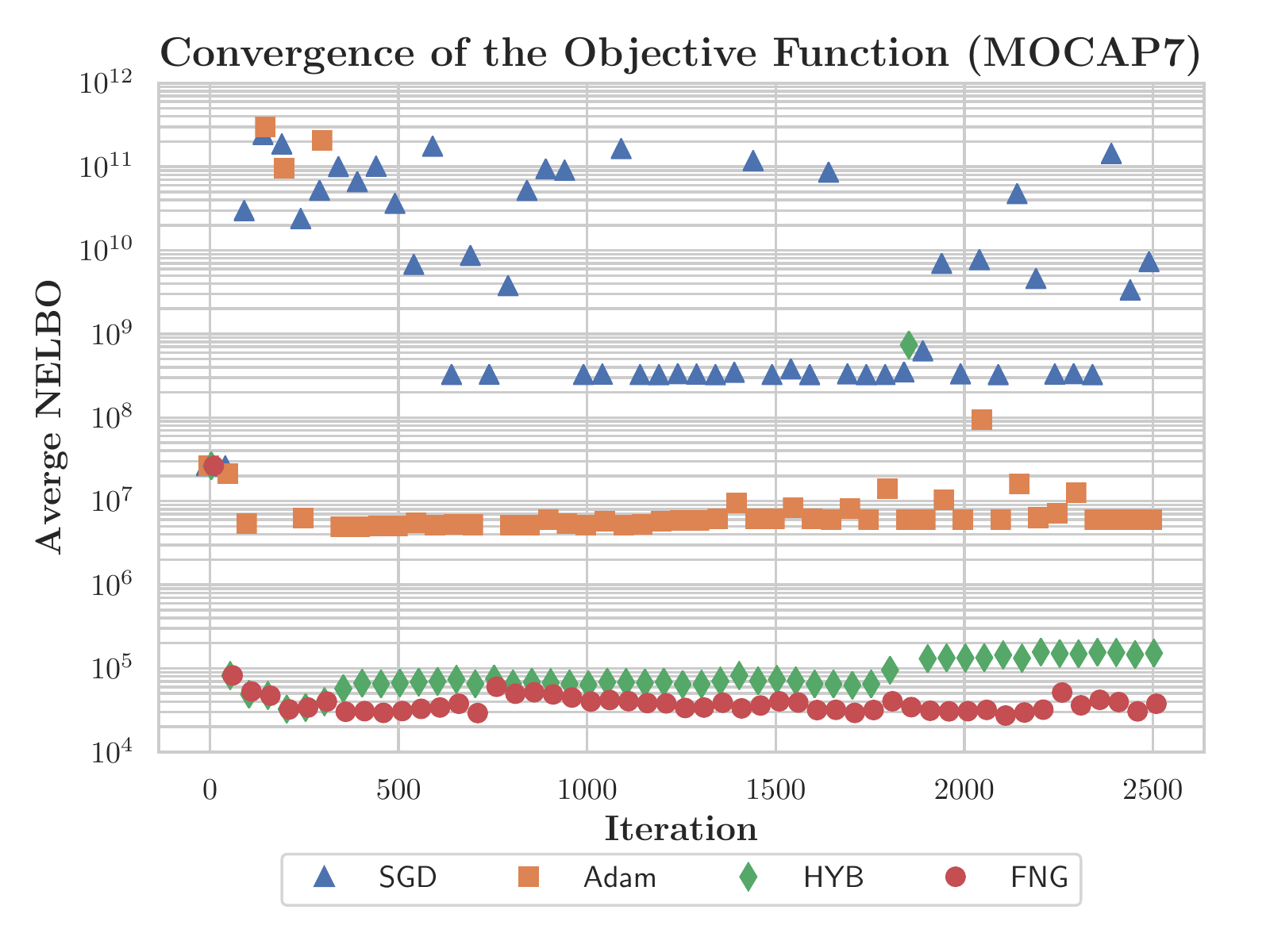}
		\hspace{-0.25cm}\includegraphics[width=0.255\textwidth,height=0.19\textheight]{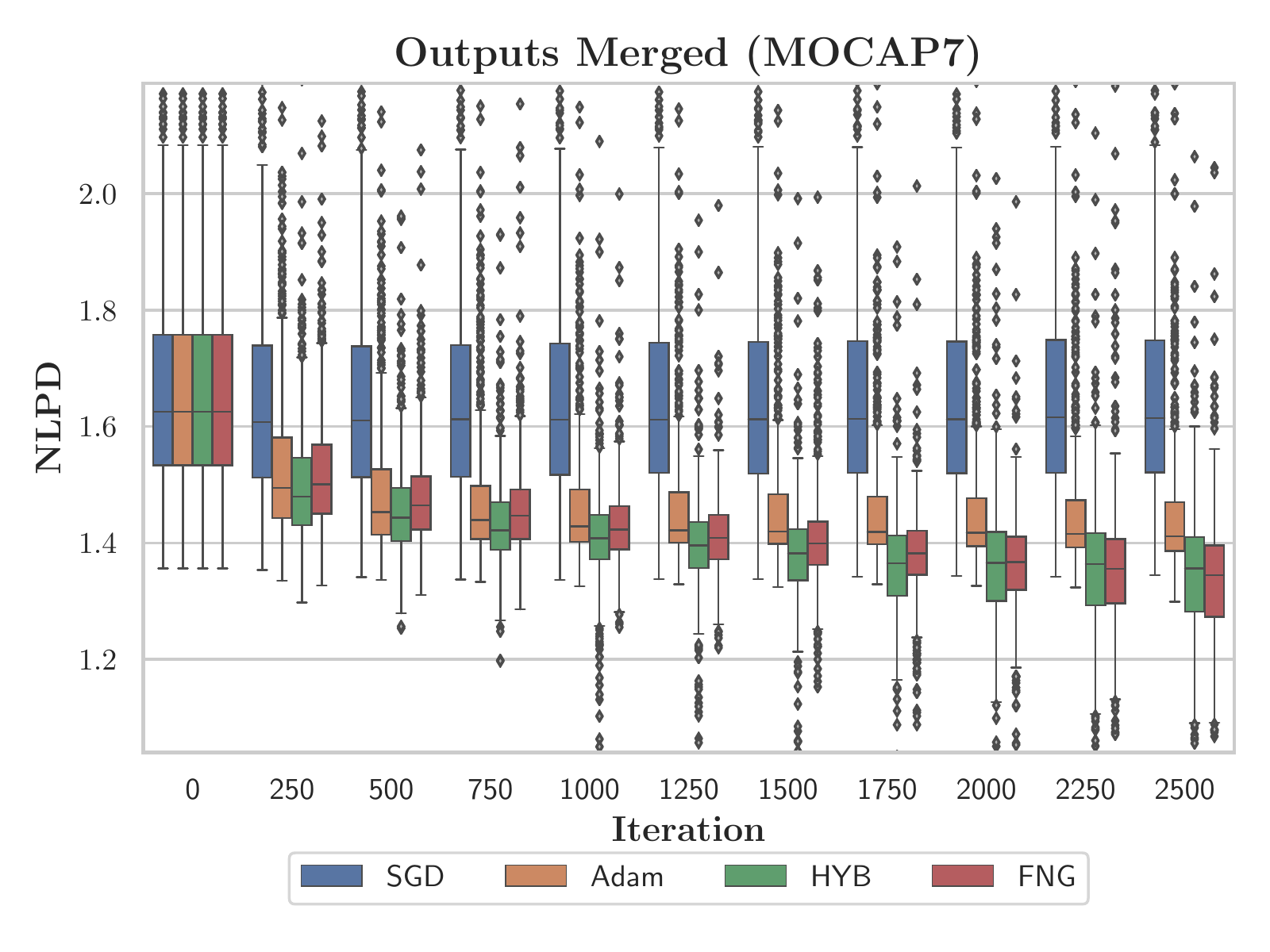}}
	\caption{Performance of the diverse inference methods on the SARCOS and MOCAP7 datasets using 20 different initialisations for HetMOGP with LMC. Sub-figures left and middle-left correspond to SARCOS; middle-right and right refer to MOCAP7. For each dataset we show the average NELBO convergence of each method and the box-plot trending of the NLPD over the test set across all output. The box-plots at each iteration follow the legend's order from left to right: SGD, Adam, HYB and FNG. The isolated diamonds that appear in the outputs' graphs represent ``outliers".}
	\label{fig:LMC_sarcos_mocap}
\end{figure*}
Figures \ref{fig:human} and \ref{fig:LMC_london_naval} show the
NELBO convergence over the training set, together with the average
NLPD performance over the test set for HUMAN, LONDON and NAVAL data,
respectively. We provide a merged NLPD along outputs for LONDON and NAVAL (see appendix H for an analysis of each specific output). With regard to the rate convergence of NELBO for HUMAN and LONDON datasets all methods converge similarly. Nonetheless, for the NAVAL dataset, our FNG approach presents a faster converge,
followed by HYB and Adam; SGD remains without much progress along the
iterations.
\begin{figure*}[hbtp]
	\centering
	{\hspace{-0.25cm}\includegraphics[width=0.255\textwidth,height=0.19\textheight]{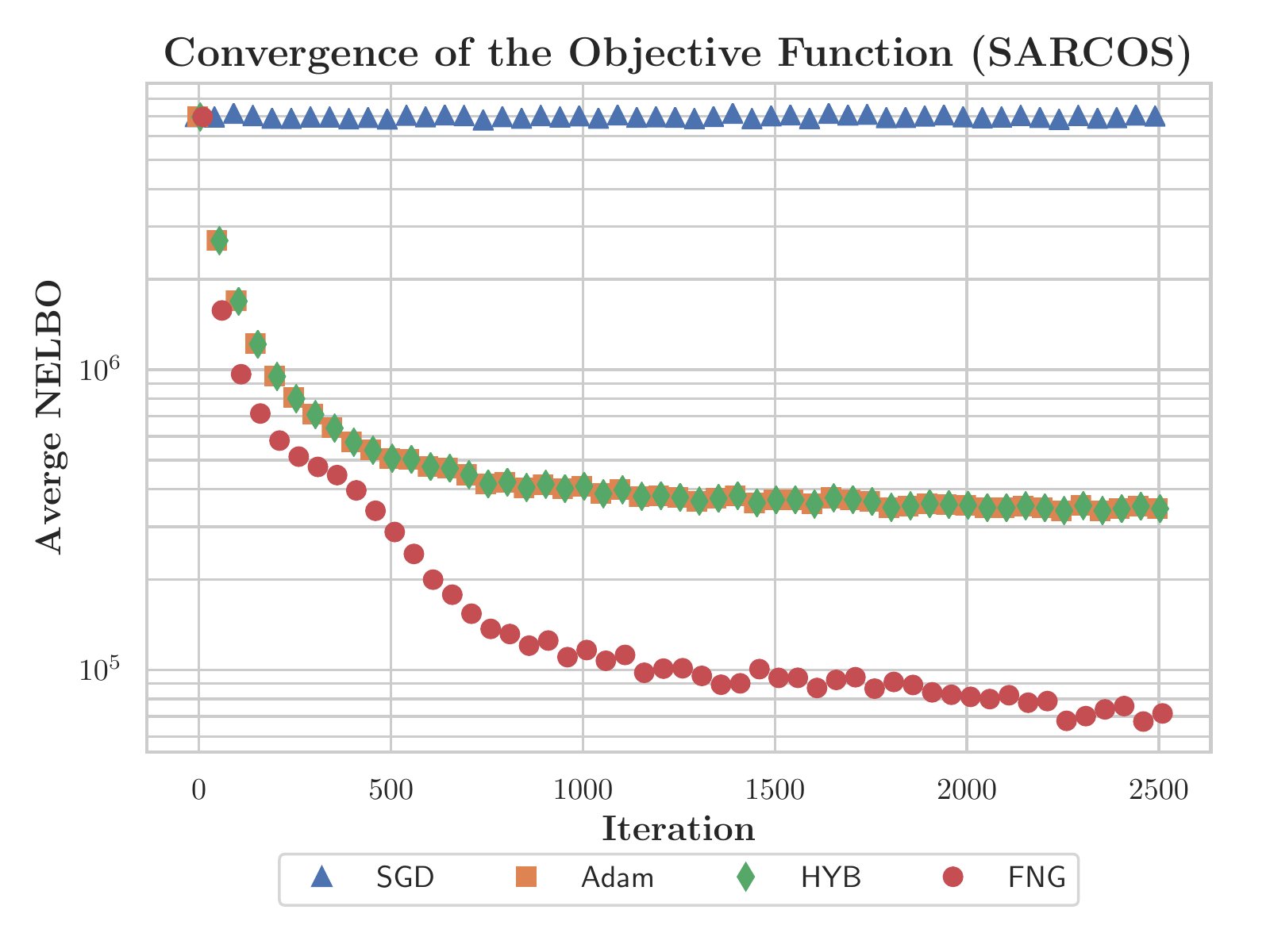}
		\hspace{-0.25cm}\includegraphics[width=0.255\textwidth,height=0.19\textheight]{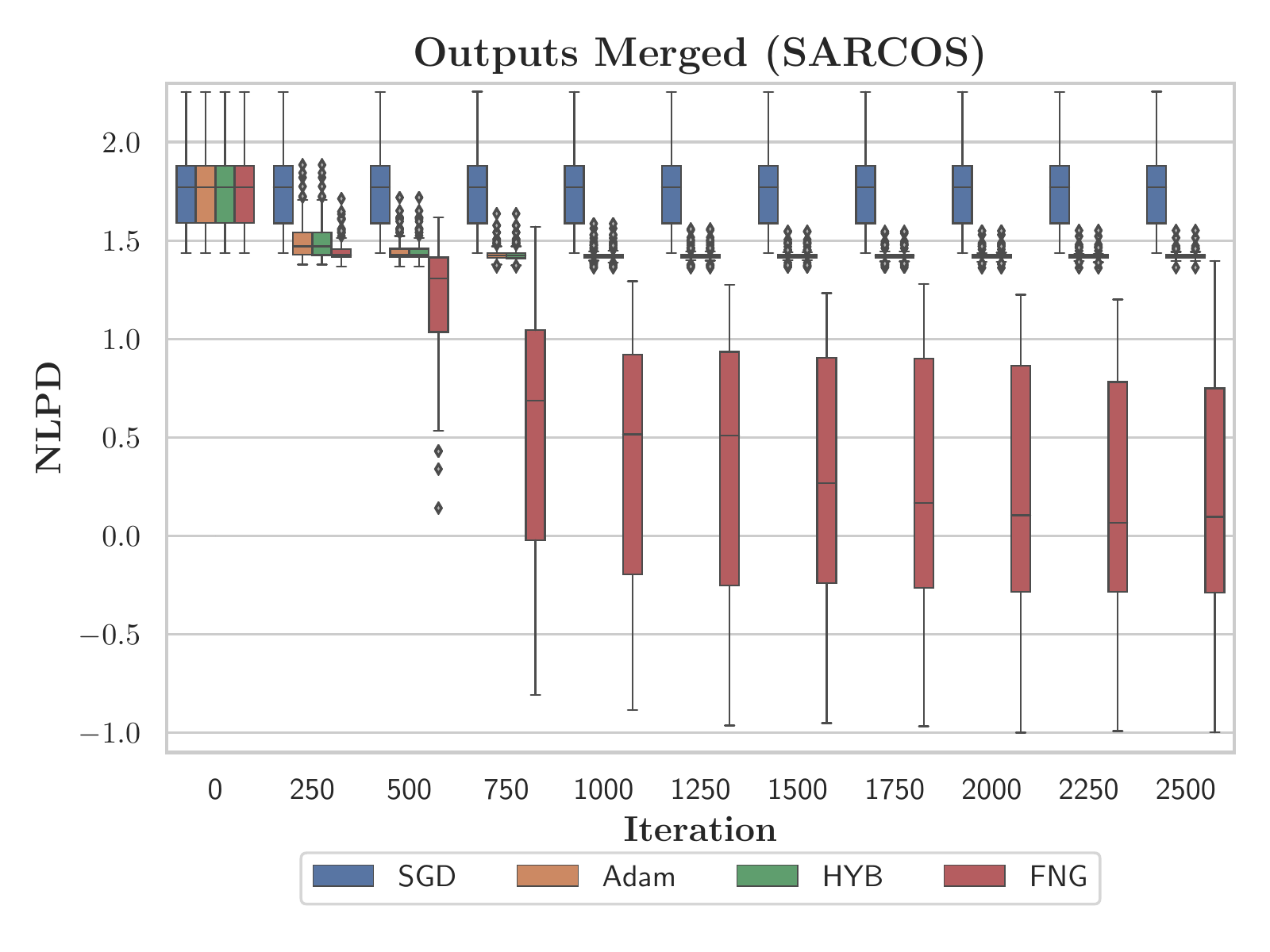}
		\hspace{-0.25cm}\includegraphics[width=0.255\textwidth,height=0.19\textheight]{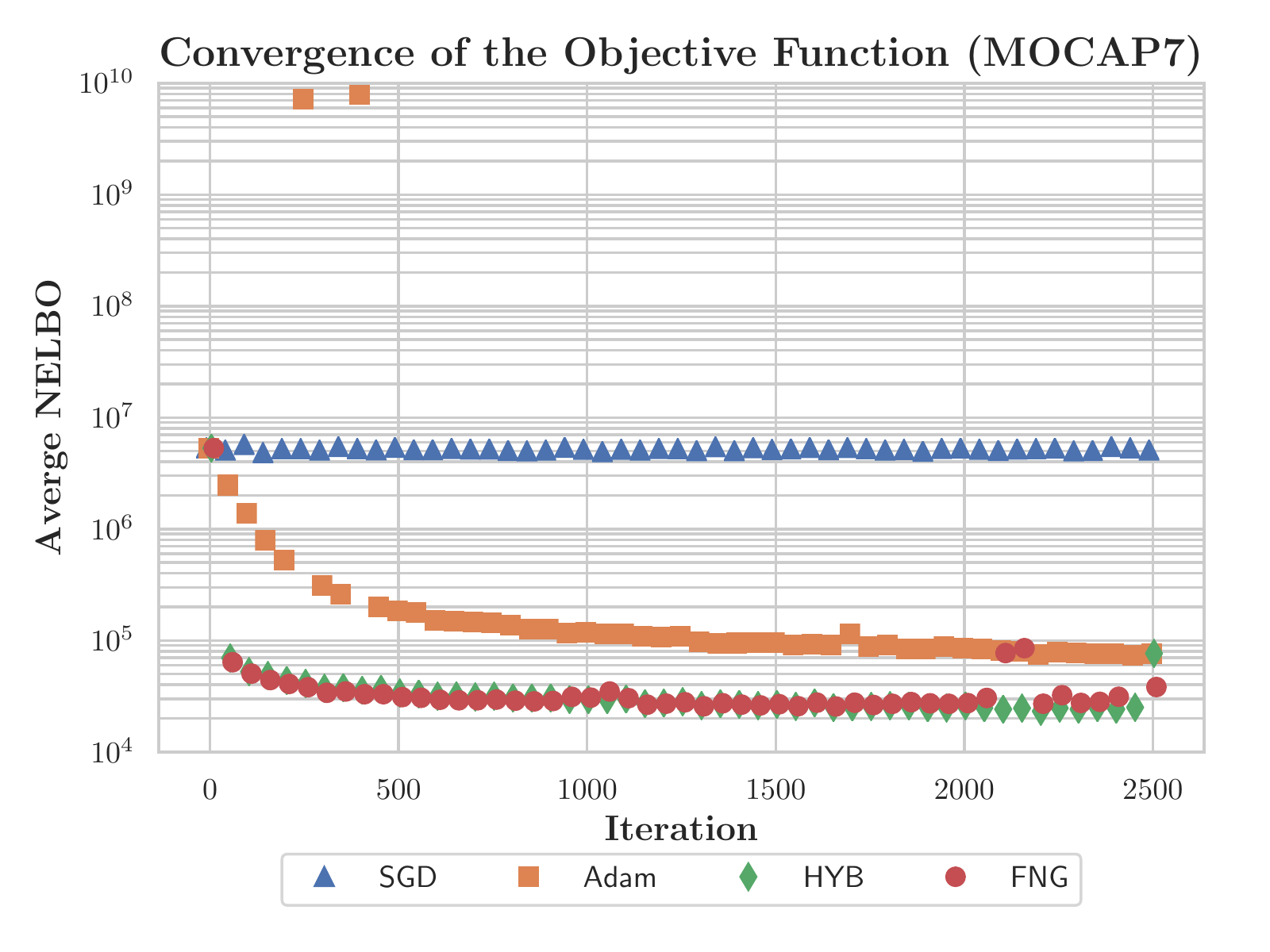}
		\hspace{-0.25cm}\includegraphics[width=0.255\textwidth,height=0.19\textheight]{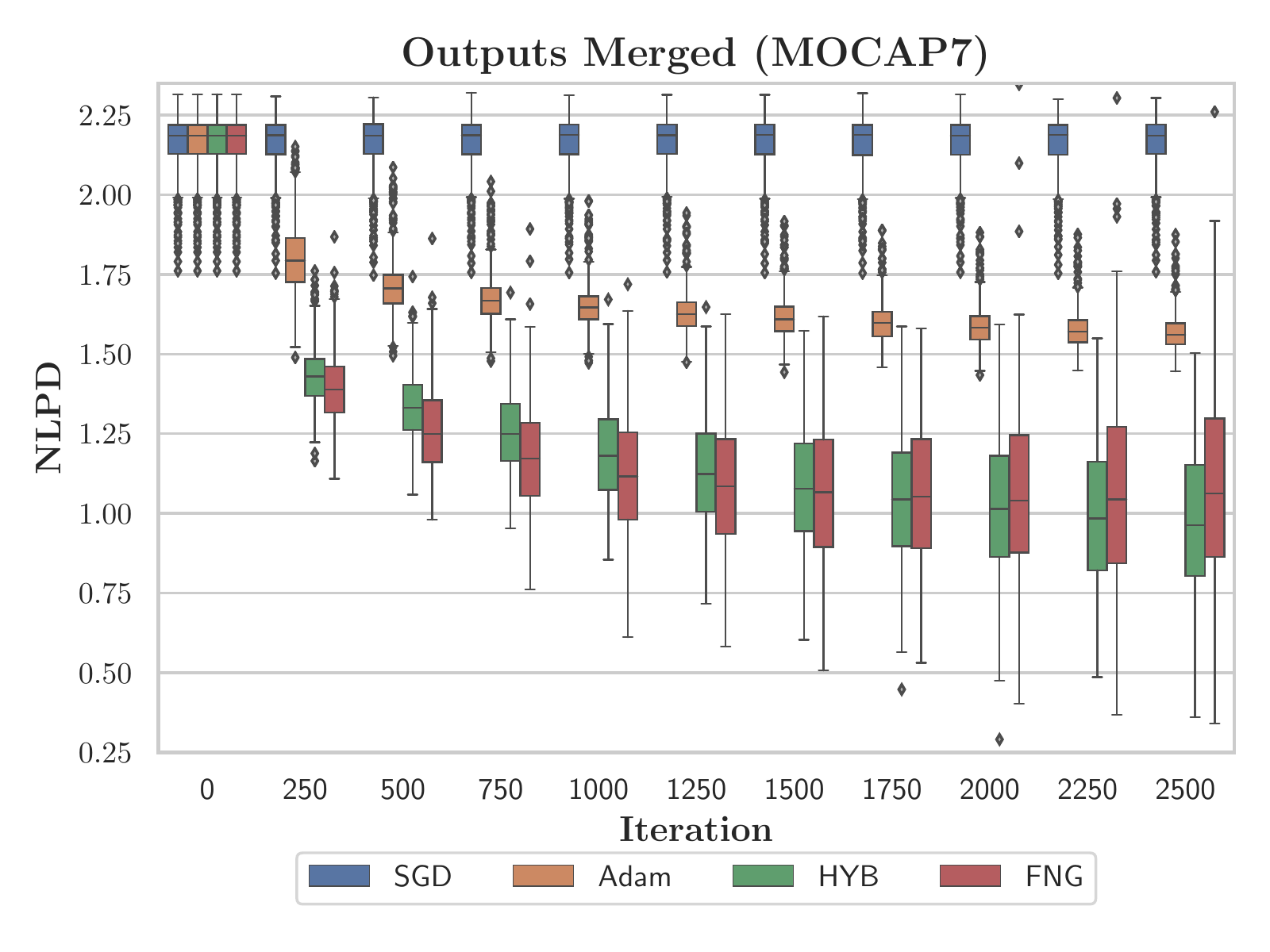}}
	\caption{Performance of the diverse inference methods on the SARCOS and MOCAP7 datasets using 20 different initialisations for HetMOGP with CPM. Sub-figures left and middle-left correspond to SARCOS; middle-right and right refer to MOCAP7. For each dataset we show the average NELBO convergence of each method and the box-plot trending of the NLPD over the test set across all output. The box-plots at each iteration follow the legend's order from left to right: SGD, Adam, HYB and FNG. The isolated diamonds that appear in the outputs' graphs represent ``outliers".}
	\label{fig:CMC_sarcos_mocap}
\end{figure*}
For the HUMAN dataset, the SGD arrives at a better minimum than Adam,
but the Adam's averaged NLPD is higher across outputs. HYB reaches
consistent solutions being better than Adam and SGD, not only in the
training process but also in testing along the HetGaussian and Beta
outputs. Though, the Bernoulli output limits the overall performance
of the method since there is not much improvement along the
iterations. Our FNG method also shows a steady performance along
outputs, commonly arriving to solutions with lower NLPD than the other
methods. Our method presents the biggest variance for the Bernoulli
output, implying strong exploration of the solutions' space for such
likelihood, allowing it to reach the lowest average NLPD.

For the LONDON dataset, Adam converges to a richer minimum of the NELBO
than SGD. Moreover, the NLPD for Adam is, on average, better than the
SGD. The HYB and FNG arrive to a very
similar value of the NELBO, both being better than Adam and SGD. HYB
and FNG methods attain akin NLPD metrics, but the average and median trend of our approach is slightly better, being more robust to the initialisation than HYB method. The NLPD performance for the NAVAL dataset shows in
Fig. \ref{fig:LMC_london_naval} that the SGD method cannot make
progress. We tried to set a bigger step-size, but usually increasing
it derived in numerical problems due to ill-conditioning. The methods
Adam and HYB show similar NLPD boxes, but at the end, Adam attains a slightly lower median with bigger variance than HYB. Regarding the NLPD, our FNG method ends up with a larger variance than SGD, Adam and HYB, but obtaining a much better mean and median trending than the others. Also, our FNG shows that the upper bar of the NLPD box is very close to the interquartile range, while the other methods present larger upper bars, this means that our FNG method concentrates in regions that provide better predictive performance than the other methods.

Fig. \ref{fig:LMC_sarcos_mocap} shows the performance achieved by
the different optimisation methods for SARCOS and MOCAP7
datasets. Since these datasets present a high number of outputs we
stacked the NLPD metric along all outputs. We can notice from the SARCOS experiment, in the first
two sub-figures to the left, that SGD cannot improve much during the
inference process both for NELBO and NLPD. Adam and HYB converge to
the same local minima achieving the same average NELBO and NLPD trend,
in contrast to our FNG method which attains the lowest values showing
a better performance. Particularly in the SARCOS experiment, figures
show how our method changes suddenly, around iteration $600$, probably
escaping from the same local minima to which Adam and HYB
converged. For the MOCAP7's experiment, the two sub-figures to the
right show that SGD slightly improves its performance in the inference
process, while Adam reaches a much better minimum for the average
NELBO. Although, these former methods do not perform better than HYB
and FNG. The HYB and FNG behave similar before 500 iteration, but in
the long term our FNG presents the lowest average NELBO. Likewise, the
NLPD shows that HYB presents a slightly better trend than FNG at the
early stages of the inference, but at the end, our FNG finds a better
NLPD metric.

\subsection{Optimising the HetMOGP with CPM on Real Data}
In this subsection we show the performance of our FNG over the
convolved MOGP for the model with heterogeneous
likelihoods. We use the datasets SARCOS and MOCAP7 with a number of outputs of $D=7$
and $D=40$, respectively. Fig. \ref{fig:CMC_sarcos_mocap} shows the
performance of the different optimisation methods for fitting the
HetMOGP with CPM over such datasets. Similarly to
Fig. \ref{fig:LMC_sarcos_mocap}, we put together the NLPD metric across all outputs. The SARCOS' experiment shows that
SGD does not improve much during the optimisation process. Adam and
HYB seem to converge to a similar minimum value since the average
NELBO and NLPD look very much alike. Otherwise, our FNG method shows
to perform much better than the other methods achieving the lowest
average NELBO. Also the NLPD trend exhibits a more
robust performance over the test set. For
MOCAP7, HYB and FNG behave similarly during the optimisation process
showing almost the same average NELBO trend. Though, the former
method presents a better behaviour when converging at the end. Our FNG
method shows a better NLPD performance during the optimisation, but at
the end HYB reaches a lower NLPD metric. Adam method accomplishes a
poor minima in comparison to HYB and FNG, though a better one than SGD. We can notice from Fig. \ref{fig:LMC_sarcos_mocap} and \ref{fig:CMC_sarcos_mocap}, both experiments over SARCOS and MOCAP7, that the FNG presents similar convergence patterns in both the LMC and CPM, reaching better solutions than SGD, Adam and HYB. The next sub-section compares the performance between these two MOGP prior schemes.   
\subsection{Comparing MOGP priors for heterogeneous likelihoods}

In this subsection we compare the MOGP models for heterogeneous
likelihoods: the one based on the LMC
and the one based on convolution processes. Table \ref{Tab:table1}
presents the different NLPD metrics over a test set when using our
proposed FNG scheme, for the real datasets in previous sections, and we have included two additional datasets for these experiments: TRAFFIC and MOCAP9 (see SM in section VIII for details about these additional datasets). The
Table shows that the CPM in general outperforms the LMC for the
different real datasets used in our experiments. The NLPD performance, for almost all datasets, shows a considerable improvement when using the
convolutional approach, only for MOCAP9 the CPM did not present an improvement over the LMC. The NLPD metric for most of the datasets presents a median very close to the
mean, unlike the HUMAN dataset which its mean differs much to the
median, though having the median a better trend. Also, we can observe
from the Table that generally the standard deviation is higher for the
CPM. This is probably due to the additional hyper-parameter set, i.e.,
the length-scales associated to each smoothing kernel which introduce a bigger
parameters' space to be explored.
\begin{table}[]
	\centering
	\caption{NLPD Performance of the Heterogeneous Schemes.}
	\setlength\tabcolsep{3.0pt}
	\begin{tabular}{c|r|r|r|r|}
		\cline{2-5}
		\multicolumn{1}{l|}{}                  & \multicolumn{2}{c|}{\textbf{LMC}}                              & \multicolumn{2}{c|}{\textbf{CPM}}                              \\ \hline
		\multicolumn{1}{|c|}{\textbf{Dataset}} & \textbf{Median} & \multicolumn{1}{c|}{\textbf{Mean $\pm$ \color{white}{0}\color{black}Std}} & \textbf{Median} & \multicolumn{1}{c|}{\textbf{Mean $\pm$ \color{white}{0}\color{black}Std}} \\ \hline
		\multicolumn{1}{|c|}{LONDON}           & 1.025           & \color{white}{0}\color{black}1.012 $\pm$ \color{white}{0}\color{black}0.331                            & 0.986           & \color{white}{0}\color{black}0.983 $\pm$ \color{white}{0}\color{black}0.396                            \\ \hline
		\multicolumn{1}{|c|}{NAVAL}            & -0.310          & \color{white}{0}\color{black}-0.318 $\pm$ \color{white}{0}\color{black}0.475                           & -0.429          & \color{white}{0}\color{black}-0.454 $\pm$ \color{white}{0}\color{black}0.527                           \\ \hline
		\multicolumn{1}{|c|}{HUMAN}            & 0.596           & \color{white}{0}\color{black}0.646 $\pm$ \color{white}{0}\color{black}0.764                            & 0.330           & \color{white}{0}\color{black}0.529 $\pm$ \color{white}{0}\color{black}0.807                            \\ \hline
		\multicolumn{1}{|c|}{SARCOS}           & 0.684           & \color{white}{0}\color{black}0.618 $\pm$ \color{white}{0}\color{black}0.581                            & 0.096           & \color{white}{0}\color{black}0.169 $\pm$ \color{white}{0}\color{black}0.605                            \\ \hline
		\multicolumn{1}{|c|}{MOCAP9}           & 0.752           & \color{white}{0}\color{black}0.774 $\pm$ \color{white}{0}\color{black}0.297                            & 1.101           & \color{white}{0}\color{black}1.172 $\pm$ \color{white}{0}\color{black}0.386                            \\ \hline
		\multicolumn{1}{|c|}{MOCAP7}           & 1.344           & \color{white}{0}\color{black}1.344 $\pm$ \color{white}{0}\color{black}0.170                            & 1.078           & \color{white}{0}\color{black}1.141 $\pm$ \color{white}{0}\color{black}0.833                            \\ \hline
		\multicolumn{1}{|l|}{TRAFFIC}          & 72.762          & 69.947 $\pm$ 25.466                          & 68.214          & 74.866 $\pm$ 35.775                          \\ \hline
	\end{tabular}
	\label{Tab:table1}
\end{table}

\section{Discussion and Conclusion}

In practice we noticed that some likelihoods (e.g. HetGaussian, Gamma)
tend to strongly influence the value of the objective function
(NELBO), so the optimisers HYB, Adam and SGD are prone to find
solutions that focus on such kind of likelihoods, while neglecting the
others with less influence, for instance a Bernoulli or Beta as shown
in Fig. \ref{fig:toy2}. On the other hand, our proposed scheme presents a more
consistent performance achieving rich solutions across the different
types of outputs' distributions. When increasing the outputs' size our
FNG presented a consistent performance for TOY and real datasets like
SARCOS and MOCAP7. We realised that HYB method presents a relevant
performance for low input dimensionalities, but when the input
dimensionality increases its performance degrades as shown for the TOY experiments when $P>1$ and for the
SARCOS experiment with $P=21$. So, our method
is the least sensible to reduce its performance when increasing the
input dimensionality, followed by the HYB and Adam methods. When using
the SGD method we had to set a very small step-size parameter, because
using large step-sizes makes the model to easily become
ill-conditioned. Also, we observed that our FNG is a suitable scheme for training another type of MOGP model like the CPM. Indeed,
our experiments show that the CPM can also be trained under a
SVI attaining better performance
than a HetMOGP based on a LMC. The new HetMOGP model based on convolution processes differs from the original one based on a LMC in the way the inducing variables are introduced. For the LMC the inducing variables are additional evaluations of the functions $u_{q}(\cdot)$. While for the CPM the inducing variables are additional evaluations of the functions $f_{d,j}(\cdot)$. We implemented the version of CPM using the same style of inducing variables as the LMC, though in the practice we realised that the assumption commonly used in the literature for the posterior, i.e., $q(\mathbf{f},\mathbf{\check{u}})=p(\mathbf{f}|\mathbf{\check{u}})q(\mathbf{\check{u}})$ is not sufficiently flexible to fit the LPFs and limits the SVI implementation. Therefore, we opted for the inducing variables procedure which does support the assumption $q(\mathbf{f},\mathbf{\check{u}})=p(\mathbf{f}|\mathbf{\check{u}})q(\mathbf{\check{u}})$.

The VO bound in Eq. \eqref{eq:new_ELBO} can be seen as a fully Bayesian treatment of the HetMOGP, where the model's parameters and hyper-parameters follow a prior distribution, where the positive constraint variables follow a Log-Normal distribution and the non-constraint ones follow a Gaussian distribution. Our VO bound benefits from the assumption of a Gaussian exploratory (or posterior) distribution for deriving in a closed-form our FNG optimisation scheme. This scheme helps to find solutions that directly improve the predictive capabilities of the HetMOGP model. For instance, since the inducing points' size is directly influenced by the input dimensionality, we believe that applying exploration over them helps to improve the model performance for high input dimensionalities as shown in the experiments.

In this paper, we have shown how a fully natural gradient scheme
improves optimisation of a heterogeneous MOGP model by generally
reaching better local optima solutions with higher test performance
rates than HYB, Adam and SGD methods. We have shown that our FNG
scheme provides rich local optima solutions, even when increasing the
dimensionality of the input and/or output space. Furthermore, we have
provided a novel extension of a stochastic scalable Heterogeneous MOGP
model based on convolution processes. Our FNG method may also
be an alternative tool for improving optimisation over a single output
GP model. As a future work, it might be worth exploring the behaviour of
the proposed scheme over other type of GP models, for instance Deep
GPs \citep{salimbeni2019}. Likewise, it would be relevant to explore a scalable way to
implement the method using a full covariance matrix
$\boldsymbol{\Sigma}$ which can exploit full correlation between all
hyper-parameters. Modelling multi-modal data is another venue
for future work. One might potentially want to combine ideas
from the work in \citep{Lazaro2012}, with the HetMOGP model and
the optimisation schemes proposed in this work. Also, ideas for the model selection problem of the number $Q$ of latent functions, like the ones based on Indian buffet processes \citep{Guarnizo2015,Tong2019} can be further investigated in the particular context of MOGPs with Heterogeneous outputs.

\section*{Acknowledgment}
The authors would like to thank Emtiyaz Khan for the feedback
related to mathematical aspects of the method and to the authors of \citep{pablo2018} for lending their datasets HUMAN
and LONDON. The authors would also like to thank Innovate UK for funding under the project 104316. JJG is being funded by a scholarship
from the Dept. of Comp. Science, University of Sheffield. MAA
has been financed by the EPSRC Research Projects EP/R034303/1 and
EP/T00343X/1.

\small
\bibliographystyle{plainnat} 
\bibliography{biblioJournal2019}

\begin{appendices}
	\section{Variational Optimisation Example}
	
	It is important to highlight that the idea of exploration is better understood in the context of variational optimisation for optimising a deterministic variable. Hence, with the purpose of
	illustrating such an idea of introducing exploration over
	a deterministic variable through a variational optimisation scheme,
	we build an example similar to one in the work ``Variational Adaptive-Newton Method for Explorative Learning" (M. E. Khan, W. Lin, V. Tangkaratt, Z. Liu, and D. Nielsen (2017)). Let us define a function with
	multiple local minima, $g(\theta)=2\exp(-0.09\theta^2)\sin(4.5\theta)$, which we
	are interested in optimising:
	\begin{align}
	\theta^{\star}=\arg \min_{\theta} g(\theta)=\arg \min_{\theta} 2\exp(-0.09\theta^2)\sin(4.5\theta).\label{eq:objective_g}
	\end{align}
	We intend to introduce exploration over the variable $\theta$ by
	assuming it as stochastic. To this end, we can solve an analogous
	optimisation problem that consists of minimising a bound $\mathcal{\tilde{L}}$ as follows: 
	\begin{align}
	\min_{\theta} g(\theta) \leq	\mathcal{\tilde{L}} =\mathbb{E}_{q(\theta)}[g(\theta)],
	\label{eq:objective_L}
	\end{align}
	\begin{figure}[h]
		\centering
		{\includegraphics[width=0.5\textwidth]{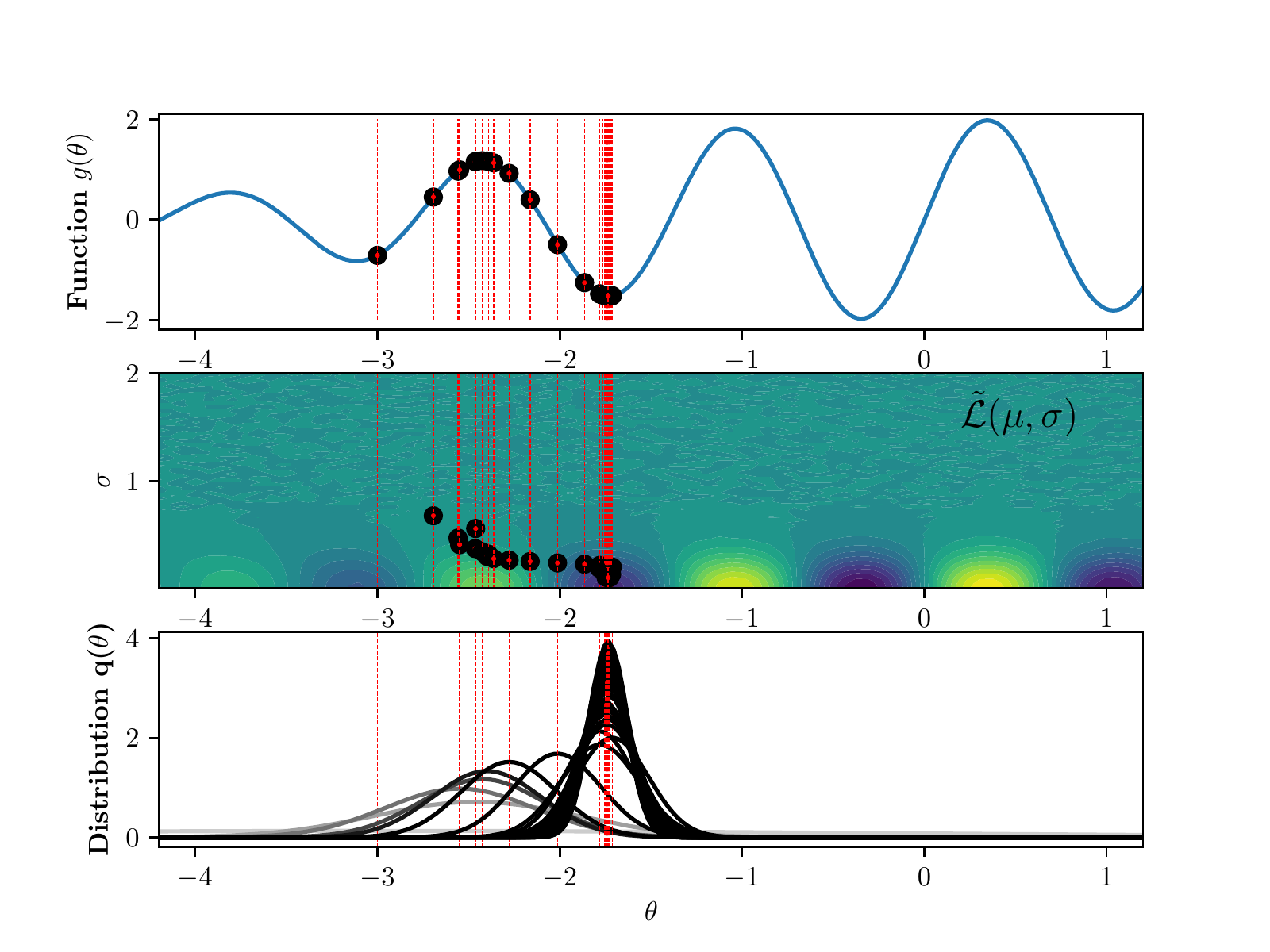}}
		\caption{Variational Optimisation for the function $g(\theta)=2\exp(-0.09\theta^2)\sin(4.5\theta)$.}
		\label{fig:VO}
	\end{figure}
	w.r.t the distribution $q(\theta)$, where this
	$q(\theta):=\mathcal{N}(\theta|\mu,\sigma^2)$ represents a variational
	distribution over $\theta$, with parameters mean $\mu$ and variance
	$\sigma^2$. We built an experiment to show what happens at
	each iteration of the optimisation process. Figure \ref{fig:VO} shows three
	perspectives of a such experiment, where we initialise the parameters
	$\theta=\mu=-3.0$ and $\sigma=3.0$. First row in Figure \ref{fig:VO} shows what
	happens from the perspective of the original function $g(\theta)$, the
	black dots on the graph represent the position of $\theta=\mu$ at each
	iteration. Second row shows a contour graph that let us visualise the
	space of solutions for the bound $\mathcal{\tilde{L}}$ for different
	values of $\sigma$ and $\mu$, and how these parameters change along
	the inference process. Here the black dots refer to the position
	achieved by $\sigma$ and $\mu$ at each iteration, and the colour
	intensities change from blue to yellow relating to lower and higher
	values of $\mathcal{\tilde{L}}$. Third row shows the behaviour of the
	exploratory Gaussian distribution $q(\theta)$. For each Gaussian bell,
	we use a colour code from light-gray to black for representing initial
	to final stages of the inference. All sub-graphs in Figure \ref{fig:VO} present
	vertical lines in red colour, these lines intend to align all
	sub-graphs regarding iterations, i.e., from left to right the vertical
	lines represent the occurrence of an iteration, being the furthest to
	the left the initial one. In order to avoid excessive overlapping of
	many graphs, the third row in the Figure \ref{fig:VO} only shows the plots every
	two iterations. For the inference we used the natural gradient updates
	shown in the paper:
	\begin{align}
	\sigma^{-2}_{t+1}&=\sigma^{-2}_{t}+2\alpha_t\hat{\nabla}_{\sigma^{2}}\mathcal{\tilde{L}}_t\notag\\
	\mu_{t+1}&=\mu_{t}-\alpha_t\sigma^{2}_{t+1}\hat{\nabla}_{\mu}\mathcal{\tilde{L}}_t,\notag
	\end{align} 
	where $\mu_t$ and $\sigma^2_t$ are the mean and variance parameters at the instant $t$ respectively; the stochastic gradients are  $\hat{\nabla}_{\mu}\mathcal{\tilde{L}}_{t}:=\hat{\nabla}_{\mu}\mathcal{\tilde{L}}(\mu_t,\sigma_t)$ and $\hat{\nabla}_{\sigma^2}\mathcal{\tilde{L}}_{t}:=\hat{\nabla}_{\sigma^2}\mathcal{\tilde{L}}(\mu_t,\sigma_t)$.
	We can notice from Figure \ref{fig:VO} that the initial value of $\theta=\mu=-3.0$ is close the the poor minimum at $\theta\approx -3.114$ and far away from better minima solutions like the one at $\theta\approx -1.729$ and $\theta\approx -0.346$ (the global minimum). When the inference process starts, the exploratory distribution $q(\theta)$ modifies its variance and moves its mean towards a better region in the space of $\theta$. From the third row we can also see that $q(\theta)$ initially behaves as a broad distribution (in light-gray colour) with a mean located at $\mu=-3.0$, while the iterations elapse, the distribution $q(\theta)$ modifies its shape in order to reach a better local minima solution (at $\mu\approx -1.729$). The distribution presents such behaviour in spite of being closer to other poor local minima like the one between the interval $(-4,-3)$. Additionally, when the mean $\mu$ is close to the local minimum at $\theta\approx -1.729$, the variance parameter reduces constantly making the distribution look narrower, which means the variance parameter tends to collapse to zero ($\sigma^2\rightarrow 0$) increasing the certainty of the solution. This behaviour implies that in the long term the distribution will become a Dirac's delta $q(\theta)=\delta(\theta-\mu)$, where $\mu=\theta$. Therefore, a feasible minima solution for the original objective function $g(\theta)$ is $\theta=\mathbb{E}_{q(\theta)}[\theta]=\mu$. This can be seen in the first sub-graph where at each iteration $\theta=\mu$, in fact, at the end $\mu$ is fairly close to the value $\theta\approx -1.729$, a local minima. Though we could notice, in Figure \ref{fig:VO}, an exploratory behaviour of $q(\theta)$ that helped avoiding the poor local minima at $\theta\approx -3.114$, the rapid collapsing effect of the variance parameter limits the exploration of $\theta$'s space. In order to reduce such a collapsing effect of $q(\theta)$ and gain additional exploration, we can take advantage of a Kullback-Leibler (KL) diverge to penalise the Eq. \eqref{eq:objective_L} as follows, 
	\begin{align}
	\min_{\theta} g(\theta) \leq	\mathcal{\tilde{L}} =\mathbb{E}_{q(\theta)}[g(\theta)]+\mathbb{D}_{KL}(q(\theta)||p(\theta)),\notag
	\end{align}
	where we force the exploratory distribution $q(\theta)$ to trade-off between minimising the expectation $\mathbb{E}_{q(\theta)}[g(\theta)]$ and not going far away from an imposed $p(\theta)$ penalization (M. E. Khan, Z. Liu, V. Tangkaratt, and Y. Gal (2017)).\medskip\\
	
\begin{figure}[]
		\centering
		{\includegraphics[width=0.5\textwidth]{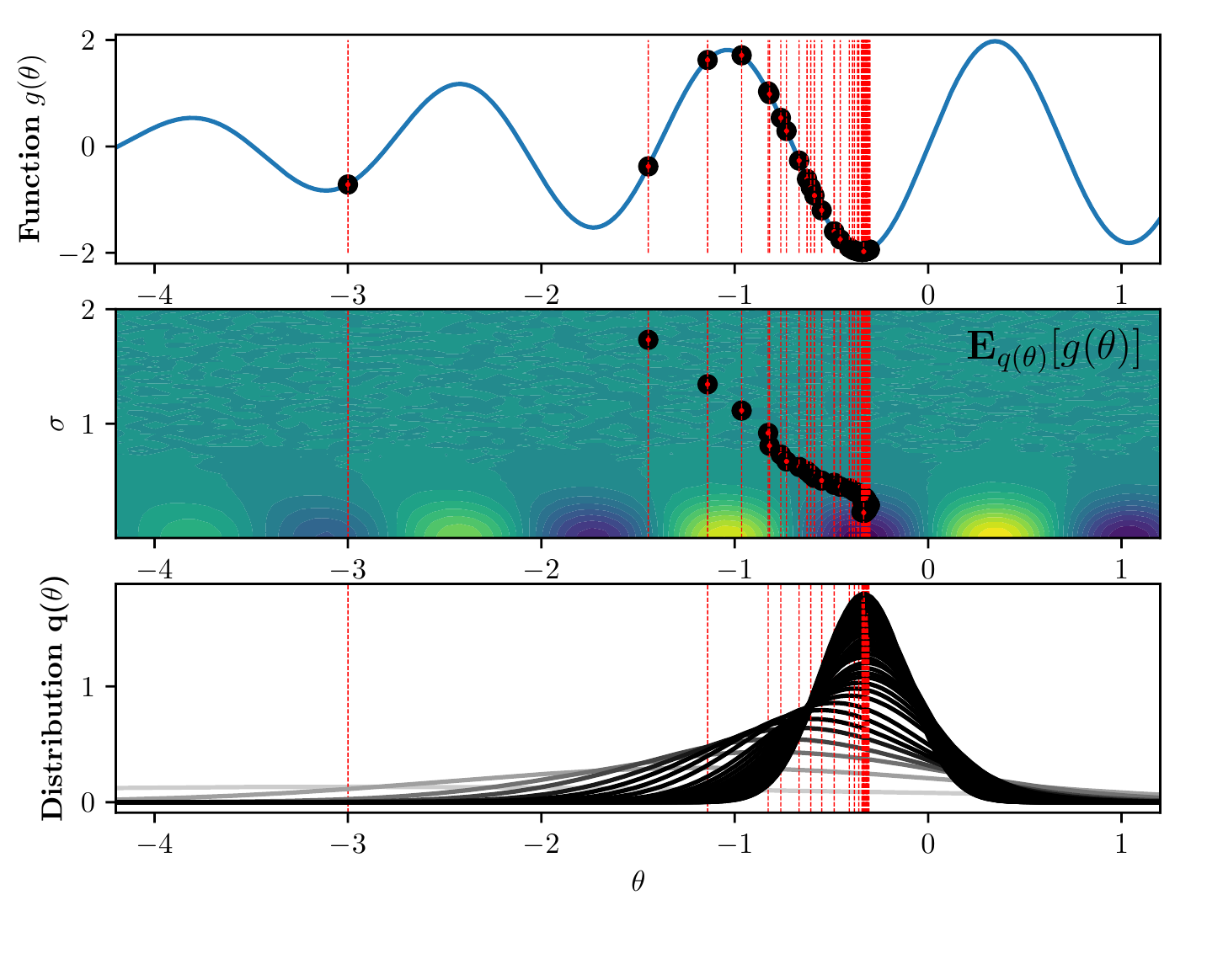}}
		\caption{Variational Optimisation with KL penalisation for the function  $g(\theta)=2\exp(-0.09\theta^2)\sin(4.5\theta)$.}
		\label{fig:VO_KL}
	\end{figure}

\begin{figure}[]
		\centering
		{\includegraphics[width=0.5\textwidth]{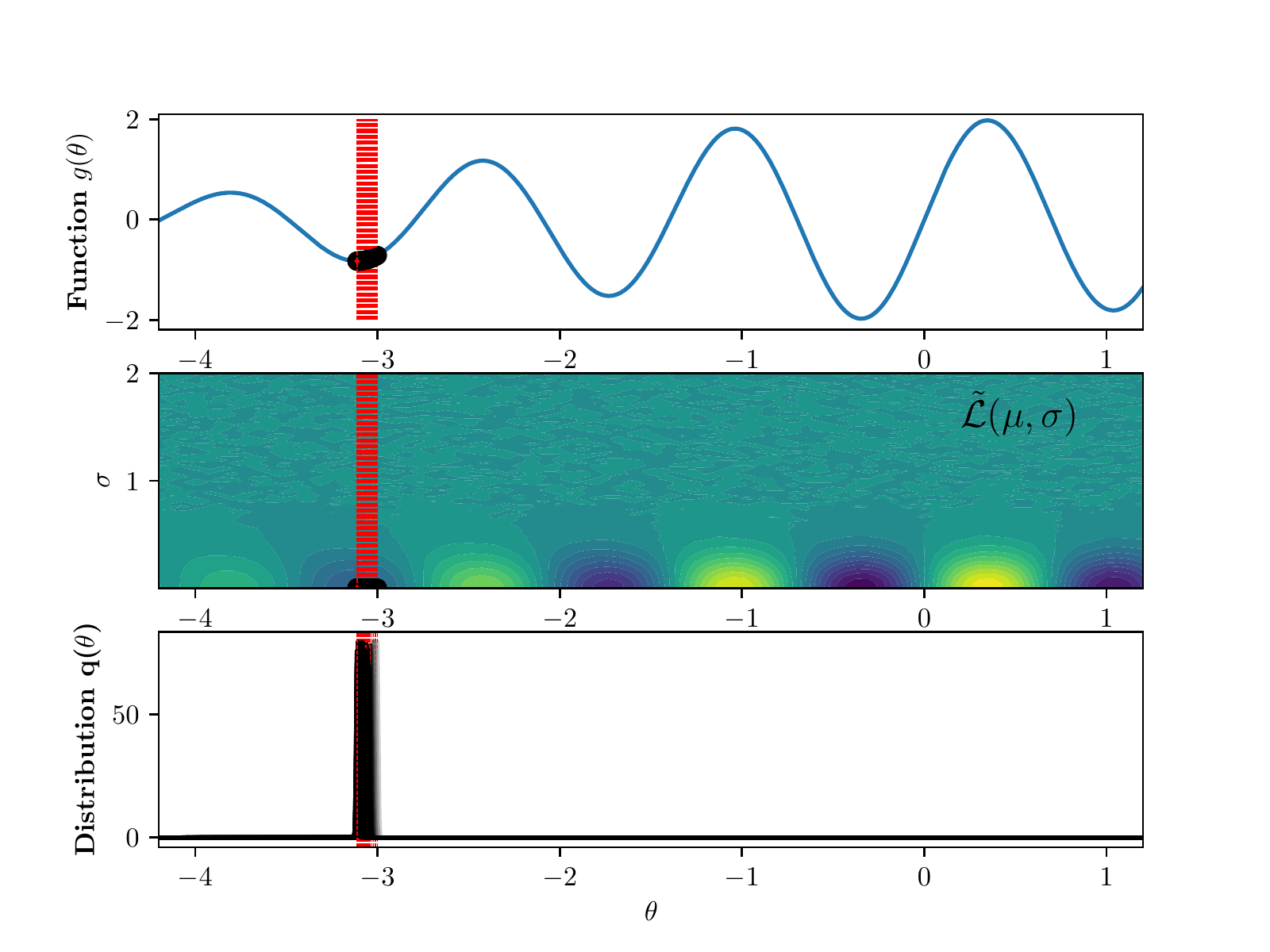}}
		\caption{Using Newton's method for optimising the function  $g(\theta)=2\exp(-0.09\theta^2)\sin(4.5\theta)$.}
		\label{fig:Newton}
	\end{figure}

	Indeed, our work makes use of the KL divergence, which represents a variational optimisation case that introduces a penalisation that avoids the variational distribution's variance to collapse. For instance, we can look at such effect following again
	the example of  $g(\theta)=2\exp(-0.09\theta^2)\sin(4.5\theta)$. Figure \ref{fig:VO_KL} shows the
	inference process behaviour when we apply the KL divergence, we used a
	penalisation (or prior) distribution
	$p(\theta)=\mathcal{N}(0,\lambda^{-1})$, with $\lambda=1.5$. Similar to
	Figure \ref{fig:VO}, the initial value of $\theta=\mu=-3.0$. When the inference
	process starts, the distribution $q(\theta)$ moves its mean in
	direction to a better region in the space of
	$\theta$. Notice that, in second row of
	Figure \ref{fig:VO_KL}, we refer to $\mathbb{E}_{q(\theta)}[g(\theta)]$ instead of
	$\mathcal{\tilde{L}}$ in order to ease the comparison to Figure
	1. From the third row we can also see that $q(\theta)$ initially
	behaves as a broad distribution (in light-gray color) with a mean
	located at $\mu=-3.0$. While the iterations elapse, the distribution
	$q(\theta)$ modifies its shape in order to reach the solution at
	$\mu\approx -0.346$. When the mean $\mu$ is close to $\theta\approx -0.346$ (the global
	minimum), the variance parameter $\sigma^2$ decreases significantly in comparison to
	its initial stage, though we realise that its collapsing
	effect widely reduces in comparison to the case without KL divergence of Figure \ref{fig:VO}. Also, we can notice that there is a faster convergence in comparison to the case without KL divergence.\medskip\\ 
	On the other hand, we reproduced the same experiment without
	introducing any exploration over the variable $\theta$. We used the
	Newton's method for minimising Eq. \eqref{eq:objective_g}, with the
	same initial point $\theta=-3.0$. Figure \ref{fig:Newton} presents the process of
	convergence of the Newton's method, this figure uses the same style
	used for Figure \ref{fig:VO}. Though here, from right to left the vertical red
	lines represent the occurrence of an iteration, being the furthest to
	the right the initial one. As it can be seen from
	Figure \ref{fig:Newton}, the optimisation carried out in the space of $g(\theta)$,
	with no exploration mechanism, converges to a poor local minima
	located between the interval $(-4,-3)$. Also, we can analogously view
	the Eq. \eqref{eq:objective_L} as:
	\begin{align}
	g(\theta)=\mathcal{\tilde{L}}=\mathbb{E}_{q(\theta)}[g(\theta)],\notag
	\end{align}
	where this equality holds if the distribution is a Dirac's delta
	$q(\theta)=\delta(\theta-\mu)$ and $\mu=\theta$. Indeed, we can think
	of $q(\theta)$ as a Gaussian distribution with its variance collapsed
	to zero ($\sigma^2=0$), that is why the second sub-graph in Figure \ref{fig:Newton}
	shows the black dots only moving along $\theta$-axis (where
	$\theta=\mu$) while $\sigma=0$, and the third sub-graph depicts the
	Dirac's delta distributions.  
	We have shown through this example how variational optimisation helps to induce exploration over a deterministic variable avoiding poor local minima solutions, we compared the VO approaches with and without KL penalisation, both introduce exploration during the inference process, but the former allows additional exploration as shown in the previous experiments.
	
	\section{Details on: From Mirror Descent to the Natural-Gradient}
	The mirror descent algorithm in the mean-parameters space of the distribution $q(\boldsymbol{\theta})$ bases on solving the following iterative sub-problems: 
	\begin{align}
	\boldsymbol{\eta}_{t+1}=\arg \min_{\boldsymbol{\eta}} \langle\boldsymbol{\eta},\hat{\nabla}_{\boldsymbol{\eta}}\tilde{\mathcal{L}}_{t}\rangle+\frac{1}{\alpha_t}\mathbb{D}_{KL}(q(\boldsymbol{\theta})||q_t(\boldsymbol{\theta})).
	\notag
	\end{align}  
	The intention of the above formulation is to exploit the parametrised distribution's structure by controlling its divergence w.r.t its older state $q_t(\boldsymbol{\theta})$. Thus,  we can solve for the mirror descent algorithms setting to zero, 
	\begin{align}
	\langle\boldsymbol{\eta},\hat{\nabla}_{\boldsymbol{\eta}}\tilde{\mathcal{L}}_{t}\rangle+\frac{1}{\alpha_t}\mathbb{D}_{KL}(q(\boldsymbol{\theta})||q_t(\boldsymbol{\theta}))=0\notag\\
	\langle\boldsymbol{\eta},\hat{\nabla}_{\boldsymbol{\eta}}\tilde{\mathcal{L}}_{t}\rangle+\frac{1}{\alpha_t}\mathbb{E}_{q(\boldsymbol{\theta})}[\log q(\boldsymbol{\theta})-\log q_t(\boldsymbol{\theta})]=0,\notag
	\end{align}
	since we can express the distribution $q(\boldsymbol{\theta})$ in the exponential-family form as follows:
	\begin{align}
	q(\boldsymbol{\theta})=h(\boldsymbol{\theta})\exp\big(\langle\boldsymbol{\lambda},\phi(\boldsymbol{\theta})\rangle-A(\boldsymbol{\lambda})\big)\notag
	\end{align}
	we can replace it in the above KL divergence as,
	\begin{align}
	\langle\boldsymbol{\eta},\hat{\nabla}_{\boldsymbol{\eta}}\tilde{\mathcal{L}}_{t}\rangle+\frac{1}{\alpha_t}\mathbb{E}_{q(\boldsymbol{\theta})}\big[\langle\boldsymbol{\lambda},\phi(\boldsymbol{\theta})\rangle-A(\boldsymbol{\lambda})\big]-\frac{1}{\alpha_t}\mathbb{E}_{q(\boldsymbol{\theta})}\big[\langle\boldsymbol{\lambda}_t,\phi(\boldsymbol{\theta})\rangle-A(\boldsymbol{\lambda}_t)\big]=0\notag
	\end{align}
	\begin{align}
	\langle\boldsymbol{\eta},\hat{\nabla}_{\boldsymbol{\eta}}\tilde{\mathcal{L}}_{t}\rangle+\frac{1}{\alpha_t}\big[\langle\boldsymbol{\lambda},\mathbb{E}_{q(\boldsymbol{\theta})}[\phi(\boldsymbol{\theta})]\rangle-A(\boldsymbol{\lambda})\big]-\frac{1}{\alpha_t}\big[\langle\boldsymbol{\lambda}_t,\mathbb{E}_{q(\boldsymbol{\theta})}[\phi(\boldsymbol{\theta})]\rangle-A(\boldsymbol{\lambda}_t)\big]=0,\notag
	\end{align}
	given that $\mathbb{E}_{q(\boldsymbol{\theta})}[\phi(\boldsymbol{\theta})]=\boldsymbol{\eta}$ represent the mean-parameters, we can write again,
	\begin{align}
	\langle\boldsymbol{\eta},\hat{\nabla}_{\boldsymbol{\eta}}\tilde{\mathcal{L}}_{t}\rangle+\frac{1}{\alpha_t}\big[\langle\boldsymbol{\lambda},\boldsymbol{\eta}\rangle-A(\boldsymbol{\lambda})-\langle\boldsymbol{\lambda}_t,\boldsymbol{\eta}\rangle+A(\boldsymbol{\lambda}_t)\big]=0\notag
	\end{align}
	deriving w.r.t $\boldsymbol{\eta}$ we arrive to:
	\begin{align}
	\hat{\nabla}_{\boldsymbol{\eta}}\tilde{\mathcal{L}}_{t}+\frac{1}{\alpha_t}\big[\boldsymbol{\lambda}-\boldsymbol{\lambda}_t\big]=0.\notag
	\end{align}
	Where the recursive update comes from making $\boldsymbol{\lambda}:=\boldsymbol{\lambda}_{t+1}$:
	\begin{align}
	\boldsymbol{\lambda}_{t+1}&=\boldsymbol{\lambda}_{t}-\alpha_t\hat{\nabla}_{\boldsymbol{\eta}}\tilde{\mathcal{L}}_{t}\notag
	\end{align}
	where $\hat{\nabla}_{\boldsymbol{\eta}}\tilde{\mathcal{L}}_{t}=\mathbf{F}^{-1}\hat{\nabla}_{\boldsymbol{\lambda}}\tilde{\mathcal{L}}_{t}$ as per the work “The information geometry of mirror descent,” (G. Raskutti and S. Mukherjee (2015)), where the authors provide a formal proof of such equivalence.
	\section{Details on Bound Derivation for HetMOGP with Linear Model of Coregionalisation}
	We build the objective ELBO for the linear model of coregionalisation by assuming a variational distribution $q(\mathbf{f},\mathbf{u})=p(\mathbf{f}|\mathbf{u})q(\mathbf{u})$ as follows: 
	\begin{align} 
	\mathcal{L}&=\mathbb{E}_{q(\mathbf{f},\mathbf{u})}\Big[\log \frac{p(\mathbf{y}|\mathbf{f})p(\mathbf{f}|\mathbf{u})p(\mathbf{u})}{q(\mathbf{f},\mathbf{u})}\Big]\notag\\
	&=\mathbb{E}_{p(\mathbf{f}|\mathbf{u})q(\mathbf{u})}\Big[\log \frac{p(\mathbf{y}|\mathbf{f})\cancel{p(\mathbf{f}|\mathbf{u})}p(\mathbf{u})}{\cancel{p(\mathbf{f}|\mathbf{u})}q(\mathbf{u})}\Big]\notag\\
	&=\mathbb{E}_{p(\mathbf{f}|\mathbf{u})q(\mathbf{u})}\Big[\log p(\mathbf{y}|\mathbf{f})\Big]+\mathbb{E}_{q(\mathbf{u})}\Big[\log \frac{p(\mathbf{u})}{q(\mathbf{u})}\Big].\notag
	%\label{eq:ELBO}
	\end{align}
	Notice that the right hand side term in the equation above does not depend on $p(\mathbf{f}|\mathbf{u})$ then only $q(\mathbf{u})$ remains. The left hand side term does not depend on $\mathbf{u}$ so we can integrate it out as follows:
	\begin{align}
	q(\mathbf{f}_{d,j})&=\int p(\mathbf{f}_{d,j}|\mathbf{u})q(\mathbf{u})d\mathbf{u},\notag
	\end{align}
	with this result, the marginal posterior over all the latent parameter functions is build as,
	\begin{align}
	q(\mathbf{f})=\prod_{d=1}^{D}\prod_{j=1}^{Jd}q(\mathbf{f}_{d,j}),\notag
	\end{align}
	this way we can keep developing the ELBO,
	\begin{align}
	\mathcal{L}&=\mathbb{E}_{q(\mathbf{f})}\Big[\log p(\mathbf{y}|\mathbf{f})\Big]+\mathbb{E}_{q(\mathbf{u})}\Big[\log \prod_{q=1}^{Q}\frac{p(\mathbf{u}_q)}{q(\mathbf{u}_q)}\Big]\notag\\
	&=\mathbb{E}_{q(\mathbf{f})}\Big[\log \prod_{d=1}^{D}\prod_{n=1}^{N}p(y_{d,n}|\psi_{d,1}(\mathbf{x}_n),...,\psi_{d,J_d}(\mathbf{x}_n))\Big]+\mathbb{E}_{q(\mathbf{u})}\Big[\log \prod_{q=1}^{Q}\frac{p(\mathbf{u}_q)}{q(\mathbf{u}_q)}\Big]\notag\\
	&=\sum_{d=1}^{D}\sum_{n=1}^{N}\mathbb{E}_{q(\mathbf{f})}\Big[\log p(y_{d,n}|\psi_{d,1}(\mathbf{x}_n),...,\psi_{d,J_d}(\mathbf{x}_n))\Big]-\sum_{q=1}^{Q}\mathbb{D}_{\text{KL}}(q(\mathbf{u}_q)||p(\mathbf{u}_q)).\notag    
	\end{align}
	We write again as a negative ELBO:
	\begin{align}
	\mathcal{\tilde{L}}=\sum_{n=1}^{N}\sum_{d=1}^{D}\mathbb{E}_{q(\mathbf{f}_{d,1}) \cdots q(\mathbf{f}_{d,J_d})}\left[g_{d,n}\right]+\sum_{q=1}^{Q} \mathbb{D}_{KL}\left(q(\mathbf{u}_{q}) \| p(\mathbf{u}_{q})\right),\label{eq:HetMOGP_appendix}
	\end{align}
	where $g_{d,n}=-\log p(y_{d,n}|\psi_{d,1}(\mathbf{x}_n),...,\psi_{d,J_d}(\mathbf{x}_n))$ is the NLL function associated to each output.
	
	\section{Details on Bound Derivation for HetMOGP with Convolution Processes}
	We derive the ELBO for the Heterogeneous MOGP with convolution processes, assuming a variational distribution $q(\mathbf{f},\mathbf{\check{u}})=p(\mathbf{f}|\mathbf{\check{u}})q(\mathbf{\check{u}})$ as follows: 
	\begin{align} 
	\mathcal{L}&=\mathbb{E}_{q(\mathbf{f},\mathbf{\check{u}})}\Big[\log \frac{p(\mathbf{y}|\mathbf{f})p(\mathbf{f}|\mathbf{\check{u}})p(\mathbf{\check{u}})}{q(\mathbf{f},\mathbf{\check{u}})}\Big]\notag\\
	&=\mathbb{E}_{p(\mathbf{f}|\mathbf{\check{u}})q(\mathbf{\check{u}})}\Big[\log \frac{p(\mathbf{y}|\mathbf{f})\cancel{p(\mathbf{f}|\mathbf{\check{u}})}p(\mathbf{\check{u}})}{\cancel{p(\mathbf{f}|\mathbf{\check{u}})}q(\mathbf{\check{u}})}\Big]\notag\\
	&=\mathbb{E}_{p(\mathbf{f}|\mathbf{\check{u}})q(\mathbf{\check{u}})}\Big[\log p(\mathbf{y}|\mathbf{f})\Big]+\mathbb{E}_{q(\mathbf{\check{u}})}\Big[\log \frac{p(\mathbf{\check{u}})}{q(\mathbf{\check{u}})}\Big].\notag
	%\label{eq:ELBO}
	\end{align}
	Since the right hand side term in the equation above does not depend on $p(\mathbf{f}|\mathbf{\check{u}})$ then only $q(\mathbf{\check{u}})$ remains in the expectation. Regarding the left hand side term, $p(\mathbf{y}|\mathbf{f})$ does not depend on $\mathbf{\check{u}}$, so we can integrate out $q(\mathbf{\check{u}})$, as follows:
	\begin{align}
	q(\mathbf{f})&=\int p(\mathbf{f}|\mathbf{\check{u}})q(\mathbf{\check{u}})d\mathbf{\check{u}}.\notag\\
	&=\int\prod_{d=1}^{D}\prod_{j=1}^{Jd} p(\mathbf{f}_{d,j}|\mathbf{\check{u}}_{d,j})q(\mathbf{\check{u}}_{d,j})d\mathbf{\check{u}}_{d,j},\notag
	\end{align}
	Hence the marginal posterior over all the latent parameter functions is build as,
	\begin{align}
	q(\mathbf{f})=\prod_{d=1}^{D}\prod_{j=1}^{Jd}q(\mathbf{f}_{d,j}),\notag
	\end{align}
	where each
	\begin{align}
	q(\mathbf{f}_{d,j})&=\int p(\mathbf{f}_{d,j}|\mathbf{\check{u}}_{d,j})q(\mathbf{\check{u}}_{d,j})d\mathbf{\check{u}}_{d,j}.\notag
	\end{align}
	This way we can keep developing the ELBO,
	\begin{align}
	\mathcal{L}&=\mathbb{E}_{q(\mathbf{f})}\Big[\log p(\mathbf{y}|\mathbf{f})\Big]+\mathbb{E}_{q(\mathbf{\check{u}})}\Big[\log \prod_{d=1}^{D}\prod_{j=1}^{J_d}\frac{p(\mathbf{\check{u}}_{d,j})}{q(\mathbf{\check{u}}_{d,j})}\Big]\notag\\
	&=\mathbb{E}_{q(\mathbf{f})}\Big[\log \prod_{d=1}^{D}\prod_{n=1}^{N}p(y_{d,n}|\psi_{d,1}(\mathbf{x}_n),...,\psi_{d,J_d}(\mathbf{x}_n))\Big]+\mathbb{E}_{q(\mathbf{\check{u}})}\Big[\log \prod_{d=1}^{D}\prod_{j=1}^{J_d}\frac{p(\mathbf{\check{u}}_{d,j})}{q(\mathbf{\check{u}}_{d,j})}\Big]\notag\\
	&=\sum_{d=1}^{D}\sum_{n=1}^{N}\mathbb{E}_{q(\mathbf{f})}\Big[\log p(y_{d,n}|\psi_{d,1}(\mathbf{x}_n),...,\psi_{d,J_d}(\mathbf{x}_n))\Big]-\sum_{d=1}^{D}\sum_{j=1}^{J_d}\mathbb{D}_{\text{KL}}(q(\mathbf{\check{u}}_{d,j})||p(\mathbf{\check{u}}_{d,j})).\notag    
	\end{align}
	We write again as a negative ELBO:
	\begin{align}
	\mathcal{\tilde{L}}=\sum_{n=1}^{N}\sum_{d=1}^{D}&\mathbb{E}_{q(\mathbf{f}_{d,1}) \cdots q(\mathbf{f}_{d,J_d})}\left[g_{d,n}\right]+\sum_{d=1}^{D}\sum_{j=1}^{J_d} \mathbb{D}_{KL}\left(q(\mathbf{\check{u}}_{d,j}) \| p(\mathbf{\check{u}}_{d,j})\right),\label{eq:convHetMOGP_appendix}
	\end{align}
	where $g_{d,n}=-\log p(y_{d,n}|\psi_{d,1}(\mathbf{x}_n),...,\psi_{d,J_d}(\mathbf{x}_n))$ is the NLL function associated to each output.
	
	\section{Computing the Gradients w.r.t the Posterior' Parameters}
	The computation of the gradients
	$\hat{\nabla}_{\boldsymbol{\Sigma}}\tilde{\mathcal{F}}$ and
	$\hat{\nabla}_{\boldsymbol{\mu}}\tilde{\mathcal{F}}$ is directly
	influenced by the penalisation (or prior) distribution
	$p(\boldsymbol{\theta})=\mathcal{N}(\boldsymbol{\theta}|\mathbf{0},\lambda_1^{-1}\mathbf{I})$ with precision $\lambda_1>0$. Using the Gaussian
	identities, we can express the gradients as follows: 
	\begin{align}
	\hat{\nabla}_{\boldsymbol{\mu}}\tilde{\mathcal{F}}&=\mathbb{E}_{q(\boldsymbol{\theta})}\big[\hat{\nabla}_{\boldsymbol{\theta}}\tilde{\mathcal{L}}\big]+\lambda_1\boldsymbol{\mu}\notag\\
	\hat{\nabla}_{\boldsymbol{\Sigma}}\tilde{\mathcal{F}}&=\frac{1}{2}\mathbb{E}_{q(\boldsymbol{\theta})}\Big[\hat{\nabla}^{2}_{\boldsymbol{\theta}\boldsymbol{\theta}}\tilde{\mathcal{L}}\Big]+\frac{1}{2}\lambda_1\mathbf{I}-\frac{1}{2}\boldsymbol{\Sigma}^{-1}\notag.
	\end{align}
	The other gradients $\hat{\nabla}_{\mathbf{m}_{(\cdot)}}\tilde{\mathcal{F}}=\mathbb{E}_{q(\boldsymbol{\theta})}[\hat{\nabla}_{\mathbf{m}_{(\cdot)}}\tilde{\mathcal{L}}]$ and $\hat{\nabla}_{\mathbf{V}_{(\cdot)}}\tilde{\mathcal{F}}=\mathbb{E}_{q(\boldsymbol{\theta})}[\hat{\nabla}_{\mathbf{V}_{(\cdot)}}\tilde{\mathcal{L}}]$ depend on the inner gradients $\hat{\nabla}_{\mathbf{m}}\tilde{\mathcal{L}}$ and $\hat{\nabla}_{\mathbf{V}}\tilde{\mathcal{L}}$ of the negative ELBO in Eq. \eqref{eq:HetMOGP_appendix} for LMC, or in Eq. \eqref{eq:convHetMOGP_appendix} for CPM.
	\subsection{Particular Gradients for Linear Model of Coregionalisation}
	Taking the derivative of $\mathcal{\tilde{L}}$ for the LMC w.r.t each parameter $\mathbf{m}_q$ and $\mathbf{V}_q$ we arrive to,
	\begin{align}
	\hat{\nabla}_{\mathbf{m}_q}\tilde{\mathcal{L}}&=\sum_{d=1}^{D}\sum_{j=1}^{J_d}\mathbf{A}_{\mathbf{f}_{d,j}\mathbf{u}_q}^\top \mathbf{g_m}_{d,j}+\mathbf{K}^{-1}_{\mathbf{u}_q\mathbf{u}_q}\mathbf{m}_{q},\label{eq:grad_mq}\\ \hat{\nabla}_{\mathbf{V}_q}\tilde{\mathcal{L}}&=\sum_{d=1}^{D}\sum_{j=1}^{J_d}\mathbf{A}_{\mathbf{f}_{d,j}\mathbf{u}_q}^{\top}\text{diag}(\mathbf{g_v}_{d,j})\mathbf{A}_{\mathbf{f}_{d,j}\mathbf{u}_q}-\frac{1}{2}\big[\mathbf{V}^{-1}_{q}-\mathbf{K}^{-1}_{\mathbf{u}_q\mathbf{u}_q}\big]\label{eq:grad_vq},
	\end{align}
	where $\mathbf{A}_{\mathbf{f}_{d,j}\mathbf{u}_q}=\mathbf{K}_{\mathbf{f}_{d,j}\mathbf{u}_q}\mathbf{K}^{-1}_{\mathbf{u}_q\mathbf{u}_q}$, the vector $\mathbf{g_m}_{d,j} \in \mathbb{R}^{N\times 1}$ has entries $\mathbb{E}_{q_{f_{d,1,n}},...,q_{f_{d,Jd,n}}}[{\nabla}_{f_{d,j,1}}g_{d,n}]$, the vector $\mathbf{g_v}_{d,j} \in \mathbb{R}^{N\times 1}$ has entries $\frac{1}{2}\mathbb{E}_{q_{f_{d,1,n}},...,q_{f_{d,J_d,n}}}[{\nabla}^{2}_{f_{d,j,n}f_{d,j,n}}g_{d,n}]$, and $\text{diag}(\mathbf{g_v}_{d,j})$ is a new matrix with the elements of $\mathbf{g_v}_{d,j}$ on its diagonal. Notice that each distribution $q_{f_{d,j,n}}$ represents the n-th marginal of each distribution $q_{\mathbf{f}_{d,j}}$. The above equations allow us to use mini-batches at each iteration of the inference process. Then, instead of using all data observations $N$, we randomly sample a mini-batch $\mathbf{X}_B\in \mathbb{R}^{B\times P}$ and $\mathbf{y}_B\in \mathbb{R}^{B\times D}$ from the dataset $D=\{\mathbf{X},\mathbf{y}\}$, here $B$ accounts for the mini-batch size. We simply construct: the matrix $\mathbf{A}_{\mathbf{f}_{d,j}\mathbf{u}_q}$ which becomes $\in \mathbb{R}^{B\times M}$, and the vectors $\mathbf{g_m}_{d,j}$ and $\mathbf{g_v}_{d,j}$ which become $\in \mathbb{R}^{B\times 1}$. Then we scale the first term to the right hand side of Eq. \eqref{eq:grad_mq} and Eq. \eqref{eq:grad_vq} by a factor of $N/B$. We refer to $D_B=\{\mathbf{X}_B,\mathbf{y}_B\}$ as the mini-batch data collection.
	\subsection{Particular Gradients for Convolution Processes Model}
	Taking the derivative of $\mathcal{\tilde{L}}$ for the CPM w.r.t each parameter $\mathbf{m}_{d,j}$ and $\mathbf{V}_{d,j}$ we find that,
	\begin{align}
	\hat{\nabla}_{\mathbf{m}_{d,j}}\tilde{\mathcal{L}}&=\mathbf{A}_{\mathbf{f}_{d,j}\mathbf{\check{u}}_{d,j}}^\top \mathbf{\check{g}}_{\mathbf{m}_{d,j}}+\mathbf{K}^{-1}_{\mathbf{\check{u}}_{d,j}\mathbf{\check{u}}_{d,j}}\mathbf{m}_{d,j},\notag\\ \hat{\nabla}_{\mathbf{V}_{d,j}}\tilde{\mathcal{L}}&=\mathbf{A}_{\mathbf{f}_{d,j}\mathbf{\check{u}}_{d,j}}^{\top}\text{diag}(\mathbf{\check{g}}_{\mathbf{v}_{d,j}})\mathbf{A}_{\mathbf{f}_{d,j}\mathbf{\check{u}}_{d,j}}-\frac{1}{2}\big[\mathbf{V}^{-1}_{d,j}-\mathbf{K}^{-1}_{\mathbf{\check{u}}_{d,j}\mathbf{\check{u}}_{d,j}}\big]\notag,
	\end{align}
	where $\mathbf{A}_{\mathbf{f}_{d,j}\mathbf{\check{u}}_{d,j}}=\mathbf{K}_{\mathbf{f}_{d,j}\mathbf{\check{u}}_{d,j}}\mathbf{K}^{-1}_{\mathbf{\check{u}}_{d,j}\mathbf{\check{u}}_{d,j}}$, the vector $\mathbf{\check{g}}_{\mathbf{m}_{d,j}} \in \mathbb{R}^{N\times 1}$ has entries $\mathbb{E}_{q_{f_{d,1,n}},...,q_{f_{d,Jd,n}}}[{\nabla}_{f_{d,j,1}}g_{d,n}]$, the vector $\mathbf{\check{g}}_{\mathbf{v}_{d,j}} \in \mathbb{R}^{N\times 1}$ has entries $\frac{1}{2}\mathbb{E}_{q_{f_{d,1,n}},...,q_{f_{d,J_d,n}}}[{\nabla}^{2}_{f_{d,j,n}f_{d,j,n}}g_{d,n}]$, and $\text{diag}(\mathbf{\check{g}}_{\mathbf{v}_{d,j}})$ is a new matrix with the elements of $\mathbf{\check{g}}_{\mathbf{v}_{d,j}}$ on its diagonal. Notice that each distribution $q_{f_{d,j,n}}$ represents the n-th marginal of each distribution $q(\mathbf{f}_{d,j})$.
	
	\section{Algorithm}
	Algorithm~\ref{alg:fng_algorithm} shows a pseudo-code
	implementation of the proposed method. In practice, we found useful
	to update the parameters $\boldsymbol{\mu}_{t+1}$ (in Eq. (25) of the manuscript) using $\sqrt{\mathbf{p}_t}$ and $\sqrt{\mathbf{p}_{t+1}}$ instead of
	$\mathbf{p}_t$ and $\mathbf{p}_{t+1}$, for improving the method's convergence.
	\begin{algorithm}
		\caption{Fully Natural Gradient Algorithm}
		\begin{algorithmic}[1]
			\renewcommand{\algorithmicrequire}{\textbf{Input:}}
			\renewcommand{\algorithmicensure}{\textbf{Output:}}
			\REQUIRE $\alpha_t,\beta_t,\gamma_t,\nu_t, \lambda_1$
			\ENSURE  $\boldsymbol{\Sigma}_{t+1}, \boldsymbol{\mu}_{t+1}$, $\mathbf{V}_{(\cdot),t+1}$, $\mathbf{m}_{(\cdot),t+1}$
			%\\ \textit{Initialisation} :
			\STATE set $t=1$ %$\boldsymbol{\Sigma}_{1}, \boldsymbol{\mu}_{1}$, $\mathbf{V}_{q,1}$, $\mathbf{m}_{q,1}$
			\WHILE {Not Converged}
			\STATE sample $\boldsymbol{\theta}_t\sim q(\boldsymbol{\theta}|\boldsymbol{\mu}_t,\boldsymbol{\Sigma}_t)$ 
			\STATE randomly sample a mini-batch $D_B$
			\STATE  $\mathbb{E}_{q(\boldsymbol{\theta})}\big[\hat{\nabla}_{\boldsymbol{\theta}}\tilde{\mathcal{L}}\big]$ and $\mathbb{E}_{q(\boldsymbol{\theta})}\big[\hat{\nabla}_{\boldsymbol{\theta}}\tilde{\mathcal{L}}\circ \hat{\nabla}_{\boldsymbol{\theta}}\tilde{\mathcal{L}}\big]$ using samples $\boldsymbol{\theta}_t$
			%\STATE compute $\hat{\nabla}_{\boldsymbol{\mu}}\tilde{\mathcal{F}}$ and $\hat{\nabla}_{\boldsymbol{\Sigma}}\tilde{\mathcal{F}}$		
			\STATE update $\mathbf{p}_{t+1}$ and $\boldsymbol{\mu}_{t+1}$
			\STATE compute $\hat{\nabla}_{\mathbf{m}_{(\cdot)}}\tilde{\mathcal{F}}$ and $\hat{\nabla}_{\mathbf{V}_{(\cdot)}}\tilde{\mathcal{F}}$
			\STATE update $\mathbf{V}_{(\cdot),t+1}$ and $\mathbf{m}_{(\cdot),t+1}$
			\STATE $\boldsymbol{\Sigma}_{t+1} = \text{diag}\left((\mathbf{p}_{t+1}+\lambda_1\mathbf{1})^{-1}\right)$
			\STATE $t=t+1$
			\ENDWHILE 
		\end{algorithmic}
		\label{alg:fng_algorithm} 
	\end{algorithm}
	
	\section{Details on: Maximum a Posteriori in the Context of Variational Inference}
	In context of Bayesian inference, posterior distribution $p(\boldsymbol{\theta}|\mathbf{X})$ is proportional to the likelihood $p(\mathbf{X}|\boldsymbol{\theta})$ times the prior $p(\boldsymbol{\theta})$ , i.e., $ p(\boldsymbol{\theta}|\mathbf{X}) \propto p(\mathbf{X}|\boldsymbol{\theta})p(\boldsymbol{\theta})$. Although, if the likelihood and prior are non-conjugate distributions, it is necessary to approximate the posterior, for instance using variational inference. In this context of variational inference, we do not have access to the true posterior, but to the approximate posterior $q(\boldsymbol{\theta})$, which it is optimised by maximising the ELBO,
	\begin{align}
	\mathcal{L}=\mathbb{E}_{q(\boldsymbol{\theta})}\big[\log p(\mathbf{X}|\boldsymbol{\theta})\big] - \mathbb{D}_{KL}(q(\boldsymbol{\theta})||p(\boldsymbol{\theta})) \leq \log p(\mathbf{X}).\notag
	\end{align}
	Notice that if we are only interested in a point estimate of the parameter $\boldsymbol{\theta}$ of the Log Likelihood
	function, then a feasible solution for the parameter is
	$\boldsymbol{\theta}^{\star}=\mathbb{E}_{q(\boldsymbol{\theta})}[\boldsymbol{\theta}]=\boldsymbol{\mu}$, where $q(\boldsymbol{\theta}):=q(\boldsymbol{\theta}|\boldsymbol{\mu},\boldsymbol{\Sigma})$. This corresponds to the MAP solution due to the fact that, 
	\begin{align}
	\boldsymbol{\theta}_{\text{MAP}}=\arg \max_{\boldsymbol{\theta}} p(\boldsymbol{\theta}|\mathbf{X}),\notag
	\end{align}
	where $p(\boldsymbol{\theta}|\mathbf{X})$ represents the true posterior. Since in the context of variational inference, we only have access to an approximate free parametrised posterior $p(\boldsymbol{\theta}|\mathbf{X}) \approx q(\boldsymbol{\theta}|\boldsymbol{\mu},\boldsymbol{\Sigma})$, therefore the equation above implies that, 
	\begin{align}
	\boldsymbol{\theta}_{\text{MAP}}=\arg \max_{\boldsymbol{\theta}} q(\boldsymbol{\theta}|\boldsymbol{\mu},\boldsymbol{\Sigma}),\notag
	\end{align}
	and it is clear that the maximum of the distribution
	$q(\boldsymbol{\theta}|\boldsymbol{\mu},\boldsymbol{\Sigma})$ is
	located at its mean, thereby
	$\boldsymbol{\theta}_{\text{MAP}}=\boldsymbol{\mu}$.
	
	\begin{figure*}[t]
		\centering
		{\hspace{-0.15cm}\includegraphics[width=0.33\textwidth,height=0.19\textheight]{ELBO_london.pdf}
			\hspace{-0.15cm}\includegraphics[width=0.33\textwidth,height=0.19\textheight]{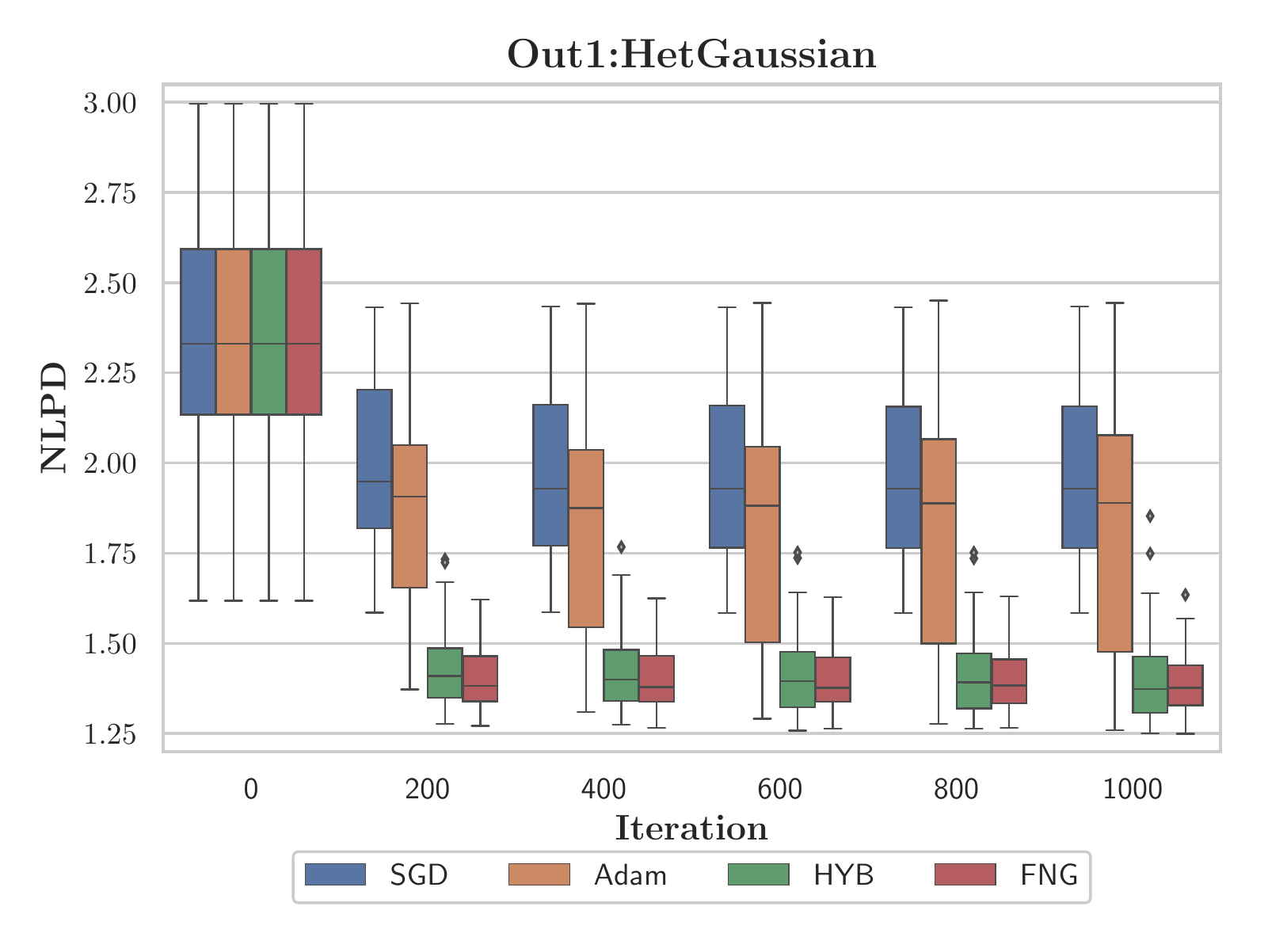}
			\hspace{-0.15cm}\includegraphics[width=0.33\textwidth,height=0.19\textheight]{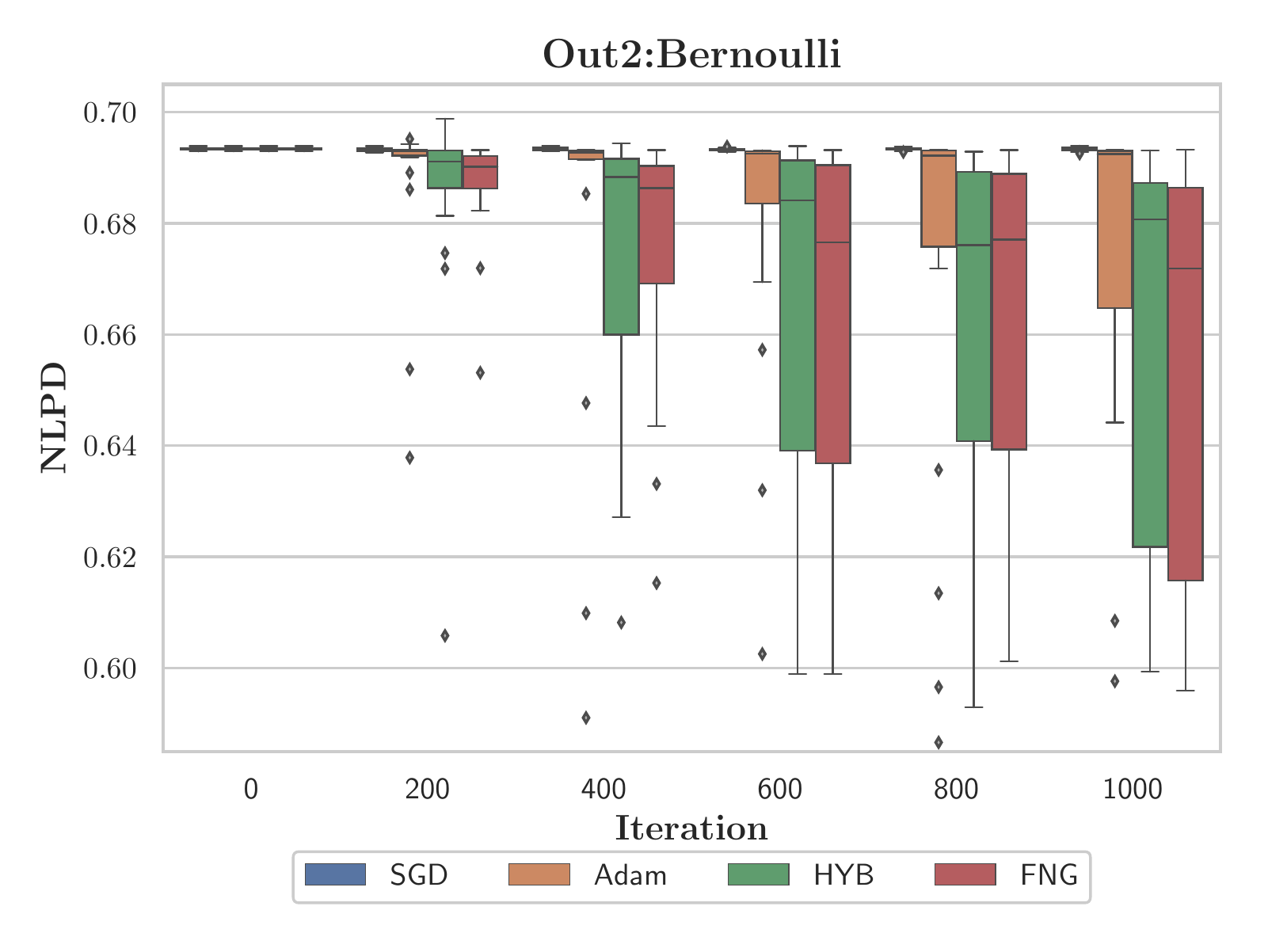}}
		\caption{Performance of the diverse inference methods on the LONDON dataset using 20 different initialisations over HetMOGP with LMC. The left sub-figure shows the average NELBO convergence of each method. The other sub-figures show the box-plot trending of the NLPD over the test set for each output. The box-plots at each iteration follow the legend's order from left to right: SGD, Adam, HYB and FNG. The isolated diamonds that appear in the outputs' graphs represent ``outliers".}
		\label{fig:london}
	\end{figure*}
	\begin{figure*}[t]
		\centering
		{\hspace{-0.15cm}\includegraphics[width=0.33\textwidth,height=0.19\textheight]{ELBO_naval.pdf}
			\hspace{-0.15cm}\includegraphics[width=0.33\textwidth,height=0.19\textheight]{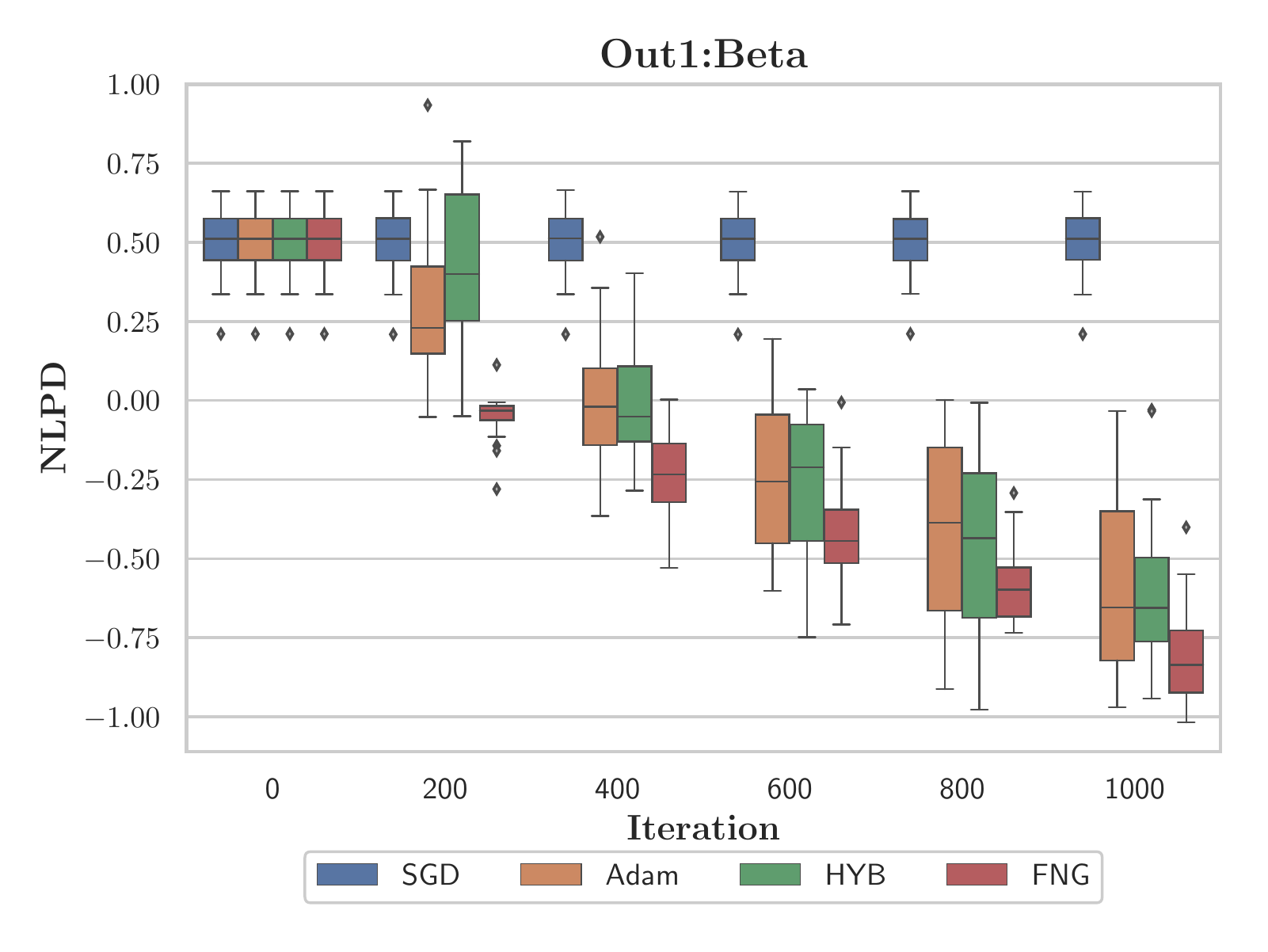}
			\hspace{-0.15cm}\includegraphics[width=0.33\textwidth,height=0.19\textheight]{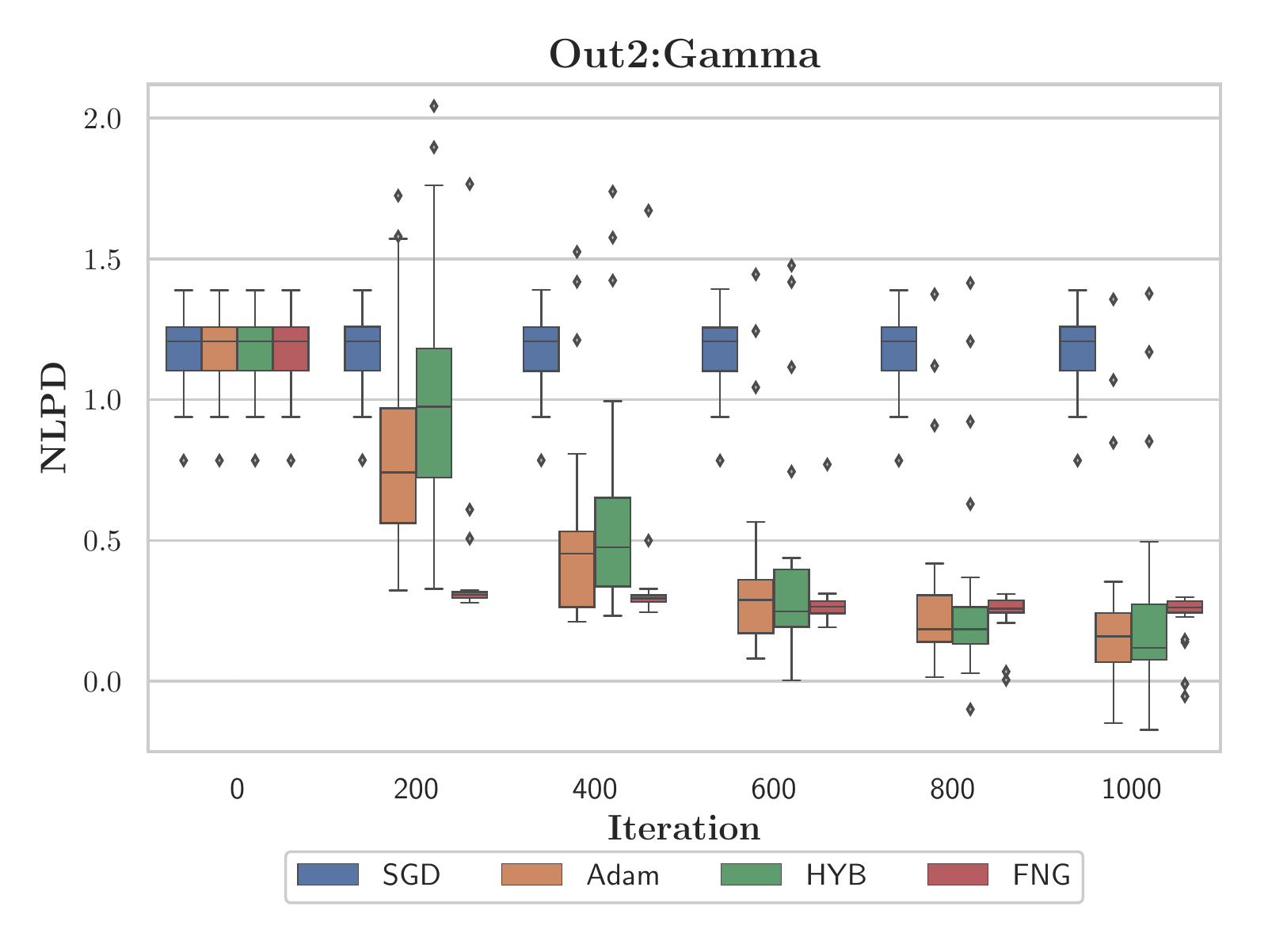}}
		\caption{Performance of the diverse inference methods on the NAVAL dataset using 20 different initialisations over HetMOGP with LMC. The left sub-figure shows the average NELBO convergence of each method. The other sub-figures show the box-plot trending of the NLPD over the test set for each output. The box-plots at each iteration follow the legend's order from left to right: SGD, Adam, HYB and FNG. The isolated diamonds that appear in the output graphs represents ``outliers".}
		\label{fig:naval}
	\end{figure*}
	
	\section{Additional Information of Datasets and Experiments Setting}
	
	\noindent\textbf{TRAFFIC Dataset:} a record of vehicles traffic, it contains a per-day-number of vehicles passing by the main roads and
	streets of London city (TRAFFIC, $N=1712, P=3, D=4$). We use a Poisson likelihood per each output of TRAFFIC and assume $Q=4$ latent functions.\medskip
	
	\noindent\textbf{MOCAP9 Dataset:} a motion capture data for a running subject (MOCAP9, $N=744, P=1, D=20$). We use a HetGaussian distribution as the likelihood for each output and assume $Q=3$ functions.\medskip
	
	The datasets used in our experiments were taken from the following web-pages:
	\begin{itemize}
		\item The HUMAN is captured using EB2 \textit{app}, visit https://www.eb2.tech/
		\item For information about LONDON dataset visit https://www.gov.uk/government/collections/price-paid-data
		\item For information about NAVAL dataset visit http://archive.ics.uci.edu/ml/datasets 
		\item For information about SARCOS dataset see http://www.gaussianprocess.org/gpml/data/
		\item See http://mocap.cs.cmu.edu/subjects.php for MOCAP dataset, subject 7 refers to MOCAP7 and subject 9 refers to MOCAP9.
		\item Visit https://data.gov.uk/dataset/208c0e7b-353f-4e2d-8b7a-1a7118467acc/gb-road-traffic-counts for information about TRAFFIC dataset.
	\end{itemize}
	\subsection{Additional Analysis per Output over LONDON and NAVAL datasets}
	For the LONDON dataset, Fig. \ref{fig:london} shows that Adam converges to a richer minimum of the NELBO
	than SGD. Moreover, the NLPD for Adam is, on average, better than the
	SGD for both HetGaussian and Bernoulli outputs. Particularly, Adam
	presents for the Bernoulli output few ``outliers" under its boxes that
	suggest it can find sporadically rich local optima, but its general
	trend was to provide poor solutions for that specific output in
	contrast to the HetGaussian output. The HYB and FNG arrive to a very
	similar value of the NELBO, both being better than Adam and SGD. HYB
	and FNG methods attain akin NLPD metrics for the HetGaussian output,
	though our method shows smaller boxes being more confident along
	iterations. Both methods present large variances for the Bernoulli
	output, but the average and median trend of our approach is much
	better, being more robust to the initialisation than HYB method.
	
	The NLPD performance for the NAVAL dataset shows in
	Fig. \ref{fig:naval} that the SGD method cannot make
	progress. We tried to set a bigger step-size, but usually increasing
	it derived in numerical problems due to ill-conditioning. The methods
	Adam and HYB show almost the same behaviour along the NELBO
	optimisation, in fact the NLPD boxes for the Beta and Gamma outputs
	look quite similar for both methods. The difference of performance can
	be noticed for the Beta output, where at the end, HYB method becomes
	more confident reducing its variance. Our FNG method ends up with a
	slightly upper NLPD solution in the Gamma output in comparison to Adam
	and HYB, but being more confident showing a smaller spread in the
	box-plot across iterations. For the Gamma output, FNG shows at the end
	some ``outliers" under the NLPD boxes, accounting for sporadic
	convergence to strong solutions. For the Beta distribution, our method
	obtains a better solution with the finest NLPD in comparison to SGD,
	HYB and Adam.
\end{appendices}

\end{document}